\documentclass[letterpaper, 10 pt, conference]{ieeeconf}  

\usepackage{cite}
\usepackage{amsmath,amssymb,amsfonts}
\usepackage{algorithmic}
\usepackage{graphicx}
\usepackage{subfigure}
\usepackage{textcomp}
\usepackage{xcolor}
\usepackage{subcaption}
\usepackage{svg}
\usepackage{amsmath}
\usepackage[lined,commentsnumbered,ruled]{algorithm2e}
\usepackage{xcolor}
\usepackage{booktabs}  
\usepackage{siunitx}   
\usepackage{array}
\usepackage{xcolor}
\usepackage{makecell}
\usepackage[hidelinks]{hyperref}
\usepackage{mathrsfs}
\usepackage{bm}
\usepackage{blkarray}   
\usepackage{booktabs, multirow, nicematrix}
\usepackage{placeins}

\newcommand{\argminx}[1]{\underset{#1}{\mathrm{arg\,min}}}
\definecolor{mygray}{gray}{0.9}

\definecolor{mygray}{gray}{0.9}

\IEEEoverridecommandlockouts                              

\title{\LARGE \bf
DFL-TORO: A One-Shot Demonstration Framework for Learning Time-Optimal Robotic Manufacturing Tasks
}

\author{
    Alireza Barekatain$^{1}$, Hamed Habibi$^{1}$, and Holger Voos$^{1}$ 
    \thanks{$^{1}$Authors are with the Automation and Robotics Research Group, Interdisciplinary Centre for Security, Reliability, and Trust (SnT), University of Luxembourg, Luxembourg. Holger Voos is also associated with the Faculty of Science, Technology, and Medicine, University of Luxembourg, Luxembourg. \tt{\small{\{alireza.barekatain, hamed.habibi, holger.voos\}}@uni.lu}}
    \thanks{This work is supported by FNR "Fonds national de
la Recherche" (Luxembourg) through Industrial Fellowship Ph.D. grant (ref. 15882013).}
    \thanks{Special thanks to Dr. Paul Kremer for his guidance and support throughout the experiments of this project.}
}

\begin{document}

\maketitle
\thispagestyle{empty}
\pagestyle{empty}

\maketitle

\maketitle

\begin{abstract}
This paper presents DFL-TORO, a novel \textbf{D}emonstration Framework for \textbf{L}earning \textbf{T}ime-Optimal \textbf{R}obotic tasks via \textbf{O}ne-shot kinesthetic demonstration. It aims at optimizing the process of Learning from Demonstration (LfD), applied in the manufacturing sector. As the effectiveness of LfD is challenged by the quality and efficiency of human demonstrations, our approach offers a streamlined method to intuitively capture task requirements from human teachers, by reducing the need for multiple demonstrations. Furthermore, we propose an optimization-based smoothing algorithm that ensures time-optimal and jerk-regulated demonstration trajectories, while also adhering to the robot's kinematic constraints. The result is a significant reduction in noise, thereby boosting the robot's operation efficiency. Evaluations using a Franka Emika Research 3 (FR3) robot for a variety of tasks further substantiate the efficacy of our framework, highlighting its potential to transform kinesthetic demonstrations in contemporary manufacturing environments. Moreover, we take our proposed framework into a real manufacturing setting operated by an ABB YuMi robot and showcase its positive impact on LfD outcomes by performing a case study via Dynamic Movement Primitives (DMPs).
\end{abstract}

\section{Introduction}
\label{sec:intro}

\subsection{Background and Motivation}
The manufacturing industry is experiencing a significant shift from mass production to mass customization . Traditionally, mass production focused on creating large quantities of standardized goods using assembly lines for efficiency, cost reduction, and high throughput, though it offered limited flexibility for changes or customization. In contrast, mass customization addresses the growing consumer demand for personalized products by combining the efficiency of mass production with the flexibility of custom-made items, allowing manufacturers to produce tailored goods in high-volume small batches while still maintaining cost-effectiveness and efficiency \cite{pedersen2016robot,cohen2019assembly}.

Robotic manipulation is crucial in this transition, as it demands more flexible manufacturing systems. Traditional industrial robots in mass production performed repetitive tasks in structured environments with consistent movements. In contrast, mass customization requires advanced robots equipped with vision technologies, AI-driven decision-making, and increased dexterity to adapt to diverse and changing products \cite{wind2001customerization,gavspar2020smart}. Collaborative robots, or cobots, enhance this transition by working alongside human operators, fostering flexible, agile production processes. These cobots improve safety and facilitate intuitive human-robot interaction, making them ideal for environments with varied and dynamic tasks that mass customization demands \cite{el2019cobot}.

With the adoption of cobots for mass customization, it is necessary to rethink programming methodologies to enhance both the efficiency of implementation and the efficiency of operation. Implementation efficiency refers to the effort, time, resources, and costs involved in transitioning a robotic task from concept to operational status, ensuring seamless integration of new tasks. Conversely, operation efficiency focuses on maximizing throughput and success rates during the robot's working phase, aiming to minimize operational costs like energy consumption and maintenance. Effective programming methodologies must balance these efficiencies to maintain high productivity and adaptability in the production process \cite{liu2022robot}.

Traditional programming for robots involves writing coordinate-based scripts to control their behavior for specific tasks. These programs are rigid and tailored for predefined tasks, which work well for mass production but struggle with the dynamic needs of mass customization \cite{sloth2020towards}. Cobots, designed to operate in shared, unstructured environments with humans, add complexity to programming. This complexity grows exponentially with the need for diverse customizations, affecting program efficiency and production cycle times. Modifying these hand-coded programs for different tasks is time-consuming and leads to increased downtime and lower efficiency \cite{heimann2020industrial}. 

To address the limitations of manual programming, motion planning and optimization methods have been developed. These methods allow for higher-level programming by creating a robot model and using algorithms to optimize actions based on predefined costs and constraints, effectively managing the complexity of tasks in dynamic and unstructured environments \cite{leger2016off}. This approach improves operational efficiency but remains highly task-specific, requiring substantial reprogramming for different tasks or customization needs, which is inefficient for high-mix, low-volume production. Additionally, both manual and optimization-based programming methods demand significant robotic expertise, necessitating an expert intermediary between the task specialist and the robot, thereby increasing implementation costs and impacting efficiency.

The most recent paradigm for flexible cobot programming is Learning from Demonstration (LfD), or Imitation Learning (IL), where robots learn tasks by observing and imitating human demonstrations \cite{barekatain2024practical, ravichandar2020recent}. Unlike manual programming, LfD allows non-experts to teach complex behaviors without needing to provide explicit motion sequences. It also contrasts with motion planning and optimization, as it does not require modeling the robot's environment or defining explicit costs and constraints. Instead, LfD learns implicit task requirements from the demonstrations, achieving optimal behavior and efficient operation. This approach enhances implementation efficiency by enabling rapid and direct programming by non-expert users, eliminating the need for a robot expert to model environments or design specific optimizations for each task.

While LfD offers significant conceptual advantages over other programming methods, its deployment in practical manufacturing settings requires careful consideration and faces performance challenges in terms of implementation and operational efficiency. To maximize implementation efficiency, it is necessary to consider various factors such as the cost of demonstration setup, as well as the human time and cognitive effort required for demonstration. According to \cite{ravichandar2020recent}, kinesthetic teaching is particularly effective due to its minimal setup requirements and ease of demonstration to cobots. To further minimize human effort and time, focusing on one-shot demonstrations is crucial \cite{li2024learning}. Among LfD approaches, Dynamic Movement Primitive (DMP) is a well-established method that can learn and generalize from a single demonstration. The choice of DMP in manufacturing is also further motivated by its high explainability and low implementation effort \cite{barekatain2024practical}.

The efficiency of operation can be directly evaluated through task success rates and production throughput. While aiming for a 100\% success rate and maximum throughput would theoretically maximize daily product output, achieving this ideal is nontrivial due to practical challenges. The noise in the recorded demonstration data, caused by joint sensor errors or hand tremors during demonstrations, negatively impacts operational efficiency in two significant aspects: accuracy and execution time.

Accuracy in LfD is determined by task-specific criteria that ensure the success of the task, similar to optimization-based approaches where constraints and costs guide an algorithm towards a successful solution. In human demonstrations, these constraints and costs are inherently encoded in the motion. For example, in a reaching task, the end-effector must approach the object from a specific direction to avoid collisions and grasp the object at the correct angle. These precise instructions are implicitly conveyed through human demonstration. However, noise can obscure these implicit cues, making them less effective. This degradation affects LfD algorithms' performance, making it challenging for the system to learn and generalize the intended motion, often leading to overfitting the noise and reducing task success rates \cite{li2024learning}.

Existing methods often rely on multiple demonstrations to regress noise-free demonstrations while maintaining accuracy \cite{tsai2020constrained, calinon2016tutorial}. To numerically model the accuracy, they use the local variance of these demonstrations as a measure of significance for each segment: high local variability indicates low accuracy and vice versa. However, this approach compromises implementation efficiency by requiring the user to provide multiple demonstrations. To overcome this challenge, it is necessary to extract such accuracy measures while providing only one demonstration.

The execution time of robotic tasks is a crucial factor that directly influences production throughput. A lower execution time leads to a higher number of products produced per day and naturally a higher profit. Therefore, it is necessary to learn the fastest possible way to execute a task from human demonstration \cite{moreno2024obstacles}. However, human demonstrations are naturally slow due to the cognitive complexity of teaching accurate, detailed motions \cite{meszaros2022learning}. Consequently, it is necessary to derive the optimal timing law from human demonstrations. 
Achieving optimal timing for robotic demonstrations is challenging due to two main issues. First, as detailed in \cite{meszaros2022learning}, demonstrations cannot typically be sped up uniformly, hence a simple uniform scaling is not feasible. Second, while existing methods allow humans to locally speed up their demonstrations, they ignore the robot's kinematic limits. Due to the presence of noise, this approach leads to velocity and acceleration spikes that make the resulting trajectory kinematically infeasible. To avoid infeasibility, it is inevitable to settle for a non-optimal timing law. Additionally, this approach leads to a high-jerk motion. the motion jerk, indicating the rate of change of acceleration, affects motion smoothness. Low jerk profiles indicate smooth, human-like movement, which reduces mechanical stress, whereas high jerk increases wear, maintenance costs, and energy consumption \cite{wang2021optimised}. Therefore, there is a need for a new approach to achieve the optimal execution time, while removing the demonstration noise and enhancing the smoothness of the motion by minimizing jerk.

\subsection{Related Work}
\label{sec:works}

\begin{figure*}[htbp]
    \centering
    \includegraphics[width=0.7\linewidth]{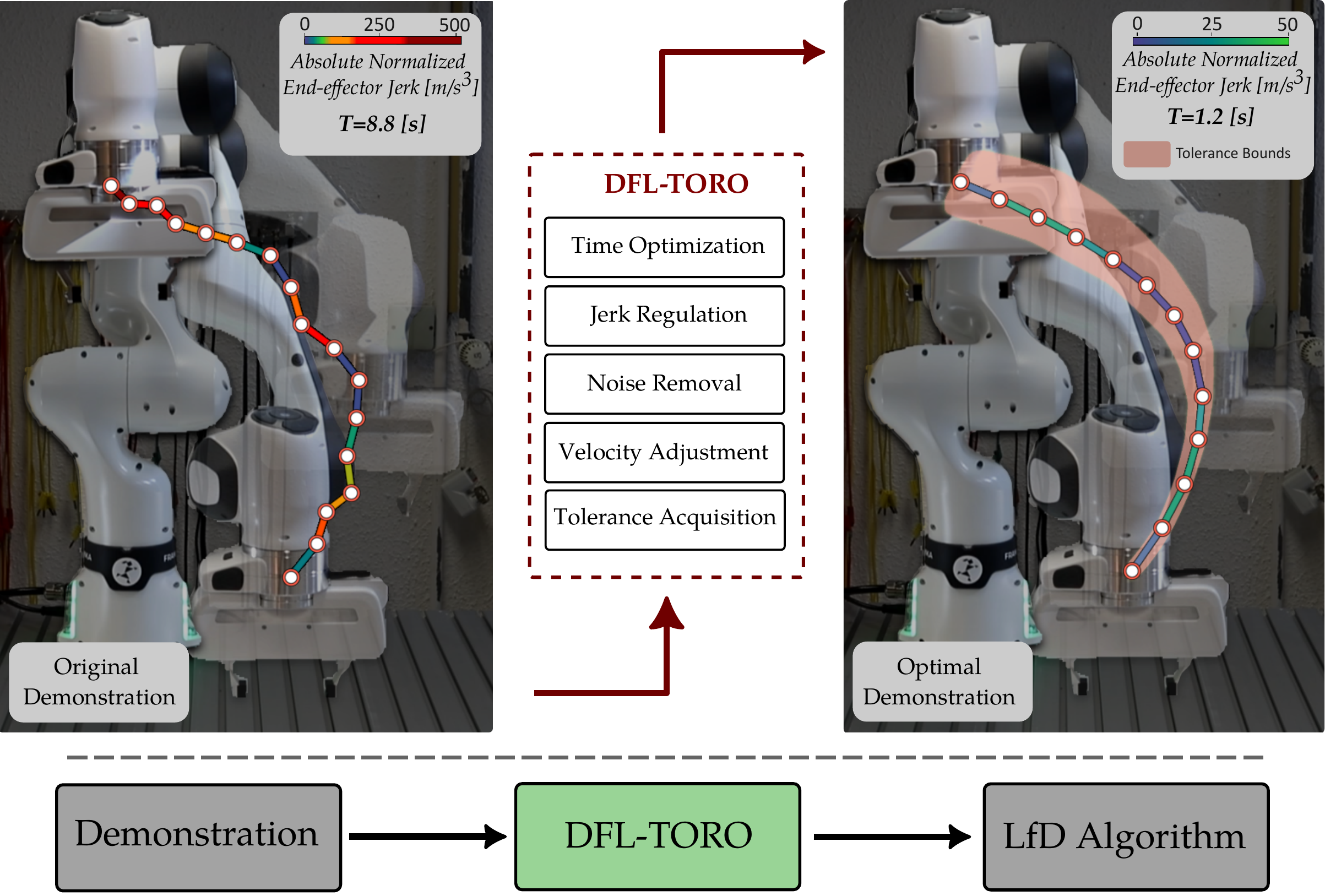}
    \caption{Advantages of DFL-TORO, Transforming original kinesthetic demonstration trajectories into time-optimal, noise-free, and jerk-regulated demonstrations with the possibility to independently refine the demonstration velocity profile. DFL-TORO acts as an intermediary layer between capturing demonstrations and feeding them into the LfD algorithm.
    }
    \label{fig:intro}
\end{figure*}

To address the issue of slow demonstrations and execution time, several studies focused on the isolation of demonstrating path and timing, allowing humans to teach the path and the timing law of demonstrations separately. Authors in \cite{nemec2018efficient} explore methods based on incremental learning to let the human teacher adjust the speed of their initial demonstrations, integrating human feedback into velocity scaling factors within the DMP formulation. The work in \cite{simonivc2021analysis} further investigates the refinement of both path and speed through kinesthetic guidance, enabling teachers to fine-tune their demonstrations in real time, thereby enhancing the quality of the learned trajectories. Moreover, the study of \cite{meszaros2022learning} utilizes teleoperated feedback to locally speed up the task execution. Their work clearly signifies the importance of a well-demonstrated timing law in achieving high task success rates. To build upon these works, it is beneficial to incorporate the robot's kinematic limits into an optimization problem, solving for the best feasible timing law for the demonstrated path. Then, to allow the human teacher to refine the timing law, instead of merely speeding up, we can permit users to ``slow down'' demonstrations to achieve reliable execution, and hence ensure the optimal timing. Via this approach, optimizing for the timing law guarantees that the best achievable execution time is calculated, while the ability to slow down the demonstration provides direct control to the human teacher in order to make the task execution reliable and successful while maintaining fast performance.


In the context of LfD, there are limited studies addressing the regulation of motion jerk. In \cite{wang2021optimised}, authors tackled jerky demonstrations by introducing Gaussian noise to smoothen the trajectories. The study in \cite{aleotti2006robust} utilized the inherent smoothing properties of B-Splines by fitting demonstration trajectories with them, an approach also adopted by \cite{biagiotti2023robot} and \cite{shyam2019improving}. B-splines are advantageous for trajectory optimization due to their smoothness, local control, and efficient computation. Their piece-wise polynomial structure ensures smooth motion paths, making them ideal for complex motion optimization \cite{de1972calculating}.
However, optimizing both timing and jerk simultaneously presents a challenge, as it can lead to an ill-defined problem. This difficulty arises because applying constraints over the demonstration path requires knowing the relative timing of each part of the path. Conversely, these timings cannot be treated as decision variables since the underlying trajectory representation in B-splines is a piece-wise polynomial \cite{yang2018new}. As the original timing of the demonstration is suboptimal and cannot be used, it is possible to address this issue by employing a two-step solution: first, optimizing the timing, and then regulating the jerk profile by removing noise.

When it comes to eliminating noise and unwanted behaviors in human demonstrations, the existing studies have taken various approaches to transform suboptimal demonstrations into optimal ones and enhance the learning and execution performance of LfD algorithms such as DMPs, Gaussian Processes (GPs), and Gaussian Mixture Models (GMMs). The performance of these methods relies heavily on the quality of the demonstrations. suboptimal demonstrations can reduce the performance of LfD and cause overfitting or underfitting on the non-optimal demonstrations \cite{li2024learning}.
One approach is to focus on augmenting the demonstrations in order to fine-tune their data efficiency \cite{ma2022efficient, qian2020hierarchical}, which aims at compensating the suboptimal demonstrations via various methods such as data augmentation or self-exploration. Another strategy focuses on extracting key information from demonstrations to capture essential implicit features \cite{he2021imitation}.
to enhance the efficiency of the demonstration process, the work in \cite{wu2021framework} assessed the effectiveness of newly added demonstrations by quantifying their information gain. The authors in \cite{el2021itp} optimized LfD outcomes via Reinforcement Learning (RL) by eliminating suboptimal demonstrations and retaining the useful ones. In \cite{ye2022bagging}, authors utilized GMM to increase the robustness of LfD to suboptimal training data, ensuring that the reproduced trajectory maintains the required precision.

A common theme in the mentioned works is the use of multiple demonstrations to optimize the demonstration process. This approach is driven by the absence of definitive ground truth in human demonstrations, making it challenging to distinguish between accurate demonstration data and noise or unwanted motion. The research question thus becomes how to develop a metric of accuracy that can serve as a ground truth while allowing some flexibility to modify or optimize the demonstration trajectories.
Studies such as \cite{tsai2020constrained} and \cite{calinon2016tutorial} highlight the use of task variability tolerances to adjust a robot's compliance. They suggest employing low impedance settings when high variability is detected, thereby accommodating more deviations in human demonstrations. The main idea is to align multiple demonstrations of a task and use the variance of each segment as a measure of significance or required accuracy. For example, a motion segment that is highly consistent across demonstrations indicates a need for precise replication, whereas segments with higher variability suggest permissible deviations while maintaining task accuracy.

Several studies have leveraged the statistical distribution of multiple demonstration trajectories, using variance to determine tolerance values that represent accuracy \cite{hu2022robot,paxton2015incremental,roveda2020assembly,wu2021framework}. Additionally, the study in \cite{delson1994robot} extracted tolerance bounds from multiple demonstrations to define obstacle-free regions. This approach provides flexibility to regress a noise-free trajectory within these bounds, ensuring the demonstration remains accurate, obstacle-free, and successful in achieving the task goal. Authors in \cite{maeda2008easy} proposed a method where the human first moves the manipulator around the workspace to define a swept volume. This swept volume is then used as tolerance bounds to constrain the optimization method, while the optimization solves for the desired objective.

The goal of all of these studies is to develop tolerance bounds as numerical representations of accuracy, which can then be used to safely constrain the optimization methods and apply desired objectives to improve demonstrations. However, since all of the mentioned works rely on multiple demonstrations, they sacrifice the clear advantages of one-shot demonstration. Only a limited number of works have attempted to address this challenge using one-shot demonstrations.
In \cite{muller2020one}, the authors used points recorded from a single demonstration to fit a piecewise polynomial trajectory within predefined and fixed tolerance bounds. Building on this, authors in \cite{biagiotti2023robot} proposed a novel approach to adaptively extract tolerance bounds from one-shot kinesthetic demonstrations. They employed surface Electromyography (sEMG) signals, where the stiffness of the human teacher's hand, recorded through sEMG measurements during the demonstration, serves as a measure of accuracy or tolerance bounds. However, this method requires an expensive setup, increasing both the cost and effort of implementation. In manufacturing contexts, it is beneficial to capture these tolerance bounds from the human teacher while maintaining the simplicity of the demonstration setup and implementation efficiency.

\subsection{Contribution and Structure}
Given the outlined challenges, this paper introduces DFL-TORO, a novel \textbf{D}emonstration \textbf{F}ramework for \textbf{L}earning \textbf{T}ime-\textbf{O}ptimal \textbf{R}obotic tasks via \textbf{O}ne-shot kinesthetic demonstration. DFL-TORO intuitively captures human demonstrations and obtains task tolerances, yielding smooth, jerk-regulated, time-optimal, and noise-free trajectories. As illustrated in Fig. \ref{fig:intro}, DFL-TORO is introduced as a pivotal layer after capturing human demonstrations and before feeding the data to the LfD algorithm. For the first time, we are addressing this issue by incorporating a new layer into the learning process.
Our main contributions are as follows:
 
\begin{itemize}
    \item An Optimization-based Smoothing algorithm, considering the robot's kinematic limits and task tolerances, which delivers time-and-jerk optimal trajectories and filters out the noise, enhancing operation efficiency. Our work is the very first attempt to optimize the original demonstration trajectory with respect to time, noise, and jerk, before feeding to the learning algorithm.
    \item A method for intuitive refinement of velocities and acquisition of tolerances, reducing the need for repetitive demonstrations and boosting operation efficiency.
    \item Evaluation of DFL-TORO for a variety of tasks via a Franka Emika Research 3 (FR3) robot, highlighting its superiority over conventional kinesthetic demonstrations
    \item A case study via DMP, to showcase the benefits of incorporating DFL-TORO before training LfD algorithms. In addition to FR3, we take our case study to a real manufacturing setting and evaluate the performance of DFL-TORO with DMP on a real manufacturing task operated by ABB dual-arm YuMi.
\end{itemize}

The rest of the paper is structured as follows. In Section \ref{sec:preliminary} the problem definition and the objectives are formulated in detail. Also, several technical preliminaries are given. In Section \ref{sec:method}, we introduce DFL-TORO along with its workflow and consisting modules. In section \ref{sec:result}, we provide and discuss the experimental results. Finally, the concluding remarks along with the limitations and potentials for future work are provided in Section \ref{sec:conclusion}. Throoughout the paper, the following notations are utilized. The bold variables, e.g. $\bm{q}$, represent a vector. $\mathbb{S}^3$ represent the the space of quaternions. Symbol ``$\leq$'' represents element-wise inequality of vectors when used with vector variables.

\section{Problem Statement and Preliminaries}
\label{sec:preliminary}


\subsection{Problem Statement}

Let $\bm{q}_o(t)$ be the initial noisy demonstration trajectory provided via kinesthetic guidance to an $n$ DoF manipulator. The path underlying the demonstration trajectory consists of $m$ waypoints, denoted by $w_i$ for $\forall i=1,\cdots,m$.
Each waypoint $w_i$ includes joint configuration $\bm{\mathfrak{q}}_{w_i} \in \mathbb{R}^n$, the end-effector's position $\bm{\mathfrak{p}}_{w_i}=[x_{w_i}, y_{w_i}, z_{w_i}]^T \in \mathbb{R}^3$ and orientation $\bm{\theta}_{w_i} \in \mathbb{S}^3$, represented as quaternion. The proposed framework aims to achieve the following objectives:

\begin{itemize}
    \item \textbf{OB1}: Given the $w_i$ and since the original timing law of $\bm{q}_o(t)$ is slow, we aim to solve for the time-optimal demonstration trajectory $\bm{q}_f(t)$. Moreover, the noise in the demonstration is inherent in the waypoints $w_i$ and needs to be eliminated in order to achieve the best achievable timing law.  We design an Optimization-based Smoothing approach that given the joint position limits $\bm{\mathfrak{q}}_{min}$, $\bm{\mathfrak{q}}_{max} \in \mathbb{R}^n$, joint velocity limits $\bm{\mathfrak{v}}_{min}$, $\bm{\mathfrak{v}}_{max} \in \mathbb{R}^n$, joint acceleration limits $\bm{\mathfrak{a}}_{min}$, $\bm{\mathfrak{a}}_{max} \in \mathbb{R}^n$, and joint jerk limits $\bm{\mathfrak{j}}_{min}$, $\bm{\mathfrak{j}}_{max} \in \mathbb{R}^n$, finds the optimal timing law for the demonstrated path constrained by $w_i$. Since jerk is the reflection of unfavored points in the demonstration \cite{li2024learning}, i.e. demonstration noise, our optimization objective also involves minimizing the motion jerk, which pushes the optimization to eliminate the noise from $w_i$ and deliver a smooth noise-free path. Finally, the path of $\bm{q}_f(t)$ must be constrained to $w_i$ via tolerance bounds to ensure that $\bm{q}_f(t)$ does not deviate from the original demonstration and the accuracy is retained.
    \item \textbf{OB2}:  The optimization process in \textbf{OB1} overrides the timing law of $\bm{q}_o(t)$ with the fastest timing law. Therefore, the objective is to provide a one-time Refinement Phase, where the human teacher can slow down the timing of $\bm{q}_f(t)$. In the meantime, the task tolerances $\bm{\epsilon}_p^i \in \mathbb{R}^3$ and $\epsilon_{\theta}^i \in \mathbb{R}$ are extracted for each waypoint $w_i$. Although the demonstrations are represented in the joint space, we extract tolerance bounds in the Cartesian space, as the end-effector pose is the ultimate point of accuracy for the task. Hence,  $\bm{\epsilon}_p^i$ is the allowed tolerance for the deviation from $\bm{\mathfrak{p}}_{w_i}$, and $\epsilon_{\theta}^i$ is the allowed tolerance for the deviation from $\bm{\theta}_{w_i}$ in terms of the value of angular difference. The outcome of this phase is fine-tuned trajectory $\bm{q}_f^r(t)$, taking tolerance values and timing modifications into account. $\bm{q}_f^r(t)$ is the final demonstration trajectory that not only respects the accuracy tolerance bounds of the task, but performs with an optimal execution time.
\end{itemize}

\begin{figure}[htbp]
    \centering
    \subfigure[Maze navigation]{
        \includegraphics[width=0.8\linewidth]{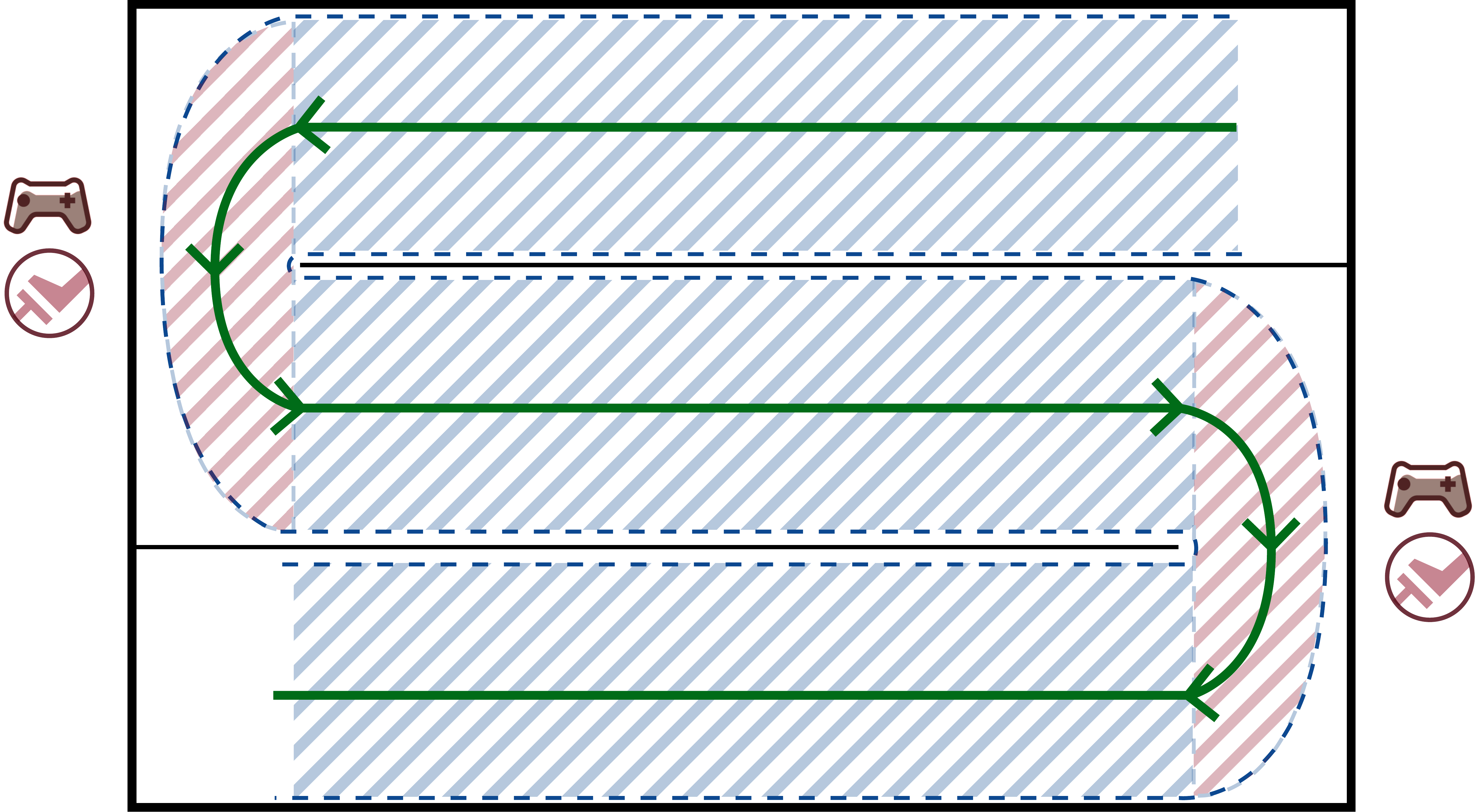}
        \label{fig:refineconcept1}
    }
    \subfigure[Narrow passage]{
        \includegraphics[width=0.4\linewidth]{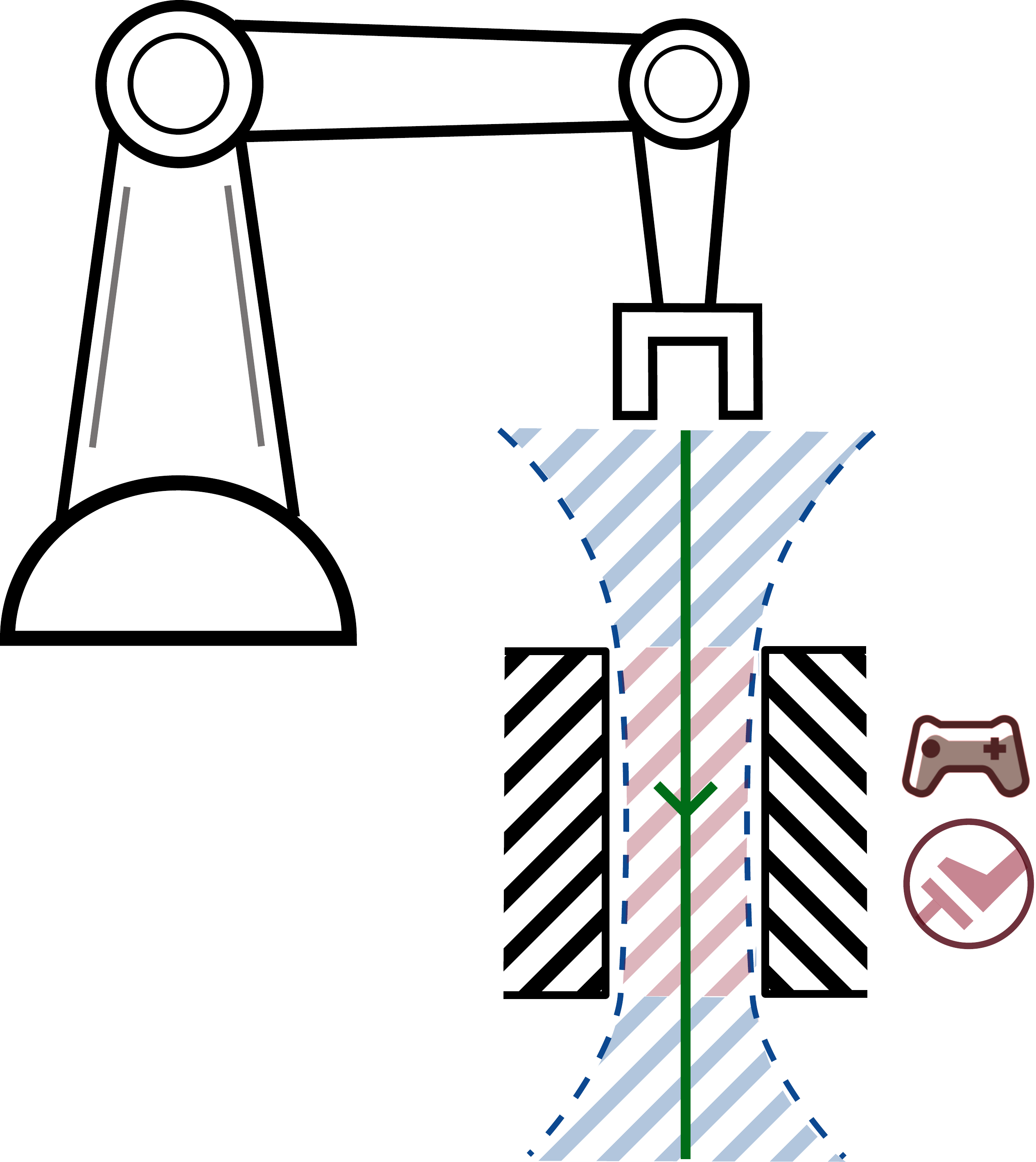}
        \label{fig:refineconcept2}
    }
    \caption{Illustrative examples to clarify the proposed refinement concept. The shaded regions represent the tolerance bounds. Red shaded areas indicate zones where the robot is intuitively expected to slow down based on teleoperated brake commands from the human operator. These regions require narrower tolerance bounds compared to other parts of the trajectory, represented by blue-shaded areas.}
    \label{fig:refineconcept}
\end{figure}

To clarify the concept of \textbf{OB2}, we present two illustrative examples in Fig. \ref{fig:refineconcept}. In the first example, depicted in Fig. \ref{fig:refineconcept1}, imagine the robot's end-effector navigating through a maze along the green line at maximum speed. The human operator can teleoperate the system to slow down the robot whenever necessary. Naturally, the robot's behavior is more critical at the maze's turning points, where it is more reliable if the robot reduces its speed. These turning points coincide with regions where the tolerance bounds must be narrower to avoid colliding with the maze walls.

Similarly, in Fig. \ref{fig:refineconcept2}, when the robot encounters a narrow passage, it is intuitive to slow down to ensure reliable execution. This passage also requires narrower tolerance bounds. Based on this conceptual analogy, our refinement objective in \textbf{OB2} is to enable the human operator to decrease the robot's speed to ensure reliable execution. Simultaneously, meaningful tolerance bounds are determined based on where a reduction in execution speed was commanded.

\subsection{Preliminary Concepts}
\subsubsection{Dynamic Movement Primitive}

The formulation of DMP  consists of two main components: a transformation system and a canonical system \cite{pastor2009learning, saveriano2023dynamic}. The transformation system governs the shape of the trajectory. It is described by a second-order differential equation resembling a damped spring-mass system with an additional forcing term to encode the desired movement shape:
\begin{align}
    &\tau \dot{z} = K(g - y) - Dz - K(g-y_0)x + Kf(x) \label{eq:dmp} \\
    &\tau \dot{y} = z, \label{eq:dmp2}
\end{align}
where $y$ and $z$ are the system's position and velocity, respectively. $\tau$ is a temporal scaling vector. $K$ and $D$ are positive constants that determine the system's stiffness and damping, respectively. $y_0$ is the start and $g$ is the goal states. $f(x)$ is a nonlinear forcing term that shapes the trajectory. $x$ is the phase variable from the canonical system. The forcing term $f(x)$ is typically represented as a weighted sum of basis functions as:
\begin{align}
    f(x) = \frac{\sum_{\mathfrak{i}=1}^{N} \psi_\mathfrak{i}(x) \mathrm{w}_\mathfrak{i} x}{\sum_{\mathfrak{i}=1}^{N} \psi_\mathfrak{i}(x)}, \label{eq:f(x)}
\end{align}
where $\psi_\mathfrak{i}(x)$ are Gaussian basis functions defined as
\begin{align}
    \psi_\mathfrak{i}(x) = \exp(-h_\mathfrak{i} (x - c_\mathfrak{i})^2).
\end{align}
$\mathrm{w}_\mathfrak{i}$ are the weights and $c_\mathfrak{i}$ and $h_\mathfrak{i}$ are the centers and widths of the basis functions, respectively.

The canonical system drives the phase variable $x$, which monotonically decreases from 1 to 0 during the course of the movement. It is governed by the following first-order differential equation:
\begin{align}
    \tau \dot{x} = -\alpha_x x,
\end{align}
where $\alpha_x$ is a constant that controls the speed of phase progression. 



\subsubsection{B-Spline}

B-Splines, or Basis Splines, are a family of piecewise-defined polynomials used in robotics to represent trajectories. B-Splines provide a powerful way to model smooth curves that can be easily manipulated by controlling a set of control points \cite{de1972calculating}. Their piecewise polynomial nature ensures smooth motion paths, making them highly suitable for complex motion planning.

The basis of B-Splines is the set of basis functions, which are defined recursively. The B-Spline basis functions of degree $k$, denoted $N_{{\hat{i}},k}(s)$, are defined over a non-decreasing sequence of real numbers called the knot vector, $U = \{u_0, u_1, ..., u_{\hat{n}+k+1}\}$. The recursive definition of the basis functions is as follows:

\begin{enumerate}
    \item Zero-degree (piecewise constant) basis functions defined as:
    \begin{align}
        N_{{\hat{i}},0}(s) &= \begin{cases} 
        1 & \text{if } u_{\hat{i}} \leq s < u_{{\hat{i}}+1} \\
        0 & \text{otherwise}
        \end{cases}   \label{eq:basis1}     
    \end{align}
    \item Higher-degree basis functions defined as:
    \begin{align}
        N_{{\hat{i}},k}(s) &= \frac{s - u_{\hat{i}}}{u_{{\hat{i}}+k} - u_{\hat{i}}} N_{{\hat{i}},k-1}(s) \nonumber \\
        &+ \frac{u_{{\hat{i}+k+1}} - s}{u_{{\hat{i}+k+1}} - u_{{\hat{i}+1}}} N_{{\hat{i}+1},k-1}(s). \label{eq:basis2}
    \end{align}
    where $N_{{\hat{i}},k-1}(s)$ and $N_{{\hat{i}}+1,k-1}(s)$ are the basis functions of degree $k-1$.
\end{enumerate}

A B-Spline curve of degree $k$ is defined as a linear combination of these basis functions. Given a set of control points $\{\bm{P}_0, \bm{P}_1, ..., \bm{P}_{\hat{n}}\}$, the B-Spline curve $\bm{\xi}(s)$ is defined by:
\begin{align}
\bm{\xi}(s) = \sum_{{\hat{i}}=0}^{\hat{n}} N_{{\hat{i}},k}(s) \bm{P}_{\hat{i}}, \label{eq:bspline}
\end{align}
where $\bm{P}_{\hat{i}}$ are the control points and $N_{{\hat{i}},k}(t)$ are the B-Spline basis functions.  




\subsection{Preliminary Functions}
The following functions are used in the optimization procedure:

\subsubsection{QuatDiff Function}

The $\mathsf{QuatDiff}(.,.) \in [0,\pi]$ function is a conventional function derived based on the dot product of quaternions to compute the absolute angular difference between two quaternions \cite{mccarthy1990introduction}. It is mathematically defined as:
\begin{align}
   \mathsf{QuatDiff}({\bm\theta}_1,\bm{\theta}_2) = 2 \cdot \arccos(|\bm{\theta}_1 \cdot \bm{\theta}_2|). \label{eq:quatdiff}
\end{align}
where $\bm{\theta}_1$ and $\bm{\theta}_2$ are the quaternions representing the orientations, and the dot product $|\bm{\theta}_1 \cdot \bm{\theta}_2|$ provides the cosine of the angle between them. This formula effectively computes the smallest angle required to rotate from one orientation to the other. During the trajectory optimization, the function ensures that the end-effector's orientation remains within a specified tolerance of the desired orientation at each waypoint.

\subsubsection{Forward Kinematics Function}
The forward kinematic function, denoted as $\mathsf{F_K}$, is used to determine the position and orientation of a robot's end-effector based on its joint configurations. This function maps the joint configuration of $\bm{q}_f(t)$ to the end-effector's position $\bm{p}_f(t)$ and orientation $\bm{\theta}_f(t)$ in the Cartesian space. Mathematically, this is expressed as:
\begin{align}
(\bm{p}_f(t),\bm{\theta}_f(t)) &=  \mathsf{F_K}(\bm{q}_f(t)). \label{eq:fk}
\end{align}
Here, $\bm{p}_f(t) = [x_f(t),y_f(t),z_f(t)] \in \mathbb{R}^3$ is a vector representing the position coordinates, and $\bm{\theta}_f(t) \in \mathbb{S}^3$ represents the orientation in quaternion form.

\textbf{Remark}: As this work focuses on representing the robot trajectory via B-Splines, the span of the knot values is defined within the range $[0,1]$ which means that $s \in [0,1]$. Therefore, $\bm{\xi}(s)$ models the trajectory over the normalized time scale $s$. An additional parameter $T$ is associated with $\bm{\xi}(s)$ which maps $s$ into actual time $t$. Finally, $\bm{P}_{\hat{i}}$ and $T$ are the decision variables to be solved via optimization. Throughout this paper, we refer to $s$ as normalized time and refer to $t$ as actual/real time.

\section{DFL-TORO Methodology}
\label{sec:method}

To achieve \textbf{OB1} and \textbf{OB2}, we have compiled our proposed methods into a comprehensive demonstration framework, DFL-TORO. This section details the methods and approaches incorporated in the DFL-TORO workflow. We begin with an overview of the workflow, illustrating how DFL-TORO transforms noisy demonstrations into optimal ones while addressing \textbf{OB1} and \textbf{OB2}. Following this, we explore the technical details of the DFL-TORO modules specifically designed to meet these objectives.

\subsection{Overall Workflow}

\begin{figure}[t]
    \centering
    \includegraphics[width=\linewidth]{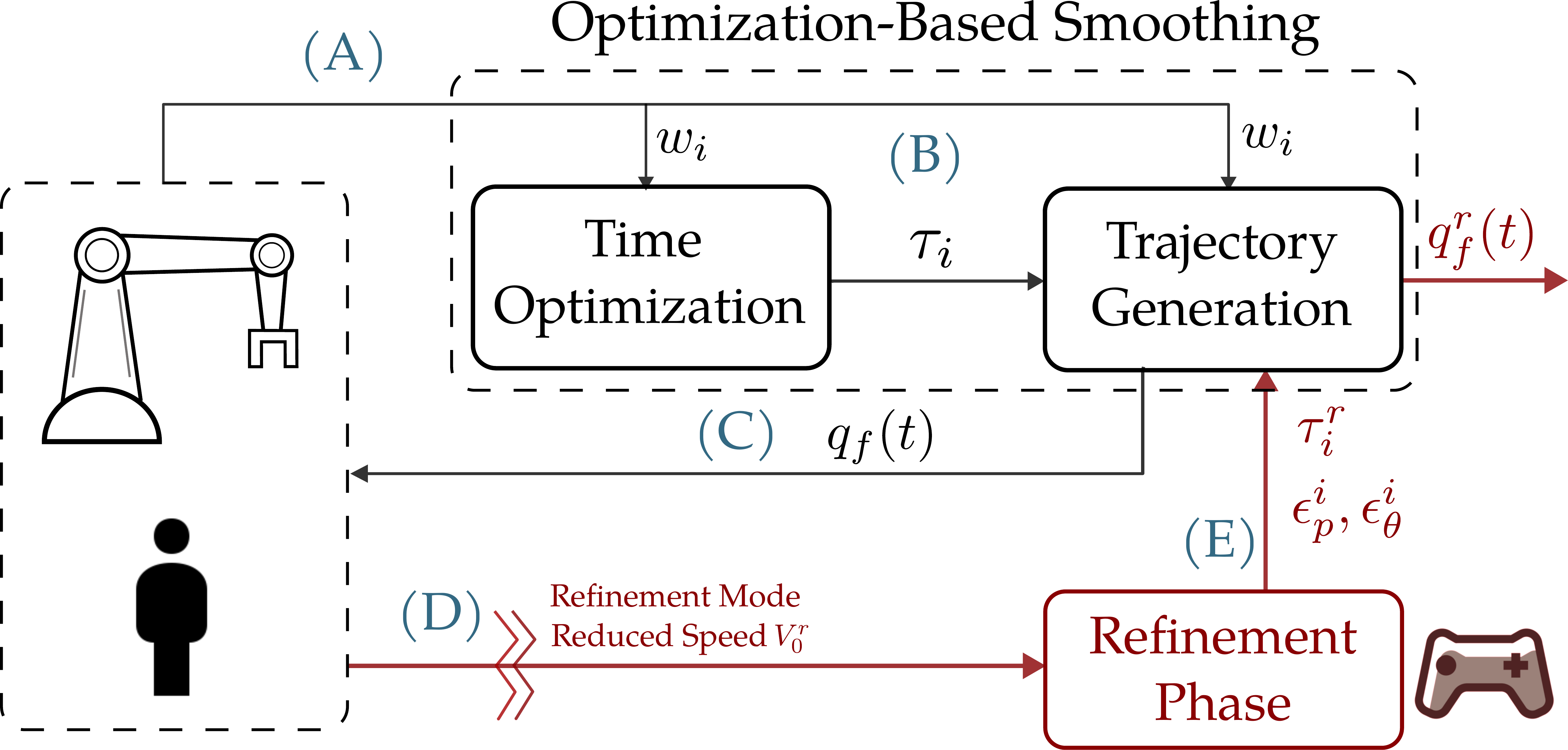}
    \caption{DFL-TORO workflow. Red arrows indicate interactive procedures with the human-robot in the loop.}
    \label{fig:overview}
    \vspace{-6mm}
\end{figure}

The overall workflow of DFL-TORO is illustrated in Fig. \ref{fig:overview}, with the steps labeled from (A) to (D). In this section, we provide a detailed walkthrough of each step, indicating where \textbf{OB1} and \textbf{OB2} are addressed. Moreover, the workflow of DFL-TORO involving these steps is provided in Algorithm \ref{alg:demo_framework}.

\begin{itemize}
    \item[\textbf{A)}] We start by recording $\bm{q}_o(t)$ from the inherently noisy human demonstration. The underlying path is then extracted and represented as $w_i$. The derivation of $w_i$ involves filtering the end-effector's position and orientation, ensuring the robot moves beyond a certain threshold before considering a new waypoint. This approach allows us to extract meaningful waypoints from the path.
    \item[\textbf{B)}] The ``Optimization-based Smoothing'' module takes $w_i$ as input and solves a comprehensive optimization problem to satisfy the requirements of \textbf{OB1}. The outcome of this module is $\bm{q}_f(t)$.
    \item[\textbf{C)}] The optimized trajectory $\bm{q}_f(t)$ is replayed on the robot under the supervision of the human teacher. In this step, the teacher analyses the execution performance of $\bm{q}_f(t)$ and checks whether the task is reliably performed under the new noise-free path and the optimized timing law. In case of inaccuracies or unreliable execution speed during the performance, the teacher proceeds with the refinement of $\bm{q}_f(t)$.
    \item[\textbf{D)}] The ``Refinement Phase'' allows the teacher to interactively slow down and correct the timing law. The robot replays the trajectory, allowing the human teacher to visually assess the execution speed. The teacher can pinpoint areas where the speed is either unreliable or unsafe and determine which segments require slowing down. Since the current trajectory $\bm{q}_f(t)$ is rapidly executed, it leaves no time for humans to observe the motion and provide feedback simultaneously. Therefore, the Refinement Phase operates at a reduced speed $\mathrm{v}_0^r \in \mathbb{R}_+$, giving humans enough reaction time to make changes.
    \item[\textbf{E)}] The Refinement Phase provides revised timing law for the trajectory, as well as the task tolerances $\bm{\epsilon}_p^i$ and $\epsilon_{\theta}^i$ extracted from human feedback. These updated values are subsequently fed to the ``Optimization-based Smoothing'' module to calculate a fine-tuned trajectory $\bm{q}_f^r(t)$ in accordance with the new timings and tolerances. This step is the last step of the workflow and the one that satisfies the requirements of \textbf{OB2}.
\end{itemize}

As mentioned, step (B) is where \textbf{OB1} is addressed via the Optimization-based Smoothing module, while \textbf{OB2} is addressed in steps (D) and (E) through the Refinement Phase. In the following sections, we elaborate on these modules and how our proposed methods tackle \textbf{OB1} and \textbf{OB2}.

\textbf{Remark}: Since $\bm{\epsilon}_p^i$ and $\epsilon_{\theta}^i$ are extracted in the Refinement Phase (Step (D)), the Trajectory Generation module utilizes a set of default tolerance values $\bm{\epsilon}_p^d$ and $\epsilon_{\theta}^d$ in step (B) to find $\bm{q}_f(t)$. After obtaining tolerance bounds, step (E) fine-tunes the trajectory to get $\bm{q}_f^r(t)$ based on the acquired tolerances, replacing the default values.

\subsection{Optimization-based Smoothing}
\label{sec:opt}

In this section, we explain our proposed method underlying the Optimization-based Smoothing, designed to transform $\bm{q}_o(t)$ into $\bm{q}_f(t)$. The objective is to find the best timing law given $w_i$ and regulate the jerk profile $\dddot{\bm{q}}_f(t)$, as an indication of noise. The constraints are the kinematic limits of the robot, as well as tolerance values binding $\bm{q}_f(t)$ to the waypoints $w_i$. Mathematically, let $t_i$ be the decision variable representing the time at which $\bm{q}_f(t)$ passes through $w_i$. Considering \eqref{eq:fk}, $\bm{q}_f(t)$ must satisfy the following constraint:
\begin{align}
- \bm{\epsilon}_p^i \leq \bm{p}_f(&t_i) - \bm{\mathfrak{p}}_{w_i} \leq \bm{\epsilon}_p^i, \label{eq:transtol} \\
 \mathsf{QuatDiff}&(\bm{\theta}_f(t_i), \bm{\theta}_{w_i}) \in [0, \epsilon_{\theta}^i], \label{eq:rottol}
\end{align}

Let $\bm{\xi}_f(s)$ with control points $\bm{P}_{\hat{i},f}$ and duration variable $T_f$ be the B-Spline model representing $\bm{q}_f(t)$. To enforce the constraints of \eqref{eq:transtol} and \eqref{eq:rottol}, we need to calculate the symbolic equation of $\bm{\xi}_f(\tau_i)$ where $\tau_i=\frac{t_i}{T_f}$, with respect to the decision variables $\bm{P}_{\hat{i},f}$ and $\tau_i$. According to \eqref{eq:bspline} this leads to:
\begin{align}
\bm{\xi}_f(\tau_i) = \sum_{{\hat{i}}=1}^{\hat{n}} N_{{\hat{i}},k}(\tau_i) \bm{P}_{\hat{i},f}. \label{eq:optbspline}
\end{align}
Following \eqref{eq:basis1} and \eqref{eq:basis2}, this requires calculation of $N_{\hat{i},0}(\tau_i)$ as:
\begin{align}
    N_{\hat{i},0}(\tau_i) &= \begin{cases} 
    1 & \text{if } u_{\hat{i}} \leq \tau_i < u_{\hat{i}+1} \\
    0 & \text{otherwise}
    \end{cases}   \label{eq:optbasis1}     
\end{align}
where $u_{\hat{i}}$ are the values from the knot vector. The issue here is that since $\tau_i$ is not known, \eqref{eq:optbasis1} leads to a highly non-linear behavior and does not result in a symbolic equation to be used for optimization. This is due to the fact that the timing of each waypoint across the B-Spline determines which basis function is used for the B-Spline equation, and since timing is a decision variable, it is not known between which knot values $w_i$ falls. Therefore, optimizing for $w_i$ and $t_i$ simultaneously leads to an ill-defined and highly nonlinear optimization formulation.

To overcome this, we approach the problem by solving a two-stage optimization problem. The first stage, i.e. the Time Optimization module, considers $\tau_i$ as decision variables and solves for the timing law. Then, in the Trajectory Generation module, $w_i$ are treated as decision variables, and a comprehensive optimization problem is solved to eliminate the noise and regulate the jerk profile. In the following, we delve into the details of the Time Optimization and the Trajectory Generation modules.

\subsubsection{Time Optimization Module}

The objective of this optimizer is to find the minimum-time trajectory $\bm{q}_t(t)$ that strictly passes through all the waypoints $\bm{\mathfrak{q}}_{w_i}$. At this stage, we ignore constraints related to acceleration and jerk since the noise is still present in the path. Adding such constraints would prevent the optimizer from finding the ideal timings. B-Spline sub-trajectories $\bm{\xi}_j$ for $\forall j=1,\cdots,m-1$ are used to represent the trajectory between every two adjacent waypoints $w_j$ and $w_{j+1}$, with control points $\bm{P}_{\hat{i},j}$ and durations $T_j$, as shown in Fig. \ref{fig:bsplinefit:timeopt}. Then, the Time Optimization problem is formulated as: 
\begin{subequations}    
\begin{align}
 (\bm{P}^*_{{\hat{i}},j},T^*_j) &= \argminx{\bm{P}_{{\hat{i}},j},T_j} \sum_{j=1}^{m-1} T_j, \label{eq:timeopt0} \\
\bm{q}_j(t_j)&=\bm{\xi}_j(\frac{t_j}{T_j}), \label{eq:timeopt1}\\
\bm{\mathfrak{q}}_{min} &\leq \bm{q}_j(t_j) \leq \bm{\mathfrak{q}}_{max}, \label{eq:timeopt1.5} \\
\bm{\mathfrak{v}}_{min} &\leq \dot{\bm{q}}_j(t_j) \leq \bm{\mathfrak{v}}_{max}, \label{eq:timeopt2} \\
\bm{\xi}_j(0) &= \bm{\mathfrak{q}}_{w_j}, \label{eq:timeopt2.5} \\
\bm{\xi}_j(1) &= \bm{\mathfrak{q}}_{w_{j+1}}, \label{eq:timeopt3} \\
\dot{\bm{\xi}}_1(0) &= \dot{\bm{\xi}}_{m-1}(1) = 0, \label{eq:timeopt4}
\end{align}
\end{subequations}
for $\forall t_j \in [0, T_j]$. \eqref{eq:timeopt1} relates the normalized and actual trajectories. We set bounds $\bm{\mathfrak{q}}_{min}$, $\bm{\mathfrak{q}}_{max}$, $\bm{\mathfrak{v}}_{min}$ and $\bm{\mathfrak{v}}_{max}$ for joint position and velocity as \eqref{eq:timeopt1.5} and \eqref{eq:timeopt2}, respectively. Continuity constraints \eqref{eq:timeopt2.5} and \eqref{eq:timeopt3} are applied at the intersection of these sub-trajectories. \eqref{eq:timeopt4} enforces the trajectory to start and rest with zero velocity. $\bm{q}_t(t)$ is a compilation of all the sub-trajectories $\bm{q}_j$ together.

It is important to note that the result of this optimization is typically an infeasible trajectory, as the existing noise causes high acceleration and jerk values. \eqref{eq:timeopt0} determines the ideal timings $T^*_j$ to move through all waypoints with highest velocity, which provides normalized time $\tau_i$ for $\forall i=1,\cdots,m$ defined as
\begin{align} \label{eq:tau_i}
\tau_i &= \begin{cases} 
\frac{\sum_{j=1}^{i-1} T^*_j}{T^*_{m-1}} & i \geq 2 \\
0 & i=1
\end{cases}.
\end{align}
This sets the groundwork for the Trajectory Generation module, which addresses the full optimization problem.

\subsubsection{Trajectory Generation Module}
\begin{figure}[htbp]
    \centering
    \subfigure[Time Optimization]{
        \includegraphics[width=\linewidth]{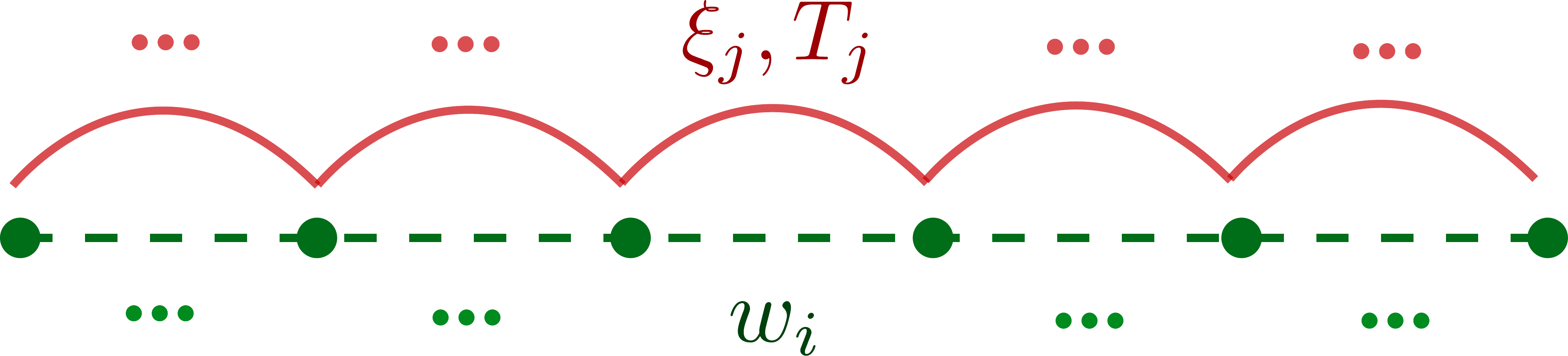}
        \label{fig:bsplinefit:timeopt}
    }
    \subfigure[Trajectory Generation]{
        \includegraphics[width=\linewidth]{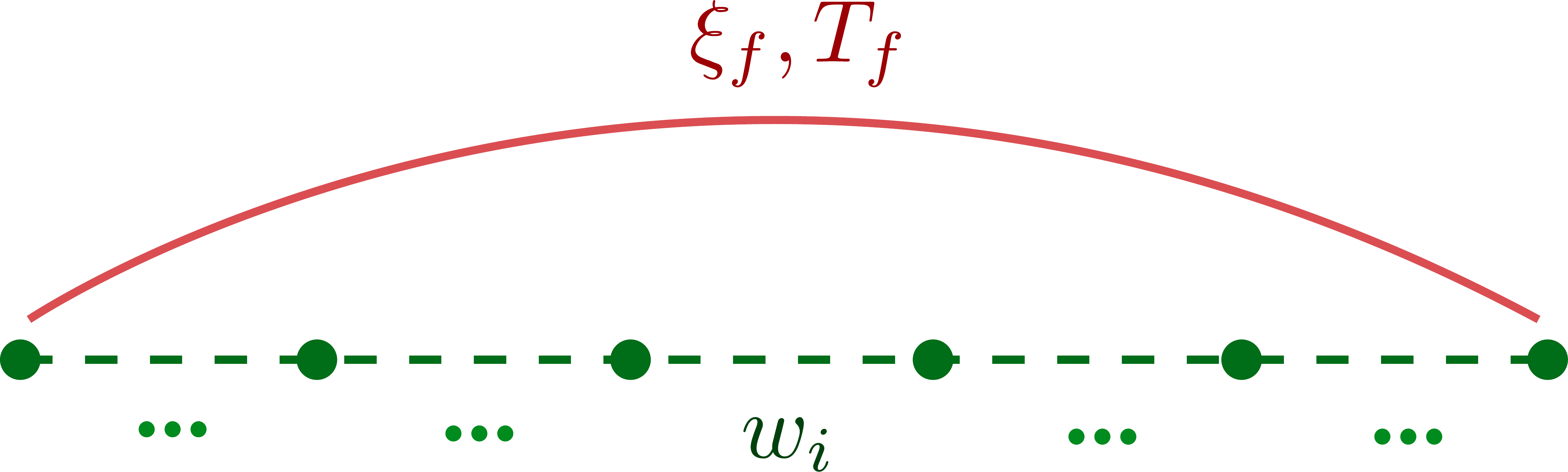}
        \label{fig:bsplinefit:trajopt}
    }
    \caption{Illustration of B-Spline fitting approach for each optimization step. The green lines represent the path and waypoints. Each red arc represents one B-Spline.}
    \label{fig:bsplinefit}
\end{figure}

Given $\tau_i$, we fit one B-Spline for $\bm{q}_f(t)$ across all the waypoints, denoted as $\bm{\xi}_f$ with control points $\bm{P}_{\hat{i},f}$ and duration variable $T_f$, as shown in Fig. \ref{fig:bsplinefit:trajopt}. We match the number of control points with the waypoints, i.e., $\hat{n}=m-1$, to give flexibility to the optimizer to locally adjust the trajectory around each waypoint. The cost function is formulated as:
\begin{align}
    J_f = \alpha T_f + \beta \int_{0}^{1}||\dddot{\bm{\xi}}_f||^2ds + \gamma  \sum_{i=1}^{m} ||\bm{P}_{\hat{i},f} -\bm{\mathfrak{q}}_{w_i}||^2, \label{eq:mainopt0}
\end{align}
where $\alpha,\beta$ and $\gamma$ are positive weights. The initial term $T_f$ involves the minimization of the overall execution time of $\bm{q}_f(t)$. Note that the Time Optimization module calculates the relative timing of waypoints in normalized time ($\tau_i$), while the actual timing law is optimized in this stage with respect to all the kinematic limits and meanwhile the noise removal. The term $\int_{0}^{1}||\dddot{\bm{\xi}}_f||^2ds$ attempts to minimize the normalized jerk profile, exploiting the tolerances of waypoints. This term is the key objective that pushes the optimization to eliminate the noise of the waypoints. The reason the normalized trajectory $\bm{\xi}_f$ is used in this term is that if $\bm{q_f}$ is used, the optimizer can simply increase $T_f$ to reduce jerk, instead of modifying the path. Therefore, using $\bm{\xi}_f$ isolates the jerk values from $T_f$ and forces the optimization to optimize the path. Finally, the term $\sum_{i=1}^{m} ||\bm{P}_{\hat{i},f} -\bm{\mathfrak{q}}_{w_i}||^2$ ensures that control points align closely with the original joint configurations at the waypoints. Given the robot manipulator's kinematic redundancy, an end-effector pose can correspond to multiple configurations. This term limits the robot's configuration null space, prompting the optimizer to remain near the initial configuration. The Trajectory Generation problem is formulated as follows: 
\begin{subequations}    
\begin{align}
\hspace{-3cm} (\bm{P}^*_{\hat{i},f},T^*_f) &= \argminx{\bm{P}_{\hat{i},f},T_f} 
\,J_f, \label{eq:mainopt0.5}\\
\hspace{-3cm} \bm{q}_f(t) &=\bm{\xi}_f(\frac{t}{T_f}), \label{eq:mainopt0.69} \\
\hspace{-3cm} \bm{\mathfrak{q}}_{min} &\leq \bm{q}_f(t) \leq \bm{\mathfrak{q}}_{max}, \label{eq:mainopt0.75}\\
\hspace{-3cm} \bm{\mathfrak{v}}_{min} &\leq \dot{\bm{q}}_f(t) \leq \bm{\mathfrak{v}}_{max}, \label{eq:mainopt1}\\
\hspace{-3cm} \bm{\mathfrak{a}}_{min} &\leq \ddot{\bm{q}}_f(t) \leq \bm{\mathfrak{a}}_{max}, \label{eq:mainopt1.5} \\
\hspace{-3cm} \bm{\mathfrak{j}}_{min} &\leq \dddot{\bm{q}}_f(t) \leq \bm{\mathfrak{j}}_{max},  \label{eq:mainopt2}\\
\hspace{-3cm} \bm{\xi}_f(0) &= \bm{\mathfrak{q}}_{w_1}, \label{eq:mainopt3} \\
\hspace{-3cm} \bm{\xi}_f(1) &= \bm{\mathfrak{q}}_{w_m}, \label{eq:mainopt4} \\
\hspace{-3cm}\dot{\bm{\xi}}_f(0) &= \dot{\bm{\xi}}_f(1) = 0, \label{eq:mainopt5} \\
\hspace{-3cm} (\bm{p}_f(t),\bm{\theta}_f(t)) &=  \mathsf{F_K}(\bm{q}_f(t)), \label{eq:mainopt5.5} \\
\hspace{-3cm} - \bm{\epsilon}_p^i \leq \bm{p}_f(&\tau_i T_f) - \bm{\mathfrak{p}}_{w_i} \leq \bm{\epsilon}_p^i, \label{eq:mainopt6} \\
\hspace{-1cm} \mathsf{QuatDiff}&(\bm{\theta}_f(\tau_i T_f), \bm{\theta}_{w_i}) \in [0, \epsilon_{\theta}^i], \label{eq:mainopt7}
\end{align}
\end{subequations}
for $\forall t \in [0, T_f]$.
We have introduced acceleration and jerk bounds  $\bm{\mathfrak{a}}_{min}$, $\bm{\mathfrak{a}}_{max}$, $\bm{\mathfrak{j}}_{min}$ and $\bm{\mathfrak{j}}_{max}$ in \eqref{eq:mainopt1.5} and \eqref{eq:mainopt2}, respectively, to ensure the trajectory's feasibility. Constraints \eqref{eq:mainopt3}-\eqref{eq:mainopt5} enforce the trajectory to start at $\bm{\mathfrak{q}}_{w_1}$ and rest at $\bm{\mathfrak{q}}_{w_m}$ with zero velocity.
\eqref{eq:mainopt5.5} represents the forward kinematic function $\mathsf{F_K}$, which is introduced in \eqref{eq:fk}. The goal is to limit the position and angle deviation of the end-effector at each waypoint $w_i$ to $\bm{\epsilon}_p^i$ and $\epsilon_{\theta}^i$, via \eqref{eq:mainopt6} and \eqref{eq:mainopt7}, respectively, at normalized time $\tau_i$. Solving \eqref{eq:mainopt0.5} yields $\bm{P}^*_{\hat{i},f}$, with which $\bm{q}_f(t)$ is computed using \eqref{eq:bspline} and \eqref{eq:mainopt0.69}.

Note that in the case of using default tolerance bounds (step (B)), the equations \eqref{eq:mainopt6} and \eqref{eq:mainopt7} are replaced with:
\begin{subequations}    
\begin{align}
    - \bm{\epsilon}_p^d \leq \bm{p}_f(&\tau_i T_f) - \bm{\mathfrak{p}}_{w_i} \leq \bm{\epsilon}_p^d, \label{eq:mainopt8} \\
    \mathsf{QuatDiff}&(\bm{\theta}_f(\tau_i T_f), \bm{\theta}_{w_i}) \in [0, \epsilon_{\theta}^d]. \label{eq:mainopt9}
\end{align}
\end{subequations}

\SetKwInput{KwOutput}{Output}

\begin{algorithm}
    \caption{DFL-TORO Architecture}
    \label{alg:demo_framework}
    
    \SetNlSty{textcolor}{rulecolor}{gray!20}
    
    \textbf{\underline{(A): Kinesthetic Guidance}} \\
    \KwIn{Noisy human demonstration $\bm{q}_o(t)$}
    \KwOutput{Path waypoints $w_i = \left( \bm{\mathfrak{q}}_{w_i}, \bm{\mathfrak{p}}_{w_i}, \bm{\theta}_{w_i} \right)$}
    \BlankLine

    \textbf{\underline{(B-1): Time Optimization}} \\
    \KwIn{$w_i$}
    Find ideal timings using \eqref{eq:timeopt0}-\eqref{eq:timeopt4} and \eqref{eq:tau_i} \\
    \KwOutput{Initial timings $\tau_i$}
    \BlankLine
    
    \textbf{\underline{(B-2): Trajectory Generation}} \\
    \KwIn{$w_i$, $\tau_i$, default tolerances $\bm{\epsilon}_p^d, \epsilon_{\theta}^d$}
    Solve for smooth trajectory using \eqref{eq:mainopt0.5}-\eqref{eq:mainopt5} and \eqref{eq:mainopt8}-\eqref{eq:mainopt9} \\
    \KwOutput{Trajectory $\bm{q}_f(t)$}
    \BlankLine

    \textbf{\underline{(C): Replay Trajectory under Supervision}} \\
    \KwIn{$\bm{q}_f(t)$}
    Human teacher assesses the performance of $\bm{q}_f(t)$
    
    \If{refinement is required}{
        Proceed to step (D)
    }
    \BlankLine

    \textbf{\underline{(D): Refinement Phase}} \\
    \KwIn{$\bm{q}_f(t)$, human command $C(t), \eta$}
    Initialize $v(0)=\mathrm{v}_0^r, s_r(0)=\mathrm{v}_0^r \cdot t$ \\
    \While{Replaying Trajectory}{
    Compute $s_r(t)$ using \eqref{eq:s_r} \\
    Send $\xi(s_r(t))$ to robot for interactive feedback
    }
    Compute new timings $\tau_i^r$ using \eqref{eq:tau_i_r}\\
    \For{each $w_i$}{
    $\bm{\epsilon}_p^i \gets \Gamma_p(\tau_i)$ using \eqref{eq:tol}\\
    $\epsilon_{\theta}^i \gets \Gamma_\theta(\tau_i)$ using \eqref{eq:tol}
    }
    \KwOutput{New timings $\tau_i^r$, new tolerances $\bm{\epsilon}_p^i, \epsilon_{\theta}^i$}
    \BlankLine

    \textbf{\underline{(E): Trajectory Fine-Tuning}} \\
    \KwIn{$w_i, \tau_i^r, \bm{\epsilon}_p^i, \epsilon_{\theta}^i$}
    Re-optimize $\bm{q}_f(t)$ via updated timings and tolerances using \eqref{eq:mainopt0.5}-\eqref{eq:mainopt7} \\
    \KwOutput{Optimal Demonstration $\bm{q}_f^r(t)$}
\end{algorithm}

\subsection{Refinement Phase}
\label{sec:refine}
Until this point, the one-shot demonstration process is finished, and $\bm{q}_f(t)$ is optimized to deliver the best timing law with default tolerances. Although the best timing law is naturally the most desirable in manufacturing and $\bm{q}_f(t)$ can be used as is, there might be cases where the human teacher prefers to slow down the execution in some parts or modify the default tolerances to imply a better accuracy and increase execution reliability and success rate. In this case, humans can optionally go through the Refinement Phase, where they can adjust the velocity of execution by slowing down $\bm{q}_f(t)$ and simultaneously teach the tolerance bounds.

\begin{figure}[htbp]
    \centering
    \includegraphics[width=0.8\linewidth]{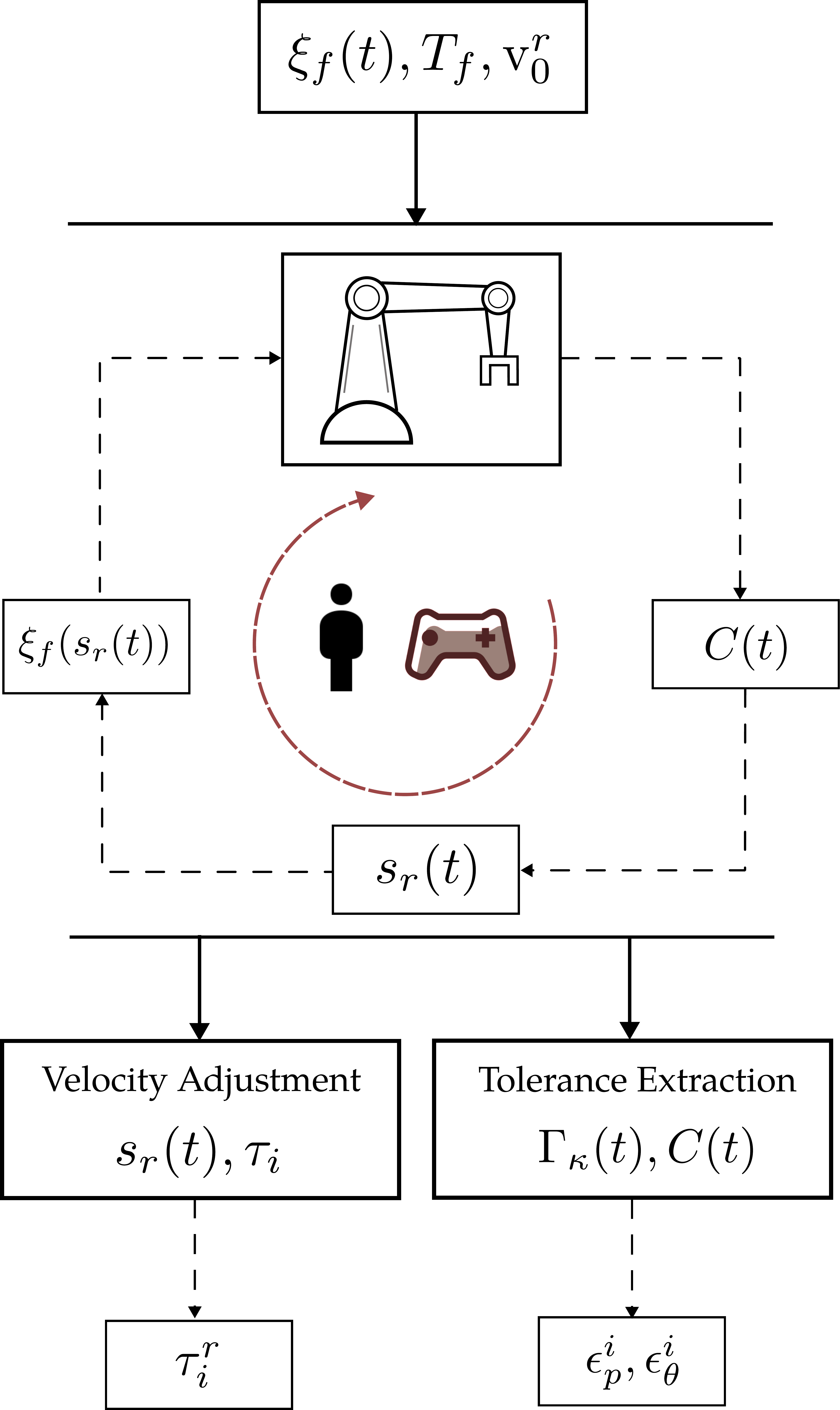}
    \caption{Visual diagram of the refinement phase. $C(t)$ and $s_r(t)$ are obtained during the interactive loop, leading to velocity adjustment and tolerance extraction.}
    \label{fig:refinediagram}
\end{figure}

The refinement process starts with the robot replaying $\bm{q}_f(t)$ under the supervision of the human teacher. the role of the human teacher is to monitor the motion and give real-time teleoperated feedback on every segment of $\bm{q}_f(t)$ that the robot executes. The human feedback is only in the form of a brake command denoted as $C(t) \in [-1,0]$, stating where the trajectory should slow down. At every time $t$ during the execution of $\bm{q}_f(t)$, $C(t)$ determines how much the velocity of execution of $\bm{q}_f(t)$ should be reduced, with $C(t)=-1$ indicating maximum reduction, and $C(t)=0$ indicating no reduction in the execution velocity. In other words, the human feedback acts as a brake pedal, decelerating $\bm{q}_f(t)$ in real time. In this interaction loop, $\bm{q}_f(t)$ executes with duration $T_f$, which is fast and makes it cognitively challenging for humans to rapidly observe, respond, and interact with the robot. Hence, $\bm{q}_f(t)$ is replayed with a reduced speed, giving enough time for the human teacher to observe the motion and provide interactive feedback in the loop.

the following will detail how we can adjust the velocity of $\bm{q}_f(t)$ and extract meaningful tolerance bounds $\bm{\epsilon}_p^i$ and $\epsilon_{\theta}^i$, leading to the refined trajectory $\bm{q}_f^r(t)$. A visual summary of the refinement process is shown in Fig. \ref{fig:refinediagram}, through which all the steps and the variables are shown through the workflow of the Refinement Phase.

\subsubsection{Velocity Adjustment}
\label{sec:vel}
To apply the deceleration effect of $C(t)$, we consider $\bm{q}_f(t)=\bm{\xi}_f(s(t))$, where $\bm{\xi}_f(s)$ is the trajectory in the normalized time in the range $[0,1]$, and $s(t)$ is the function that maps the real time $t$ to the normalized time $s$. Normally, $\bm{q}_f(t)$ is calculated by setting $s(t) = \frac{t}{T_f}$. However, here we reduce the speed by a factor of $\eta > 1$ by setting
\begin{align}
    s(t) = \mathrm{v}_0^r \cdot t, \label{eq:st}
\end{align}
where $\mathrm{v}_0^r = \frac{1}{\eta T_f}$. To apply the deceleration effect of $C(t)$, we manipulate the mapping of $s(t)$ in \eqref{eq:st}. Instead of directly altering the trajectory itself, we take $C(t)$ as a time deceleration factor, and slow down the progression of time in \eqref{eq:st}. This leads to a modified mapping $s_r(t)$. Using $C(t)$, this mapping is computed by integration the following equations throughout the interactive process:
\begin{align}
    \left\{ \begin{matrix}
        \dot v_r(t) = C(t) \\
        \dot s_r(t) = v_r(t)
    \end{matrix} \right., \quad
    \begin{matrix} 
    v_r(t) \in [\mathrm{v}_{min}^r,\mathrm{v}_0^r], \\
    s_r(t) \in [0,1]
    \end{matrix} \label{eq:s_r}
\end{align}
To prevent the robot from completely stopping, we bound $v_r(t)$ to a minimum execution speed $\mathrm{v}_{min}^r \in \mathbb{R}_+$. In this way, slowing down portions of $\bm{q}_f(t)$ is achieved via calculation and execution of $\bm{\xi}_f(s_r(t))$ throughout the interaction loop.

At the end of the interaction loop, by having $s_r(t)$, we can transform $\tau_i$ to the updated timing $\tau_i^r$ as
\begin{align} \label{eq:tau_i_r}
    \tau_i^r = \frac{s_r^{-1}(\tau_i)}{s_r^{-1}(\tau_m)},
\end{align}
where $s_r^{-1}$ is the inverse of $s_r$ obtained via interpolation. It is worth mentioning that refinement at the reduced speed $\mathrm{v}_0^r$ applies the same proportional adjustment on the real speed of execution, as $\tau_i^r$ are in normalized time.

\subsubsection{Tolerance Extraction}
\label{sec:tol}
To extract the tolerances, we pose a direct correlation between the value of $C(t)$ and the desired level of accuracy at a given point on the trajectory. This means that the more intensively the brake pedal is pressed, the greater the criticality of that trajectory portion, demanding increased precision. This is captured by introducing the functions  $\Gamma_p(t)$ and $\Gamma_\theta(t)$, defined as:
\begin{align}
    \Gamma_\kappa(t) = (\epsilon_\kappa^{max} - \epsilon_\kappa^{min})(1-(-C(t))^{\Bar{n}+\zeta}) + \epsilon_\kappa^{min}, \label{eq:tol}
\end{align}
where $\kappa \in \{p,\theta\}$. In \eqref{eq:tol}, $C(t) = 0$ is associated with maximum tolerances $\bm{\epsilon}_p^{max}$ and $\epsilon_\theta^{max}$, whereas a $C(t) = -1$ corresponds to the minimum tolerances $\bm{\epsilon}_p^{min}$ and $\epsilon_\theta^{min}$. The parameters $\zeta$ and $\Bar{n}$ determine the shape and curve of $\Gamma_\kappa$. By computing $\Gamma_\kappa(\tau_i)$ for each waypoint, we extract the requisite tolerances and pass it along with $\tau_i^r$ for re-optimization.

\section{Validation Experiments}
\label{sec:result}

In order to validate the effectiveness of our proposed demonstration framework, DFL-TORO, experiments were conducted on various robotic tasks by recording one-shot kinesthetic demonstrations. The purpose of these experiments is to validate the effectiveness of the DFL-TORO framework across different setups, including various robotic platforms and task environments. The experiments are designed to demonstrate the framework's ability to optimize trajectories' timing law, reduce noise, and regulate jerk in both laboratory and real-world manufacturing conditions. Our primary objectives are as follows:
\begin{itemize}
\item Validate DFL-TORO's performance concerning trajectory smoothness, execution time, and noise reduction.
\item Showcase the enhancement of LfD outcomes through optimal demonstrations using DFL-TORO, exemplified by a case study with DMPs.
\item Emphasize the practical applicability and generalizability of DFL-TORO by implementing it in real-world manufacturing operations.
\end{itemize}

\subsection{Validation Setup and Methodologies}

\subsubsection{Robotic Setup}
In this study, we utilize two robotic platforms for evaluation: the FR3 and the ABB Dual-Arm YuMi. FR3 robot, commonly known for its suitability in research settings, serves as the primary platform for controlled laboratory experiments. It is equipped with a 7 DoF arm, capable of precise motion control through position, velocity, and torque commands. The experiments carried out via FR3 are used for the performance analysis of DFL-TORO across its modules. In contrast, the ABB YuMi, a dual-arm collaborative robot, is deployed in a real manufacturing and production site. The ABB YuMi features a dual-arm configuration with 7-DoF per arm, designed for precise and collaborative tasks in industrial settings. The real-world deployment of ABB YuMi allows us to test the practical applicability and robustness of DFL-TORO in dynamic and complex industrial environments.

\subsubsection{Data Collection}
Demonstrations were recorded via kinesthetic guidance. These data were captured using a Robot Operating System (ROS) interface at a sampling frequency of 10 Hz. Waypoints $w_i$ are automatically selected based on an end-effector movement threshold, requiring a shift of at least 1 cm or 0.1 radians. For teleoperation, a wireless controller was employed to send continuous brake commands. The kinematic limits of FR3 and YuMi are collected from their respective manufacturer's datasheets.

\subsubsection{Software Configuration}
Regarding the software configuration of the robotic setups, the FR3 robot is operated using the ``libfranka'' ROS interface \cite{franka}. For capturing demonstrations, the robot is set to gravity-compensation mode, where torque commands are applied to the joints to compensate for the robot's weight, allowing the user to move the joints freely. For trajectory execution, the robot receives position commands at a control frequency of 1 kHz. 

The ABB YuMi robot is controlled using the ``abb robot driver'' software stack, developed by the ROS Industrial community \cite{abb_robot_driver}. This software facilitates interaction with ABB's internal controller software, enabling state monitoring of the robot and providing position or velocity commands. For capturing demonstration, YuMi operates in ``Lead-Through'' mode, a feature of ABB's internal controller similar to gravity-compensation mode, which allows for easy manual manipulation. During trajectory execution, the ROS interface controls YuMi via position commands, with a control frequency of 200 Hz.

To implement the optimization modules, we utilize the PyDrake toolbox \cite{drake}. The Time Optimization module is solved using the SNOPT solver, while the Trajectory Generation module is handled by the IPOPT solver. Additionally, the implementation of DMP was based on the ``dmpbbo'' toolbox as referenced in \cite{stulp2019dmpbbo}.

\subsubsection{Implementation details}
Regarding the implementation parameters, B-Splines of order $k=4$ are selected for smoothness up to the velocity level. The knot values $U$ are distributed in a clamped uniform manner \cite{de1972calculating} to create basis functions in \eqref{eq:bspline}. The weights in \eqref{eq:mainopt0} are set as $\alpha=1$, $\beta=0.04$, and $\gamma=1$. Finally, default tolerance values are chosen as $\bm{\epsilon}_p^d = [2,2,2]$ cm and $\epsilon_{\theta}^d = 0.1 $ radians.
In Refinement Phase, we select $\eta=5$ and $\mathrm{v}_{min}^r=0.2 \mathrm{v}_{min}^r$. Also, we choose $\bm{\epsilon}_p^{max} = [5,5,5]$ cm, $\bm{\epsilon}_p^{min}=[1,1,1]$ cm, $\epsilon_\theta^{max}=0.3$ radians, and $\epsilon_\theta^{min} = 0.1 $ radians. Furthermore, the parameters of $\Gamma_\kappa(t)$ are chosen as $\zeta=0.9$ and $\Bar{n}=1$. For the DMP case study, 15 basis functions were used in \eqref{eq:f(x)} to train the trajectories in the joint space. For the sake of comparison, DMP generalizations are calculated for the same start and goal configurations as the demonstration trajectory. 

\subsubsection{Evaluation Metrics}
The metrics used for evaluating and comparing the trajectories are the duration of the trajectory and the Maximum Absolute Normalized Jerk (MANJ). Since DFL-TORO optimizes the duration of demonstration trajectories, directly comparing the jerk profiles of these trajectories is not feasible, as a longer execution time naturally results in a lower jerk value. Therefore, we evaluate a normalized jerk value over normalized time $s \in [0,1]$ and use the maximum absolute value of each trajectory for comparison. For example, to calculate MANJ for $\bm{q}_f(t)$, we consider $\bm{\xi}_f(s)$. Let $\xi_{f,1}, \ldots, \xi_{f,n}$ be the trajectories for each individual joint. MANJ is then calculated as follows:
\begin{align}
    \text{MANJ} = \max_{l=1, \ldots, n} (\max_{s} |\dddot{\xi}_{f,l}|).
\end{align}
This metric effectively indicates the smoothness and noise level in the trajectories. The experimental results were visualized to demonstrate performance across different tasks, and a quantitative analysis was conducted to validate the improvements achieved by our method.

\subsubsection{Validation Technique}
The validation method for this study involves two main techniques. The first technique involves a performance evaluation of the Optimization-based Smoothing and Refinement Phase modules. This method assesses the improvements in time and jerk metrics by comparing the original demonstrations with those optimized by DFL-TORO. For this method, we utilize the FR3 experiments to showcase that the trajectories are free of noise, smooth, within the kinematic constraints, and accurate with respect to the tolerance bounds. The second validation is a case study using DFL-TORO with DMPs. This method highlights that incorporating DFL-TORO before using DMP significantly enhances task execution efficiency and precision. By first applying DFL-TORO to filter noise and optimize timing, the subsequent application of DMP yields more efficient and precise robotic movements compared to using DMP alone. For this technique, we utilize experiments from FR3 and YuMi to substantiate the improvements in both the quality and efficiency of robotic task execution when DFL-TORO is integrated into the demonstration process. It is worth noting that even though DMP is used in our case study, DFL-TORO applies to any other LfD algorithm and can serve as an intermediary layer to optimize demonstrations before feeding into the core algorithm.



\begin{figure}[htbp]
    \centering
    \subfigure[Kinesthetic teaching setup]{
        \centering
        \includegraphics[width=0.7\linewidth]{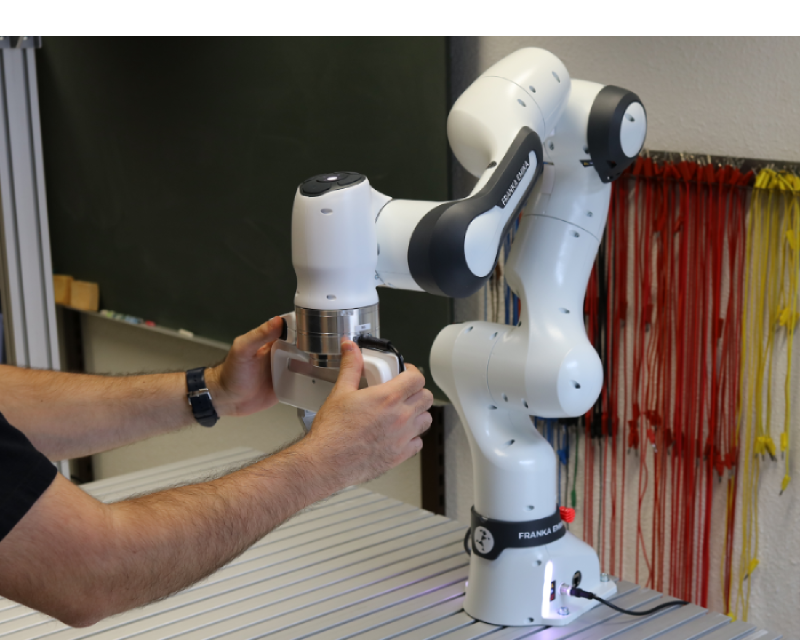}
        \label{fig:kinestheticsetup}
    }
    \subfigure[Teleoperation setup]{
        \centering
        \includegraphics[width=0.7\linewidth]{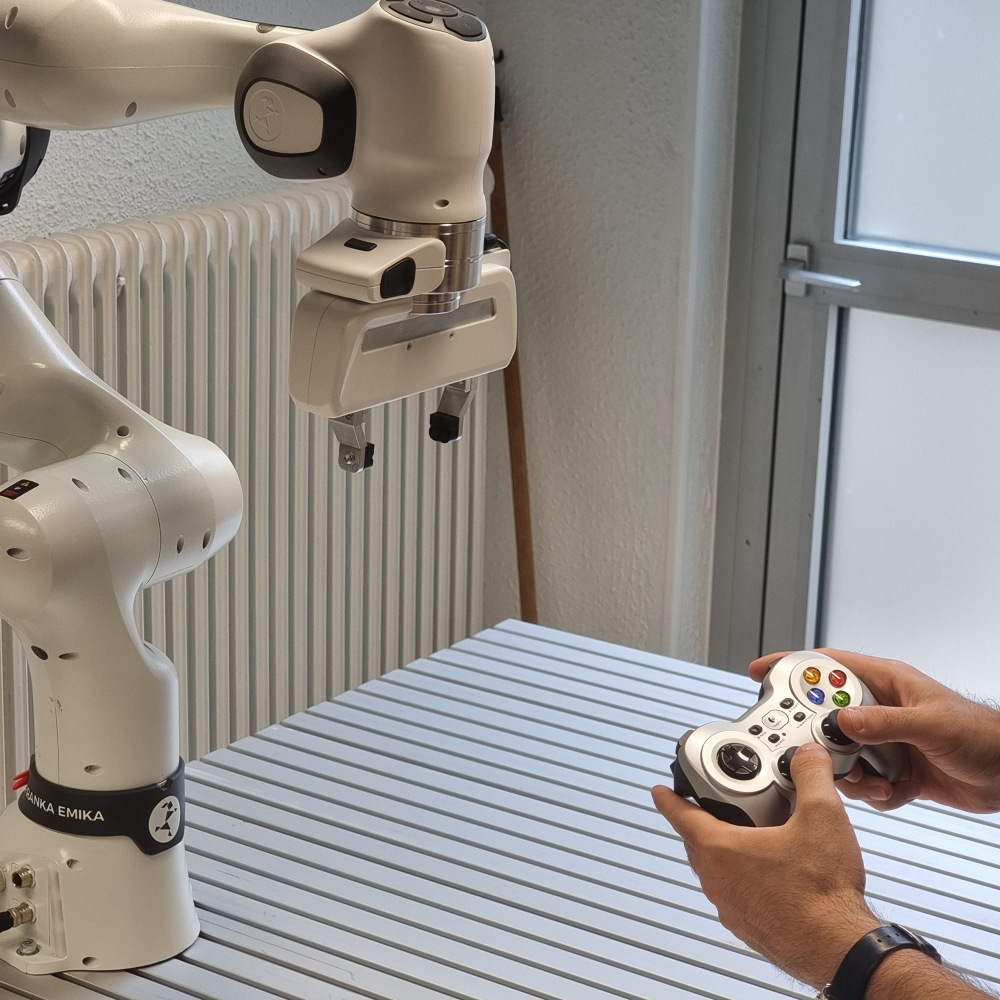}
        \label{fig:teleopsetup}
    }
    \caption{The FR3 experimental setup}
    \label{fig:exp}
\end{figure}

\subsection{Experiments with FR3}

\begin{figure*}[htbp]
    \centering
    \subfigure[RT1]{
        \includegraphics[width=0.18\linewidth]{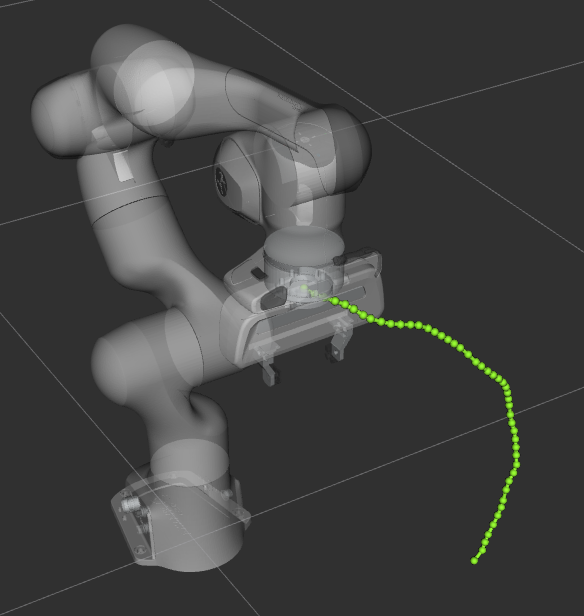}
    }
    \subfigure[RT2]{
        \includegraphics[width=0.18\linewidth]{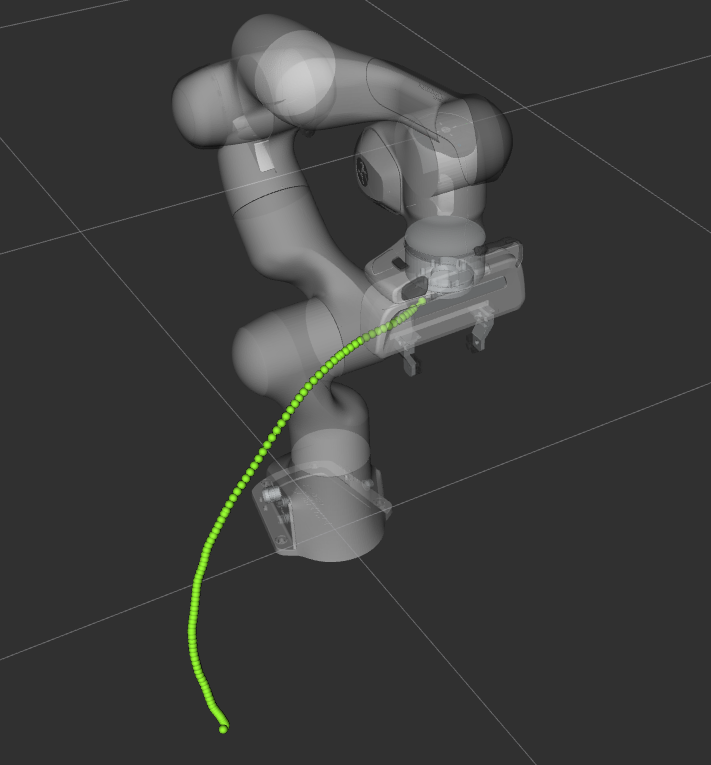}
    }
    \subfigure[RT3]{
        \includegraphics[width=0.18\linewidth]{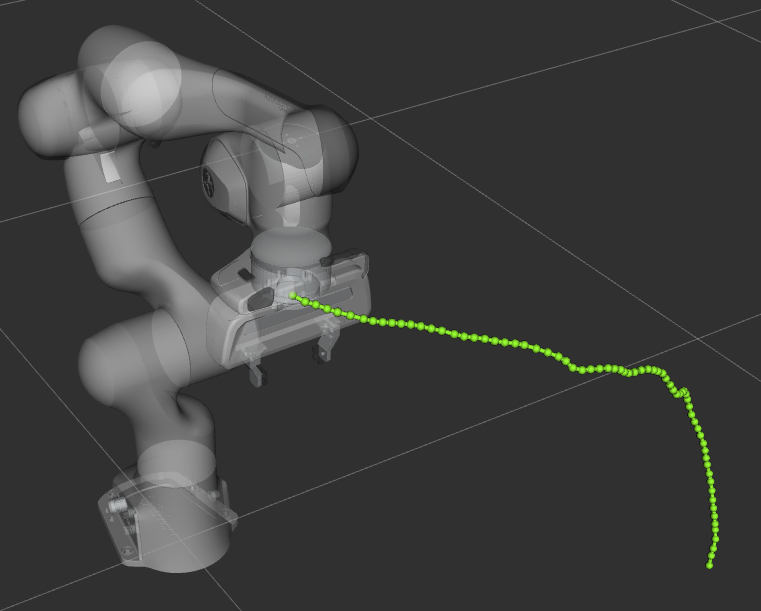}
    }
    \subfigure[RT4]{
        \includegraphics[width=0.18\linewidth]{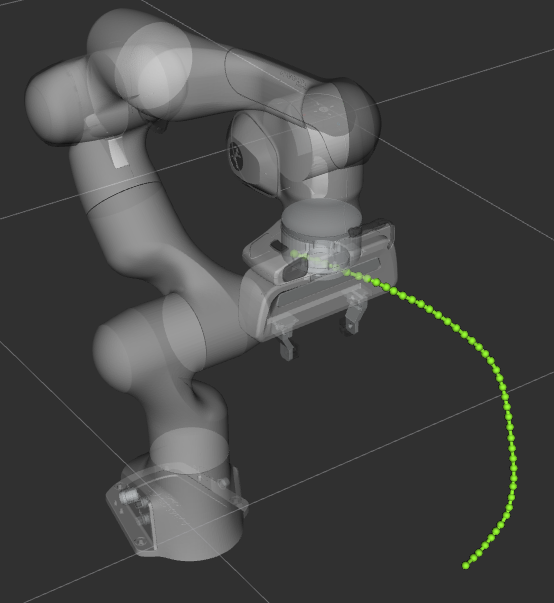}
    }
    \subfigure[RT5]{
        \includegraphics[width=0.18\linewidth]{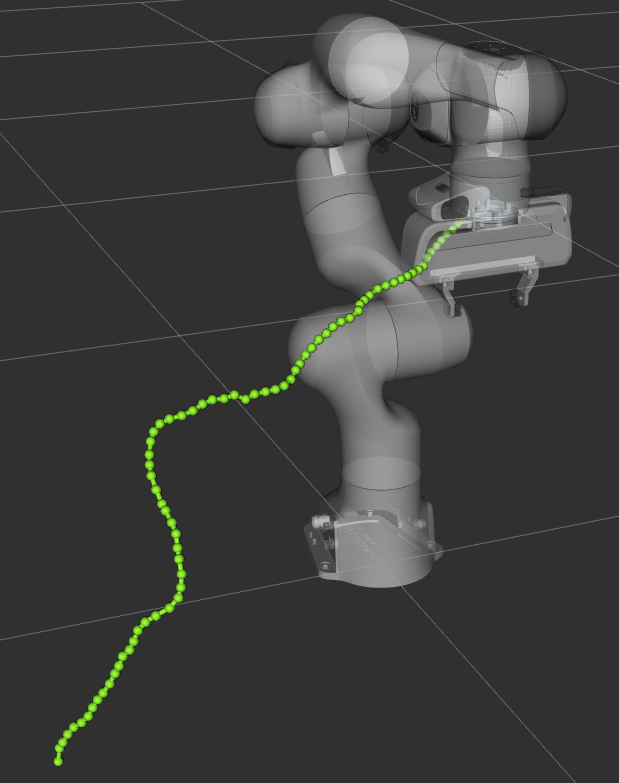}
    }
    \subfigure[MT1]{
        \includegraphics[width=0.18\linewidth]{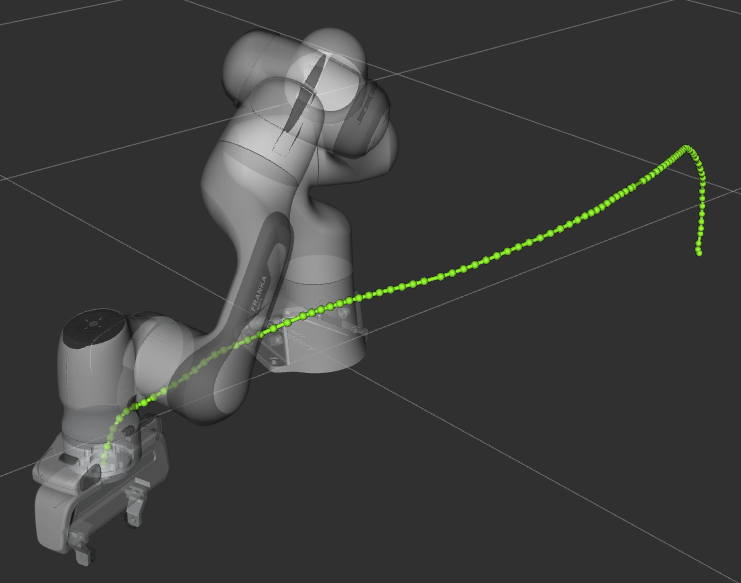}
    }
    \subfigure[MT2]{
        \includegraphics[width=0.18\linewidth]{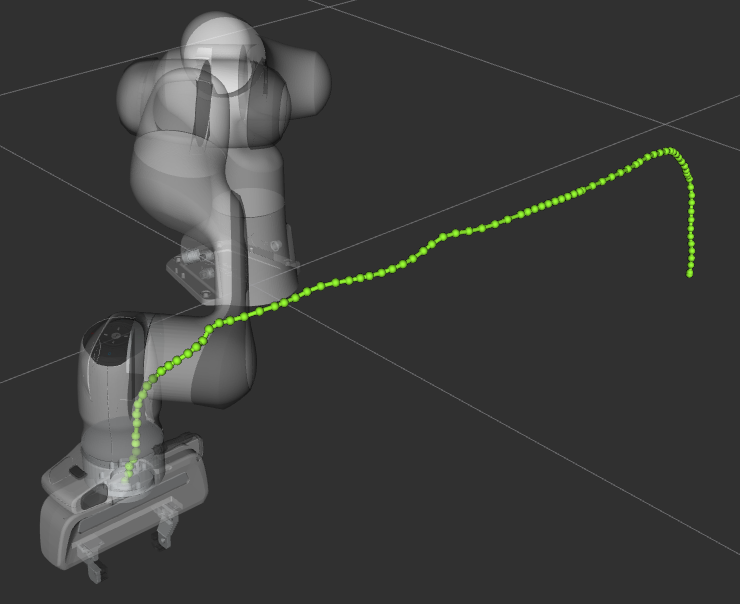}
    }
    \subfigure[MT3]{
        \includegraphics[width=0.18\linewidth]{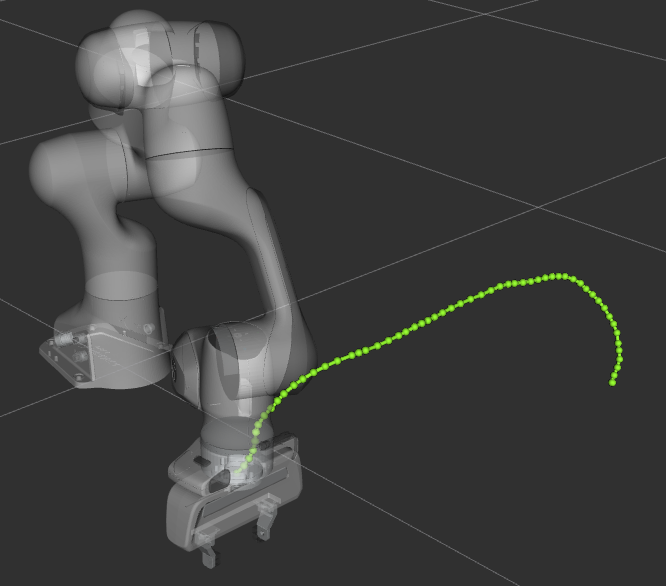}
    }
    \subfigure[MT4]{
        \includegraphics[width=0.18\linewidth]{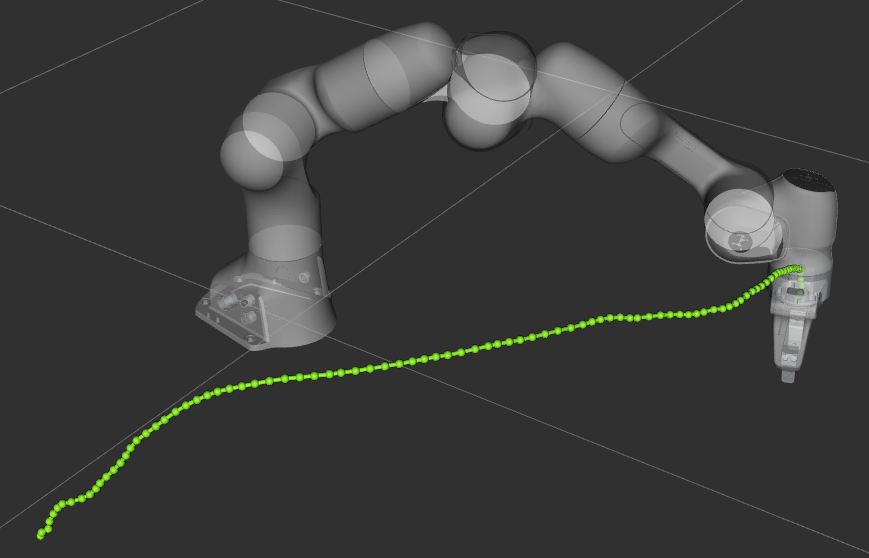}
    }
    \subfigure[MT5]{
        \includegraphics[width=0.18\linewidth]{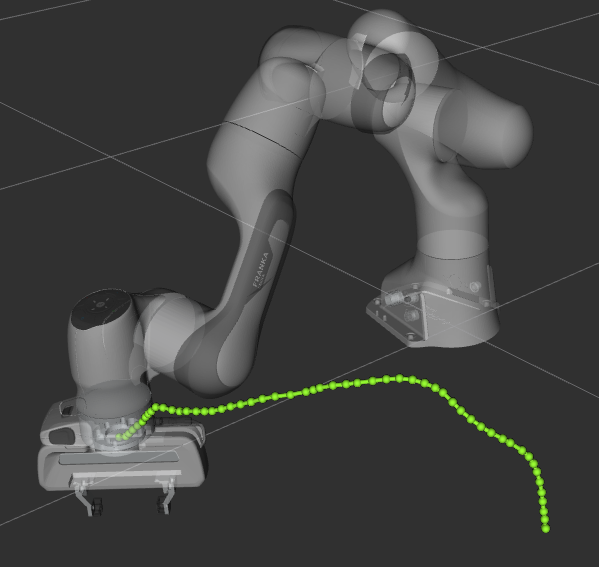}
    }
    \caption{Visualization of the demonstration trajectories recorded via FR3.}
    \label{fig:fr3demovisual}
\end{figure*}

In the experiments with FR3, A variety of motions are recorded. The experimental setup for FR3 is shown in Fig. \ref{fig:exp}. The demonstrations were recorded to cover typical robotic motions such as in pick-and-place operations, the Reaching motion to pick an object, and the moving motion to place it in another location. These tasks represent the majority of real-world applications. Overall, Five different reaching tasks (RT1-RT5) and five different moving tasks (MT1-MT5) are recorded, which is shown in Fig. \ref{fig:fr3demovisual}. While three tasks (RT1, RT2, MT1) are utilized for visualization purposes, the remaining tasks are included in a comprehensive analysis to demonstrate the robustness and effectiveness of the proposed methodology.

\begin{figure}[h!]
    \centering
    \subfigure[Original Demonstration $\bm{q}_o(t)$.]{
        \centering
        \includegraphics[width=\linewidth]{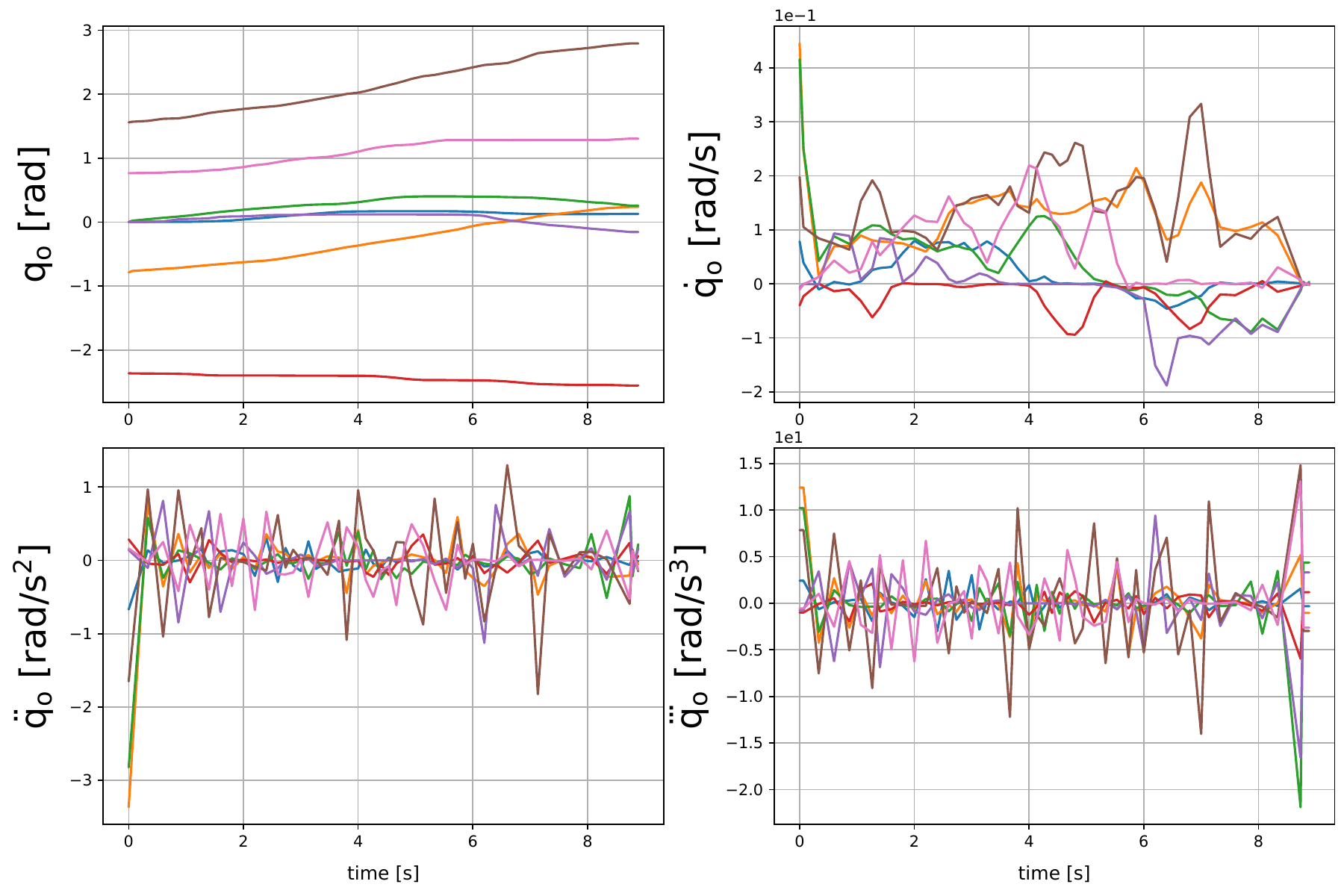}
        \label{fig:org}
    }
    
    \vspace{1em} 
    
    \subfigure[Time Optimization Output $\bm{q}_t(t)$.]{
        \centering
        \includegraphics[width=\linewidth]{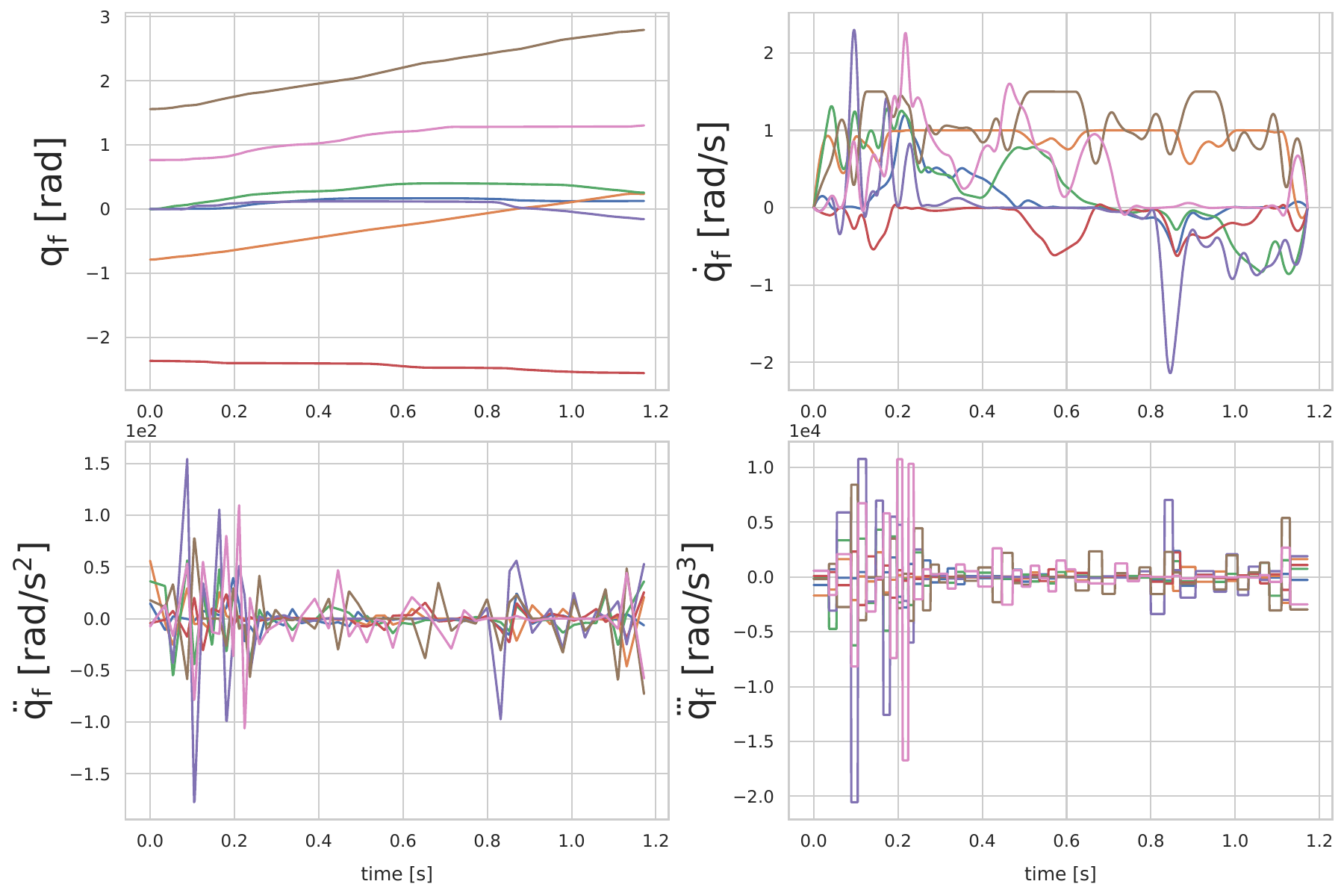}
        \label{fig:opt1}
    }
    
    \vspace{1em} 
    
    \subfigure[Trajectory Generation Output $\bm{q}_f(t)$.]{
        \centering
        \includegraphics[width=\linewidth]{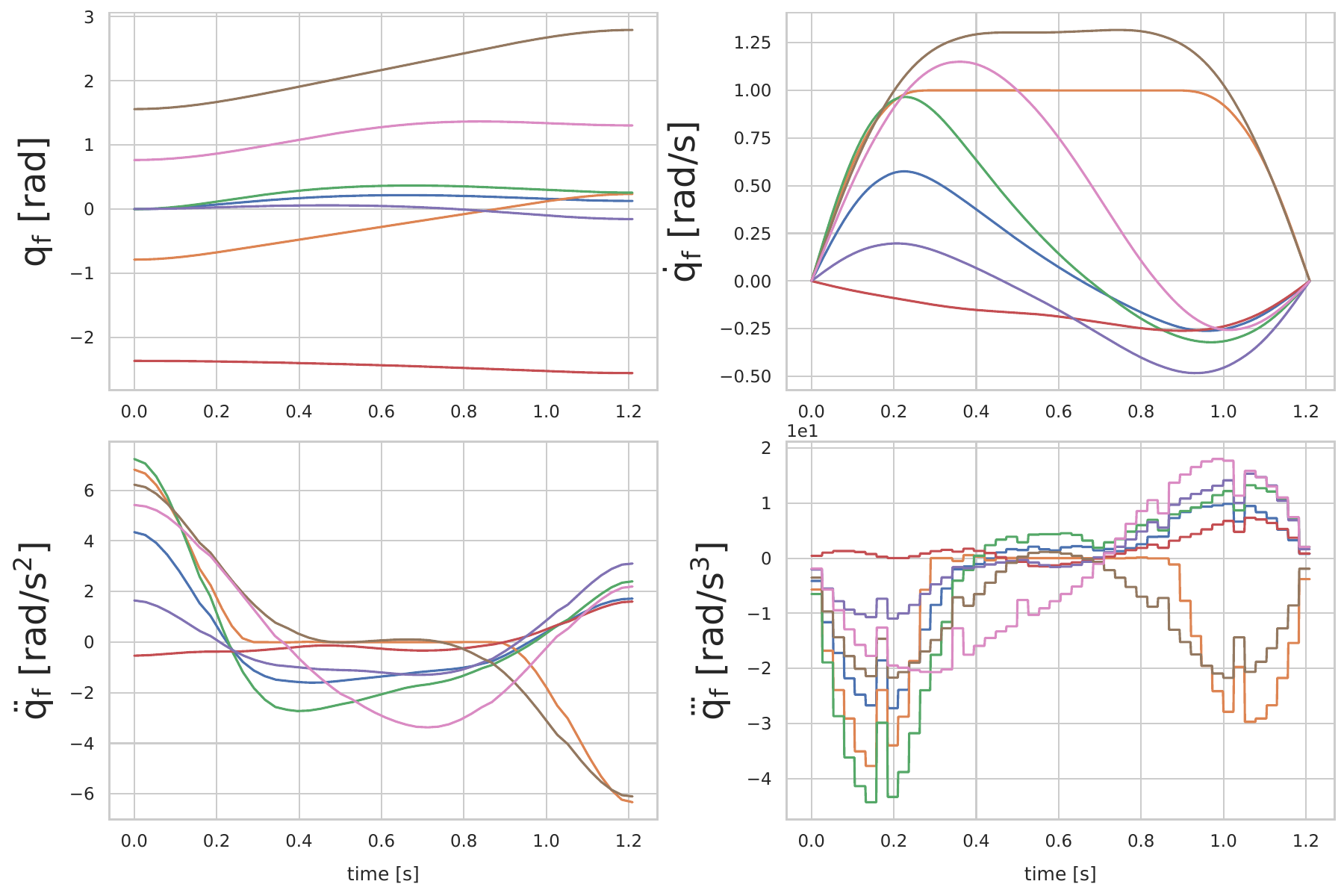}
        \label{fig:opt2}
    }
    
    \caption{Evolution of RT1 from original to smooth time-optimal joint trajectory. The motion profiles including position, velocity, acceleration, and jerk are shown for all 7 joint axes.}
    \label{fig:opt}
\end{figure}

\subsubsection{Performance of Optimization-based Smoothing}
To analyze the performance of the Optimization-based Smoothing module, we visualized the evolution of the RT1 trajectory. The original demonstration's joint trajectory, $\bm{q}_o(t)$, is shown in Fig. \ref{fig:org}. The velocity, acceleration, and jerk profiles were differentiated in a non-causal manner \cite{stulp2019dmpbbo}. Notably, the demonstration trajectory is slow with a duration of 8.88 seconds, while the presence of noise reflects itself in all the profiles, especially the jerk profile.


After applying the Time Optimization module, we obtain the trajectory $\bm{q}_t(t)$, with the relative timing law $\tau_i$ extracted using \eqref{eq:tau_i}. As shown in Fig. \ref{fig:opt1}, while this optimization reduces the trajectory duration to 1.17 seconds, the noise remains present, and both acceleration and jerk spikes reach extremely high values, up to approximately 150 $rad/s^2$ for acceleration and 20000 $rad/s^3$ for jerk. These values are unrealistic and significantly exceed the kinematic limits, as only the timing law $\tau_i$ is modified without altering the underlying path of the demonstration.


The final trajectory, $\bm{q}_f(t)$, is generated by applying $\tau_i$ along with $\bm{\epsilon}_p^d$ and $\epsilon^d_\theta$ in the Trajectory Generation module. This resulted in a smooth, noise-free, and time-optimal trajectory, depicted in Fig. \ref{fig:opt2}. The profiles clearly show the removal of path noise, adherence to kinematic limits, and a peak jerk value reduced to around 400 $rad/s^3$. The trajectory duration was adjusted to 1.21 seconds, slightly longer than the result from the Time Optimization module. This adjustment is due to the extraction and use of the normalized timing law $\tau_i \in [0,1]$ in the Trajectory Generation module and the inclusion of the term $T_f$ in the second optimization, which finds the optimal timing law while regulating jerk.


In addition to the visual analysis, we compared the original trajectory $\bm{q}_o(t)$ and the optimized trajectory $\bm{q}_f(t)$ for RT1-RT5 and MT1-MT5 in Table \ref{tab:comparison} with respect to their execution time and MANJ.

\subsubsection{Refinement Phase Analysis}
To show the performance of the Refinement Phase in locally slowing down the timing of $\bm{q}_f(t)$ and capturing meaningful tolerance values, we use $\bm{q}_f(t)$ from RT1 and pass it through the Refinement Phase to slow down the trajectory toward its end. In this experiment, we acquire $C(t)$ through teleoperated feedback. It is expected that the Refinement Phase modifies the timing law of $\bm{q}_f(t)$ to slow it down proportional to the teleoperated brake command $C(t)$. Moreover, the extracted tolerance bounds are expected to be narrower towards the end of the motion, relative to the intensity $C(t)$. Besides lowering tolerances in these portions, we assign maximum tolerance when $C(t)=0$, providing more freedom for the Trajectory Generation module to adapt the motion.

\begin{figure}[htbp]
    \centering
    \includegraphics[width=\linewidth]{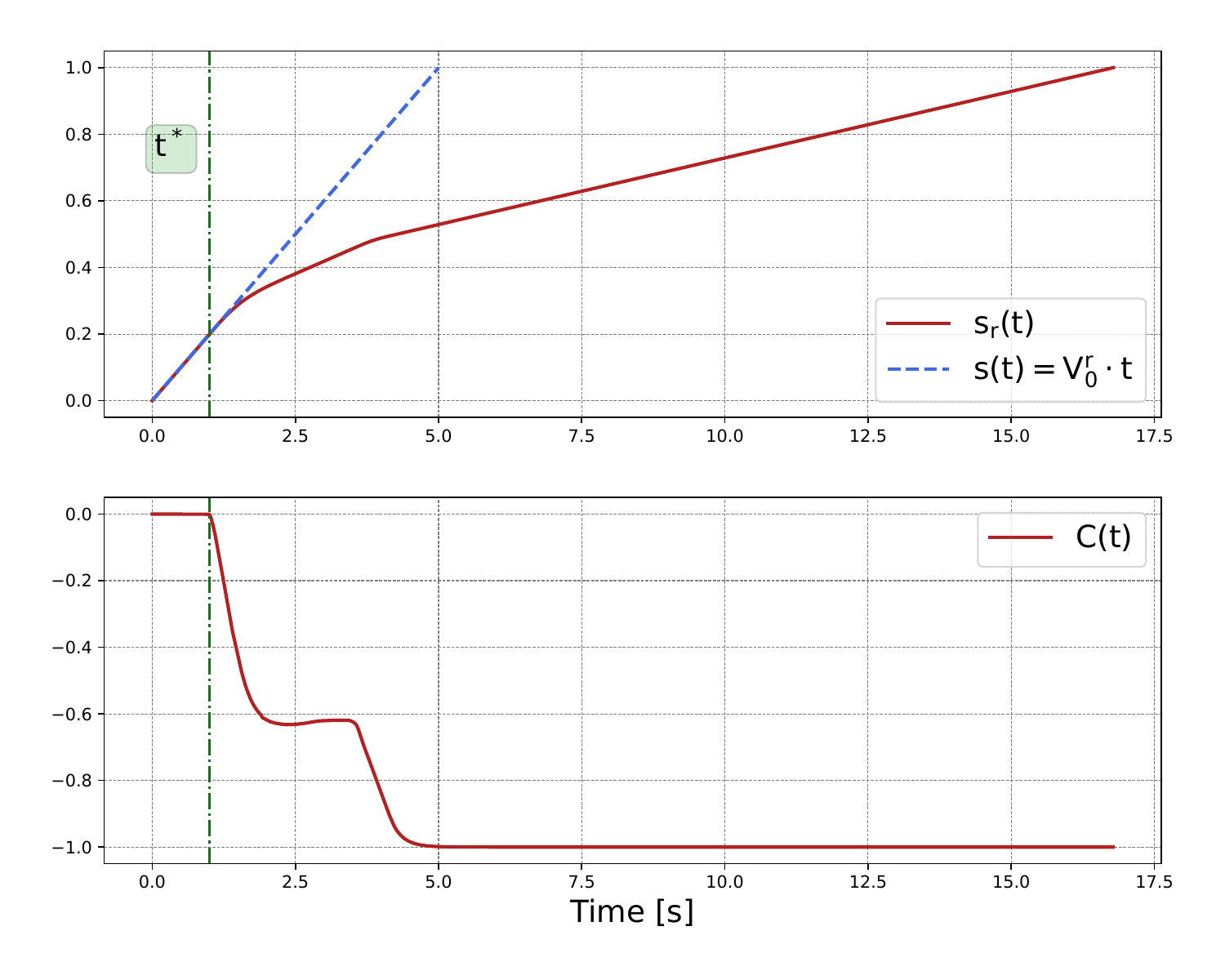}
    \caption{$s(t)$ and $s_r(t)$ during the Refinement Phase of RT1. $t^*$ marks the moment the teacher starts giving command $C(t)$.}
    \label{fig:srt}
\end{figure}

Given $C(t)$, the progression of $s(t)$ and $s_r(t)$ are calculated via \eqref{eq:st} and \eqref{eq:s_r} throughout the refinement and shown in Fig. \ref{fig:srt}. Initially, without any command, $s(t)$ and $s_r(t)$ overlap, leaving the timing law unchanged. Upon receiving the command at $t^*$, $s_r(t)$ progression changes to slow down the motion and thus extend execution time. This revised mapping leads to the derivation of $\tau_i^r$ using \eqref{eq:tau_i_r}. Also, we extract $\bm{\epsilon}_p^i$ and $\epsilon_\theta^i$ via \eqref{eq:tol}. 

\begin{figure}[htbp]
    \centering
    \subfigure[$\bm{p}_f^r(t)$ with tolerance bounds extracted based on $C(t)$]{
        \includegraphics[width=\linewidth]{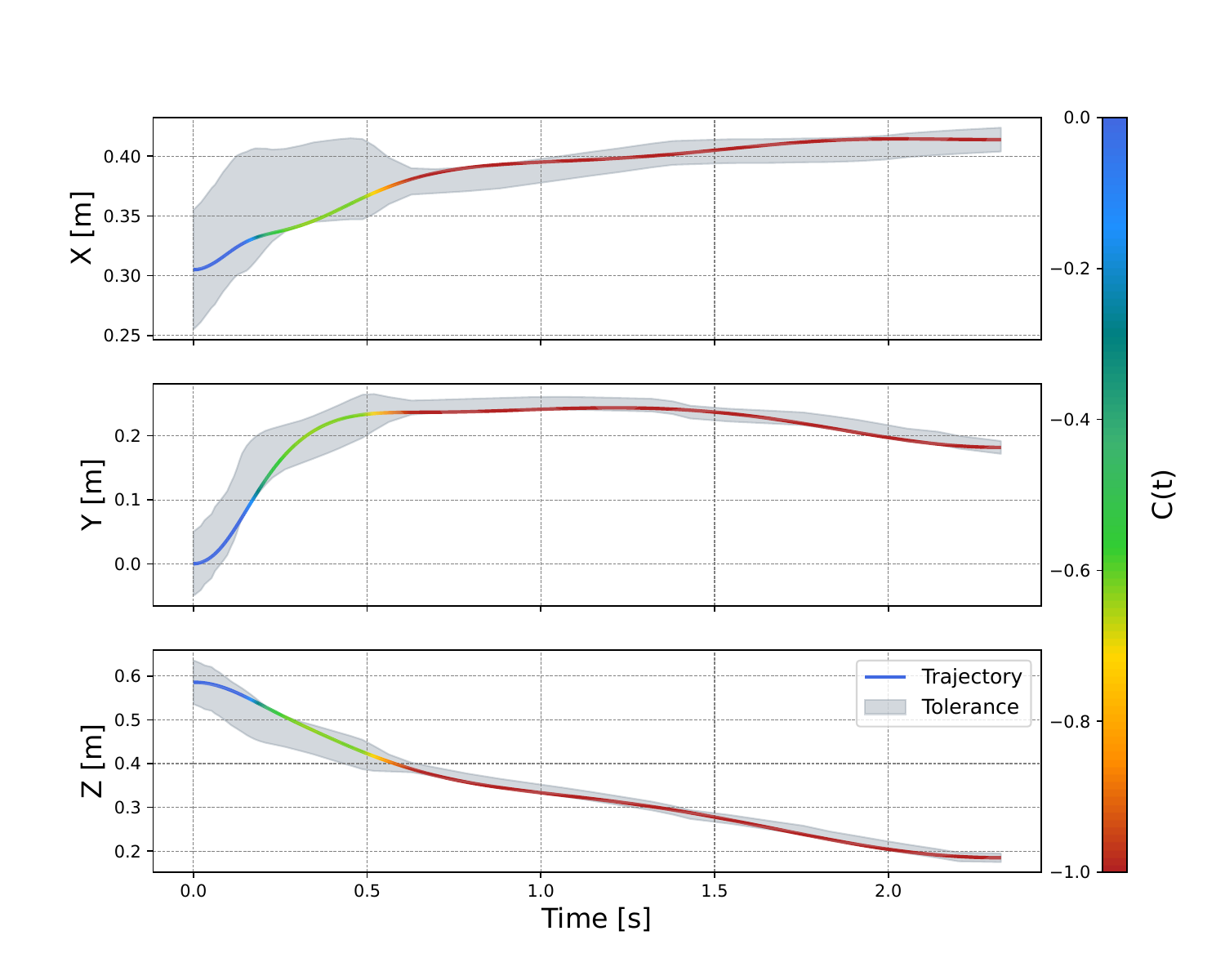}
        \label{fig:refined}
    }
    
    
    \subfigure[Waypoints timing law before and after refinement.]{
        \includegraphics[width=0.9\linewidth]{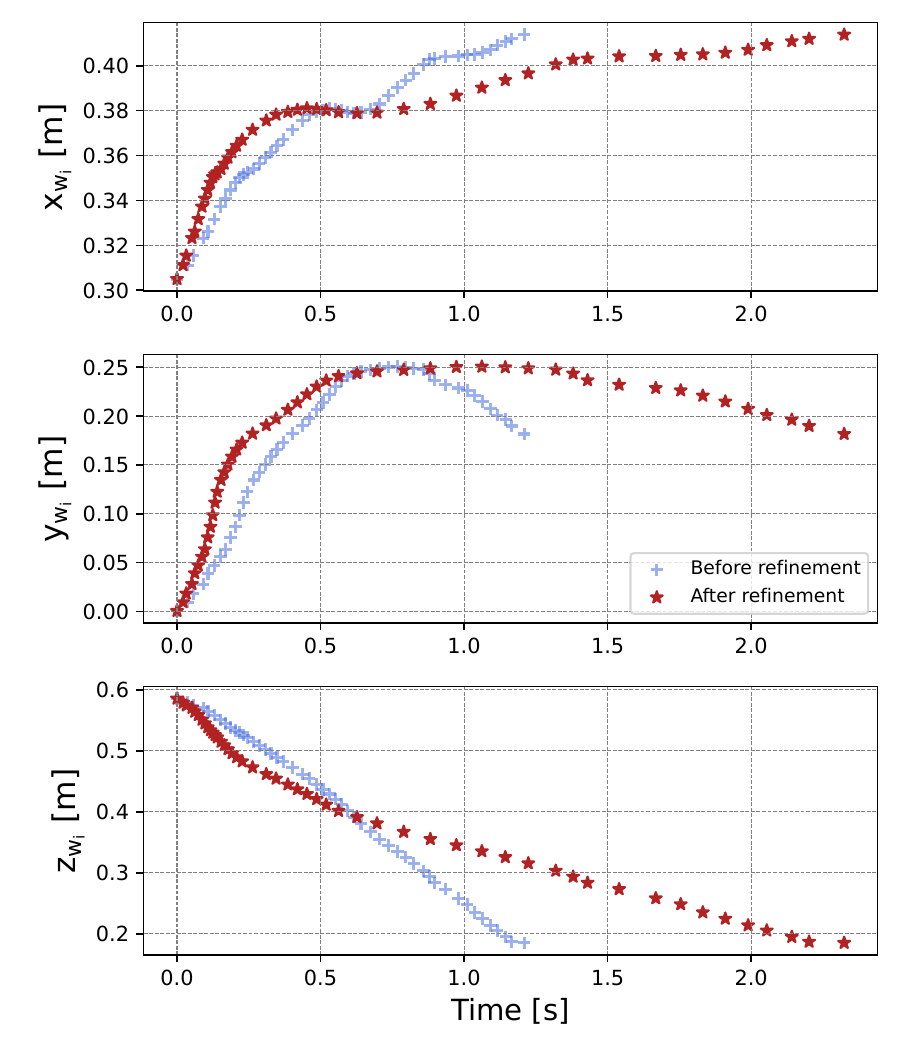}
        \label{fig:wps}
    }
    
    \caption{RT1 fine-tuning after the Refinement Phase. As the intensity of $C(t)$ increases, the tolerance bounds are narrowed, and the timing of waypoints is extended accordingly.}
    \label{fig:fine_tune}
\end{figure}

With the extracted information, The fine-tuned trajectory $\bm{p}_f^r(t) = [x_f^r(t), y_f^r(t), z_f^r(t)]^T$ is calculated and shown in Fig. \ref{fig:fine_tune}. In Fig. \ref{fig:refined}, The tolerance range is adjusted to be more precise towards the end of reaching, where slow speed was commanded. Moreover, to showcase the velocity adjustment effect on the trajectory's timing law, we present the distribution of $\bm{\mathfrak{p}}_{w_i}=[x_{w_i}, y_{w_i}, z_{w_i}]^T$ over the trajectory before and after refinement in Fig. \ref{fig:wps}. Evidently, the waypoints are stretched towards the end of the task. It is also noticeable that while the timing law of the beginning of the trajectory is left unchanged during refinement, the waypoint distribution does not overlap. This is due to the fact that since the tolerance values are increased from $\bm{\epsilon}_p^d$ to $\bm{\epsilon}_p^{max}$, as in \eqref{eq:tol}, the optimizer has managed to find a faster trajectory to traverse these waypoints.

\begin{figure}[htbp]
    \centering
    \includegraphics[width=\linewidth]{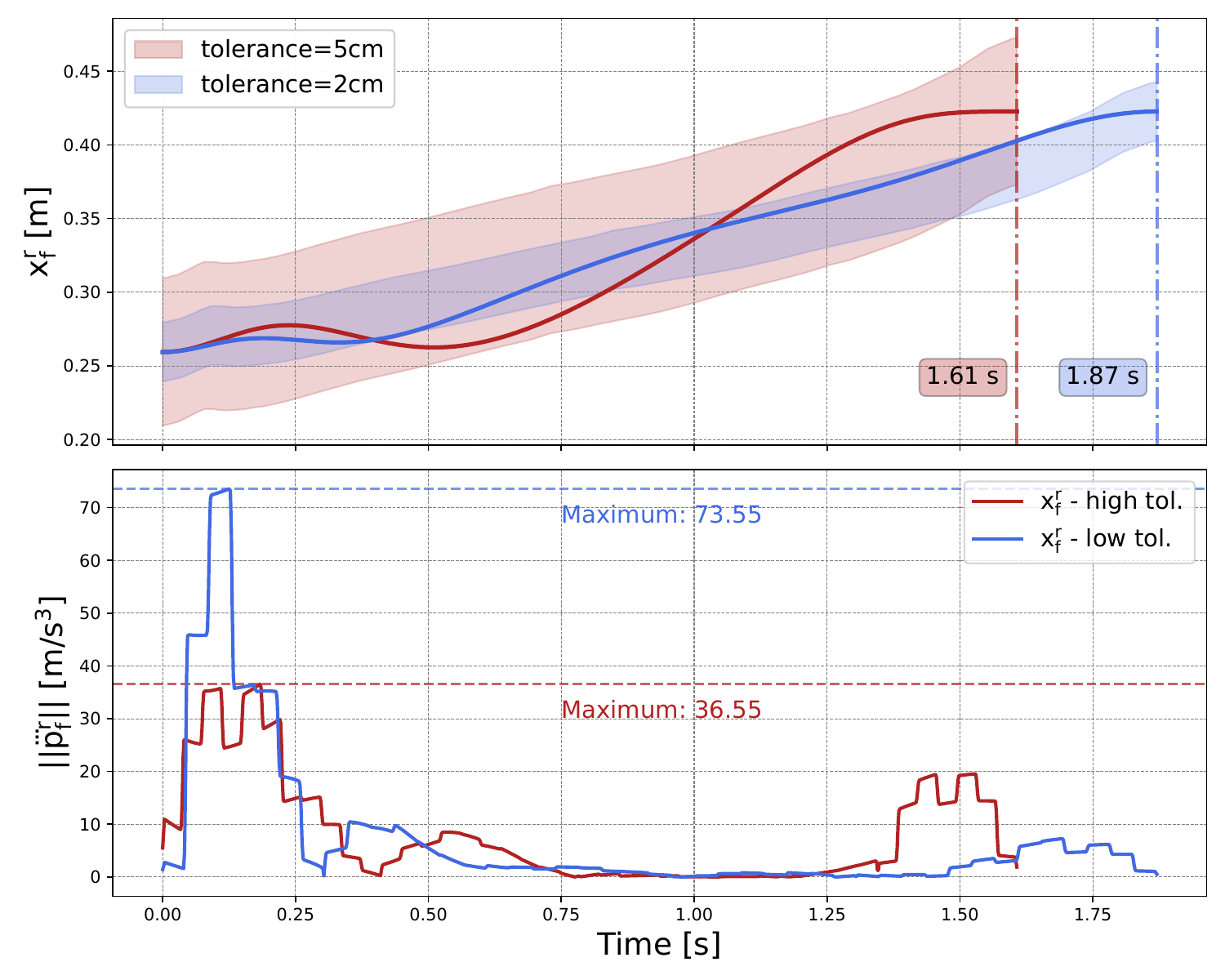}
    \caption{Comparison of $x_f^r(t)$ for RT2, considering optimized execution time and maximum end-effector jerk. Increasing the tolerance bounds enables the optimization to identify faster timing laws with reduced jerk profiles.}
    \label{fig:tols}
\end{figure}

Providing higher tolerance bounds for the optimization will allow the Trajectory Generation module to optimize for a better solution in terms of its objective in \eqref{eq:mainopt0}. It means that higher tolerance bounds lead to lower execution time and lower jerk values. To further pinpoint this, we analyze this aspect by optimizing RT2 via different tolerance bounds of 2 cm and 5 cm across all its waypoints. Fig. \ref{fig:tols} shows the effect of increasing tolerance values in RT2. We can see that via the tolerance bounds of 2 cm, the Optimization-based Smoothing solved for an execution time of 1.87 seconds and a maximum end-effector jerk of 73.55 $m/s^3$, whereas by increasing the tolerance bounds to 5 cm, the execution time is reduced by 15\% to 1.61 seconds, and the maximum end-effector jerk dropped by 50\% to 36.55 $m/s^3$.

\subsubsection{DMP Case Study}
\label{sec:fr3:dmp}
To show the benefits of incorporating DFL-TORO as an intermediate layer between the original demonstrations and the DMP algorithm, we train two sets of DMPs: ${\mathrm{DMP}}_o$ using the original trajectory $\bm{q}_o(t)$ and ${\mathrm{DMP}}_f$ using the optimized trajectory $\bm{q}_f(t)$. These are then used to reproduce the trajectories $\bm{q}_{o,dmp}(t)$ and $\bm{q}_{f,dmp}(t)$, respectively. By applying DFL-TORO prior to DMP training, we anticipate achieving smoother trajectories with a reduced jerk profile. Additionally, by eliminating noise from the original demonstration, we expect ${\mathrm{DMP}}_f$ to produce trajectories that not only adhere to kinematic limits but also have a shorter execution time compared to those generated by ${\mathrm{DMP}}_o$. This setup highlights the enhanced trajectory quality and efficiency gained by using DFL-TORO in the learning process.


Since the original trajectory $\bm{q}_o(t)$ is slower than the optimized trajectory $\bm{q}_f(t)$, we utilize the inherent scaling factor $\tau$ from \eqref{eq:dmp} in ${\mathrm{DMP}}_o$ to generate $\bm{q}_{s,dmp}(t)$, ensuring it matches the duration of $\bm{q}_{f,dmp}(t)$ for a meaningful comparison \cite{ijspeert2013dynamical}. In this context, $\bm{p}_{f,dmp}(t)$, $\bm{p}_{o,dmp}(t)$, and $\bm{p}_{s,dmp}(t)$ denote the corresponding Cartesian position trajectories. 

\begin{figure*}[htbp]
    \centering
    \subfigure[3D path of RT1.]{
        \includegraphics[width=0.4\linewidth]{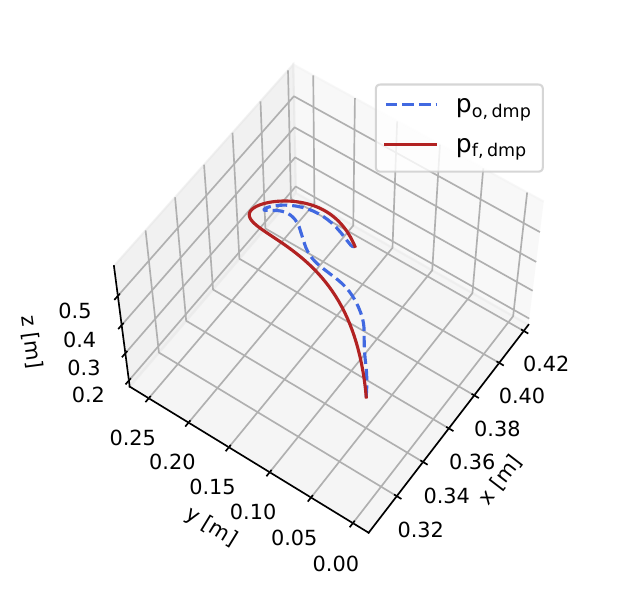}
        \label{fig:3drt1}
    }
    \subfigure[3D path of MT1.]{
        \includegraphics[width=0.4\linewidth]{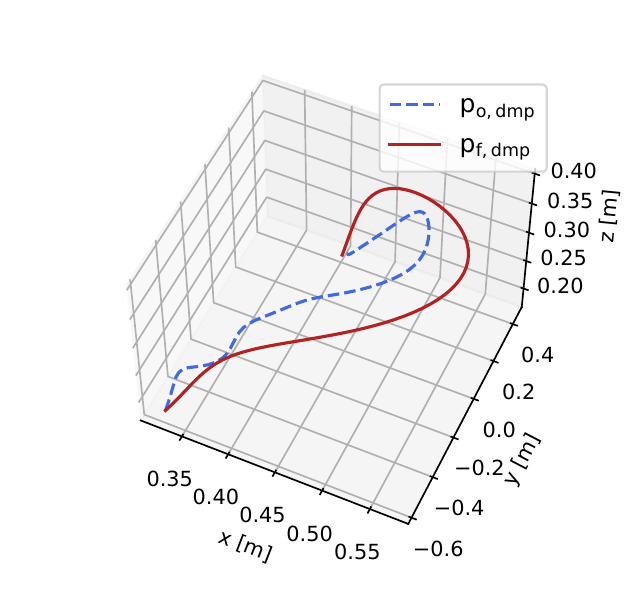}
        \label{fig:3dmt1}
    }
    \subfigure[Kinematic violations for RT1.]{
        \includegraphics[width=0.4\linewidth]{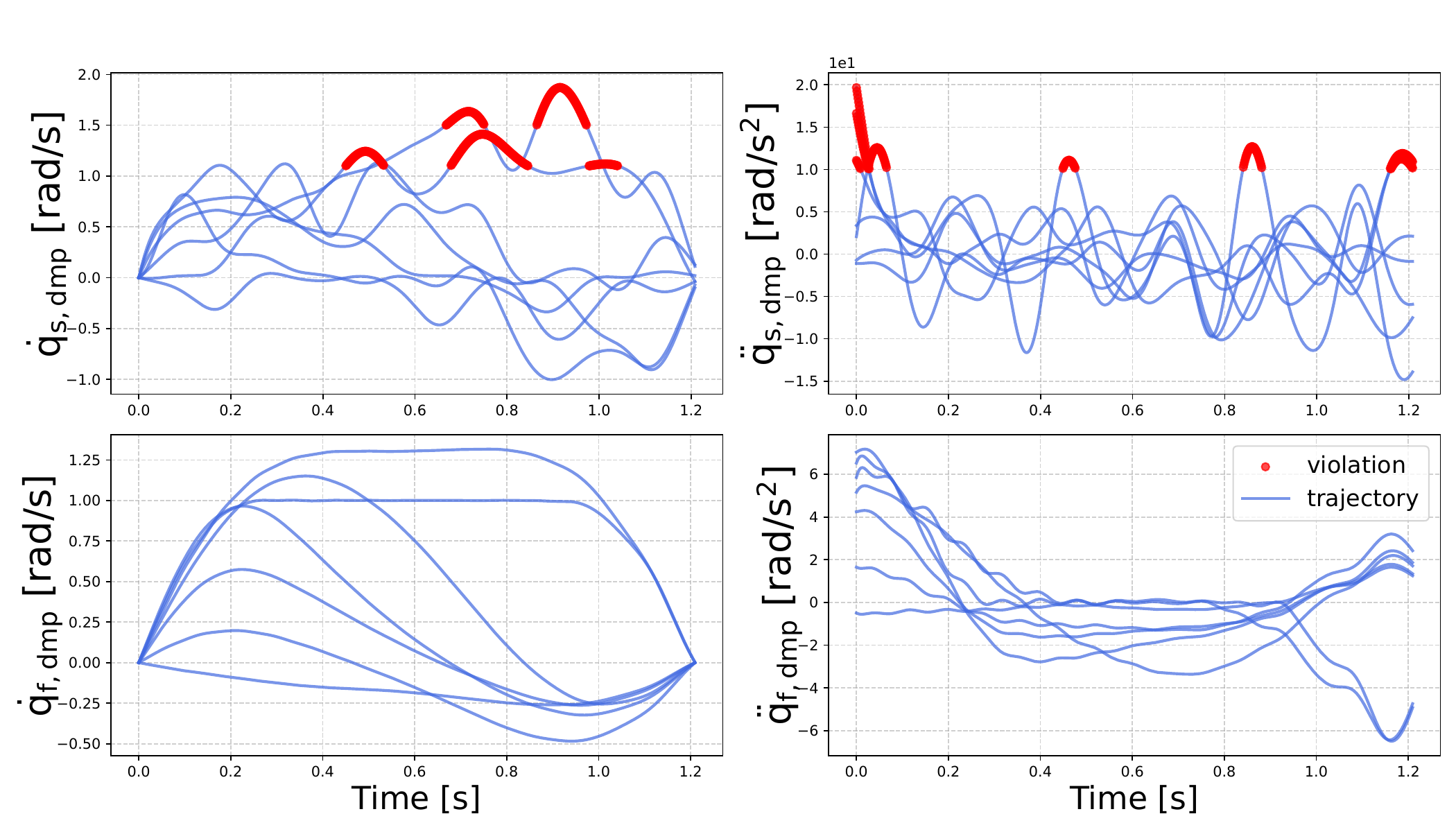}
        \label{fig:kinvolrt1}
    }
    \subfigure[Kinematic violations for MT1.]{
        \includegraphics[width=0.4\linewidth]{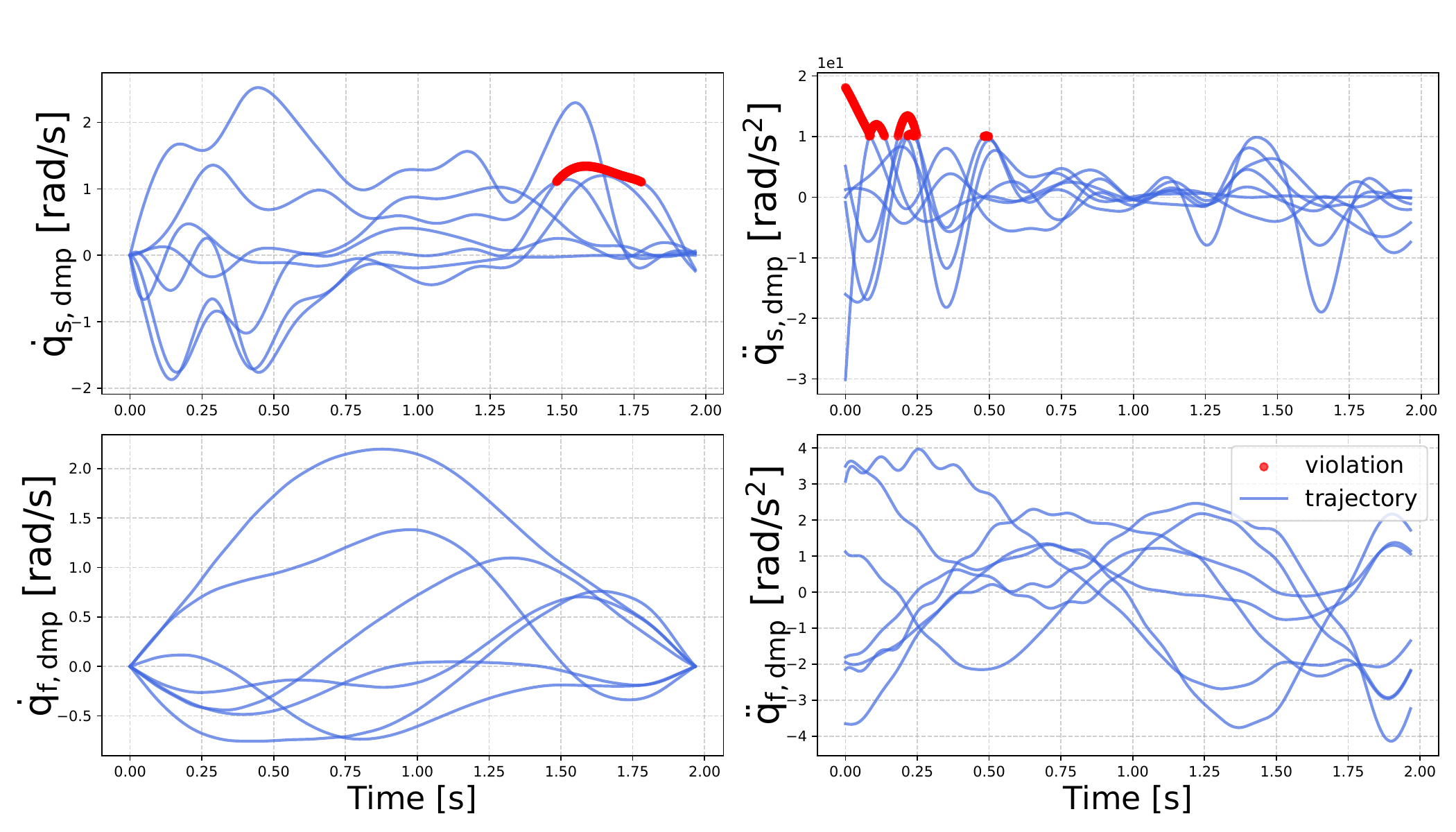}
        \label{fig:kinvolmt1}
    }
    \subfigure[Jerk profiles for RT1.]{
        \includegraphics[width=0.4\linewidth]{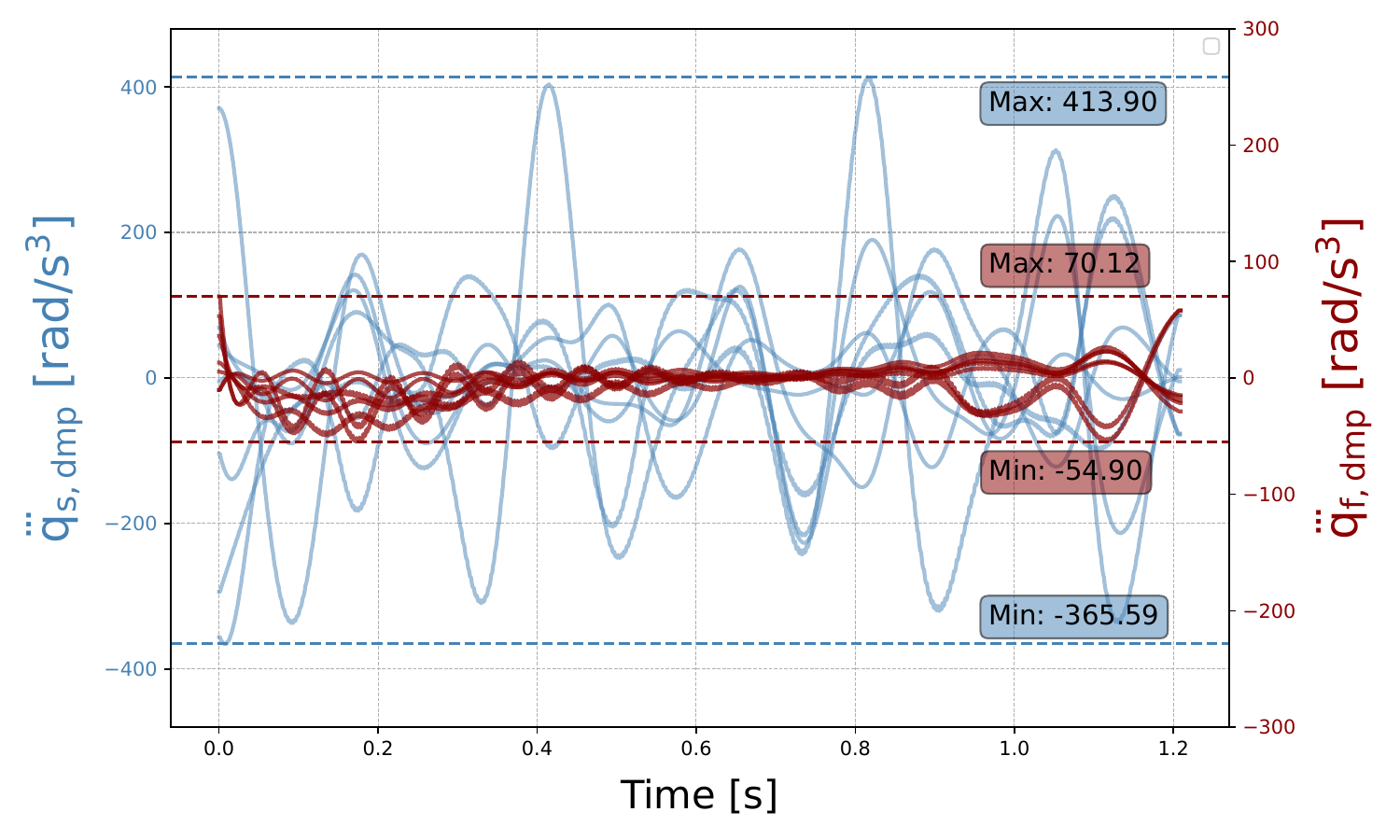}
        \label{fig:jerkrt1}
    }
    \subfigure[Jerk profiles for MT1.]{
        \includegraphics[width=0.4\linewidth]{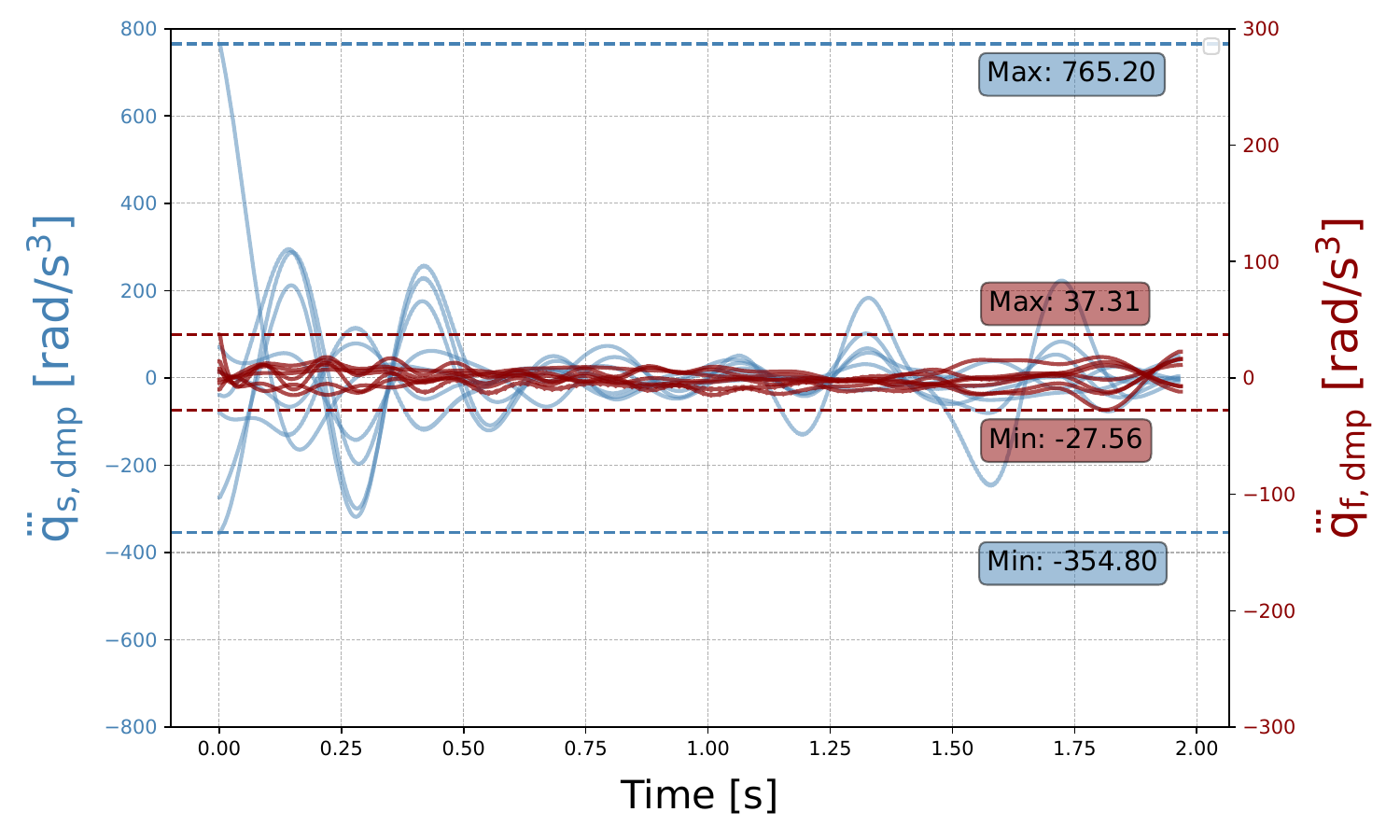}
        \label{fig:jerkmt1}
    }  
    \caption{Performance comparison of ${\mathrm{DMP}}_o$ and ${\mathrm{DMP}}_f$ with regard to kinematic feasibility and jerk profiles. ${\mathrm{DMP}}_f$ not only surpasses ${\mathrm{DMP}}_o$ in kinematic feasibility but also provides a considerably smoother jerk profile.}
    \label{fig:dmpcomp}
\end{figure*}

Figures \ref{fig:3drt1} and \ref{fig:3dmt1} illustrate the paths of $\bm{p}_{o,dmp}(t)$ and $\bm{p}_{f,dmp}(t)$ for RT1 and MT1, respectively. As shown, the path of $\bm{p}_{o,dmp}(t)$ contain unnecessary curves, which are artifacts from fitting over the noise of the original demonstration. Furthermore, the presence of noise results in $\bm{q}_{s,dmp}(t)$ becoming an infeasible trajectory in both cases, as highlighted by the kinematic violations shown in Figures \ref{fig:kinvolrt1} and \ref{fig:kinvolmt1}. In contrast, $\bm{q}_{f,dmp}(t)$ remains feasible within the same execution time. This comparison underscores that removing path noise not only reduces execution time but also leads to a more accurate and feasible timing law, enhancing the overall trajectory quality.


Furthermore, a comparison of the jerk values for $\bm{q}_{s,dmp}(t)$ and $\bm{q}_{f,dmp}(t)$ is presented in Fig. \ref{fig:jerkrt1} and \ref{fig:jerkmt1}, showcasing that $\bm{q}_{f,dmp}(t)$ results in significantly lower jerk values. This improvement is a direct consequence of eliminating noise from the original demonstrations, thereby enhancing the outcomes of the DMP algorithm. Table \ref{tab:comparison} provides a summary of the execution time and jerk values (MANJ) for RT1-RT5 and MT1-MT5. The data clearly indicate that DFL-TORO effectively reduces both execution time and MANJ metrics. This reduction is further verified through the trajectories generated by ${\mathrm{DMP}}_o$ and ${\mathrm{DMP}}_f$.


In addition to the benefits previously mentioned, it is crucial to note that the Gaussian basis functions used during the training process of DMPs, as described in \eqref{eq:f(x)}, inherently provide a smoothing effect on the demonstrations. This smoothing can act as a filter, mitigating the impact of noise in the original demonstrations. However, this effect alone is often insufficient to completely eliminate noise, as residual artifacts can persist. These residual noise elements can still cause inefficiencies in the reproduced trajectories, as observed with ${\mathrm{DMP}}_o$.

\begin{table}

\renewcommand{\arraystretch}{1.2}
\caption{Comparison of time [$s$] and Maximum Absolute Normalized Jerk (MANJ) [$rad/s^3$] for RT1-5 and MT1-5.}
\centering
\resizebox{1\linewidth}{!}{
\begin{NiceTabular}{wc{1.3cm} wc{1.3cm} wc{1.2cm} wc{1.2cm} wc{0.1cm} wc{1.2cm} wc{1.2cm}}
\CodeBefore
  \rectanglecolor{mygray}{3-2}{3-7}
  \rectanglecolor{mygray}{5-2}{5-7}
  \rectanglecolor{mygray}{7-2}{7-7}
  \rectanglecolor{mygray}{9-2}{9-7}
  \rectanglecolor{mygray}{11-2}{11-7}
  \rectanglecolor{mygray}{13-2}{13-8}
  \rectanglecolor{mygray}{15-2}{15-8}
  \rectanglecolor{mygray}{17-2}{17-8}
  \rectanglecolor{mygray}{19-2}{19-8}
  \rectanglecolor{mygray}{21-2}{21-8}
\Body 
\Xhline{3\arrayrulewidth}

\Block{2-1}{\bf Experiment} & \Block{2-1}{\bf Time/Jerk} &  \Block{1-2}{\bf Demonstration} & &  & \Block{1-2}{\bf DMP} &      \\
                                                     \cmidrule{3-4} \cmidrule{6-7} 
                                                      & & \bf $\bm{q}_o(t)$ & \bf $\bm{q}_f(t)$ & & \bf $\bm{q}_{o,dmp}(t)$ & \bf $\bm{q}_{f,dmp}(t)$  \\

\midrule
\Block{2-1}{RT1} & Time & 8.88 & \textbf{1.21} &  & 8.88 & \textbf{1.21}  \\
                    &  MANJ & 15358.07 & \textbf{78.31} &  & 2305.95 & \textbf{124.00}                        \\
\Xhline{2\arrayrulewidth}
\Block{2-1}{RT2} & Time & 11.15 & \textbf{1.87} &  & 11.15 & \textbf{1.87}  \\
                    &  MANJ & 241452.27 & \textbf{771.83} &  & 2822.72 & \textbf{582.82}                        \\
\Xhline{2\arrayrulewidth}
\Block{2-1}{RT3} & Time & 11.82 & \textbf{1.98} &  & 11.82 & \textbf{1.98}  \\
                    &  MANJ & 62706.92 & \textbf{82.25} &  & 2028.04 & \textbf{164.84}                        \\
\Xhline{2\arrayrulewidth}
\Block{2-1}{RT4} & Time & 7.08 & \textbf{1.22} &  & 7.08 & \textbf{1.22}  \\
                    &  MANJ & 9048.28 & \textbf{97.33} &  & 678.59 & \textbf{163.01}                        \\
\Xhline{2\arrayrulewidth}
\Block{2-1}{RT5} & Time & 19.75 & \textbf{2.21} &  & 19.75 & \textbf{2.21}  \\
                    &  MANJ & 101934.88 & \textbf{207.85} &  & 2993.82 & \textbf{358.07}                        \\
\Xhline{2\arrayrulewidth}
\Block{2-1}{MT1} & Time & 10.55 & \textbf{1.97} &  & 10.55 & \textbf{1.97 } \\
                    &  MANJ & 216390.32 & \textbf{132.40} &  & 5781.34 & \textbf{284.09}                        \\
\Xhline{2\arrayrulewidth}
\Block{2-1}{MT2} & Time & 14.35 & \textbf{1.49} &  & 14.35 & \textbf{1.49}  \\
                    &  MANJ & 263475.13 & \textbf{146.01} &  & 2101.84 & \textbf{152.09}                        \\
\Xhline{2\arrayrulewidth}
\Block{2-1}{MT3} & Time & 9.82 & \textbf{1.11} &  & 9.82 & \textbf{1.11}  \\
                    &  MANJ & 22481.38 & \textbf{115.51} &  & 1366.84 & \textbf{172.86}                        \\
\Xhline{2\arrayrulewidth}
\Block{2-1}{MT4} & Time & 14.95 & \textbf{1.31} &  & 14.95 & \textbf{1.31}  \\
                    &  MANJ & 97762.75 & \textbf{353.56} &  & 10462.93 & \textbf{315.19}                        \\
\Xhline{2\arrayrulewidth}
\Block{2-1}{MT5} & Time & 8.08 & \textbf{0.95} &  & 8.08 & \textbf{0.95}  \\
                    &  MANJ & 23539.90 & \textbf{102.12} &  & 1533.07 & \textbf{115.79}                        \\

\Xhline{3\arrayrulewidth}
\end{NiceTabular}
}\\
\label{tab:comparison}
\end{table}

\subsection{Experiments with ABB YuMi}

\begin{figure*}[htbp]
    \centering
    \begin{minipage}{0.9\textwidth}
    \subfigure[YM1-1]{
        \includegraphics[width=0.23\linewidth]{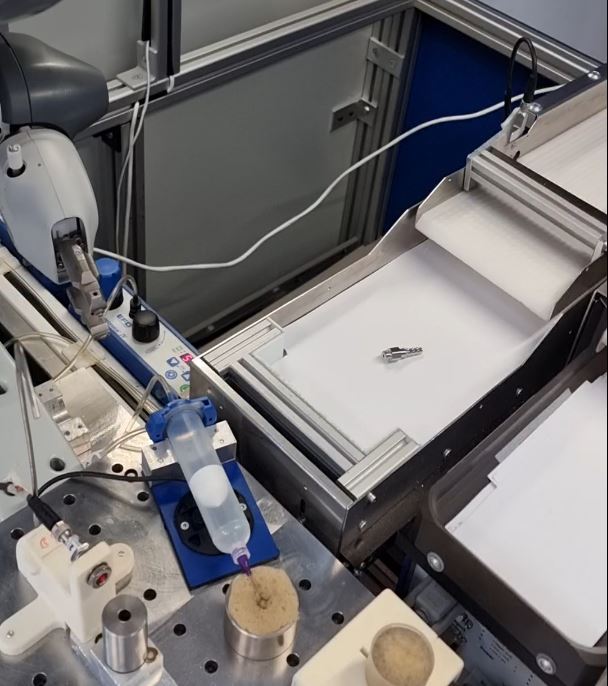}
        \label{fig:yumitask1}
    }
    \subfigure[YM1-2]{
        \includegraphics[width=0.23\linewidth]{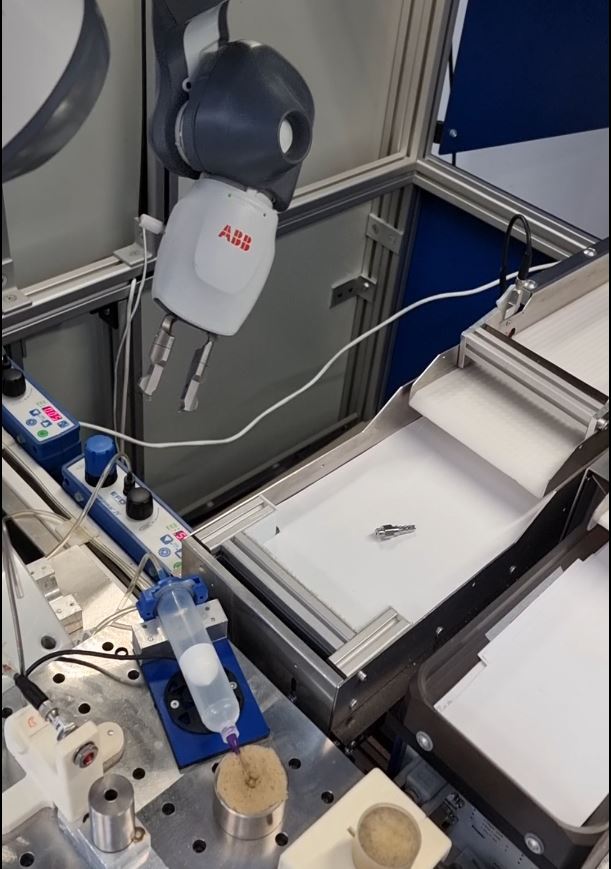}
        \label{fig:yumitask2}
    }
    \subfigure[YM1-3]{
        \includegraphics[width=0.23\linewidth]{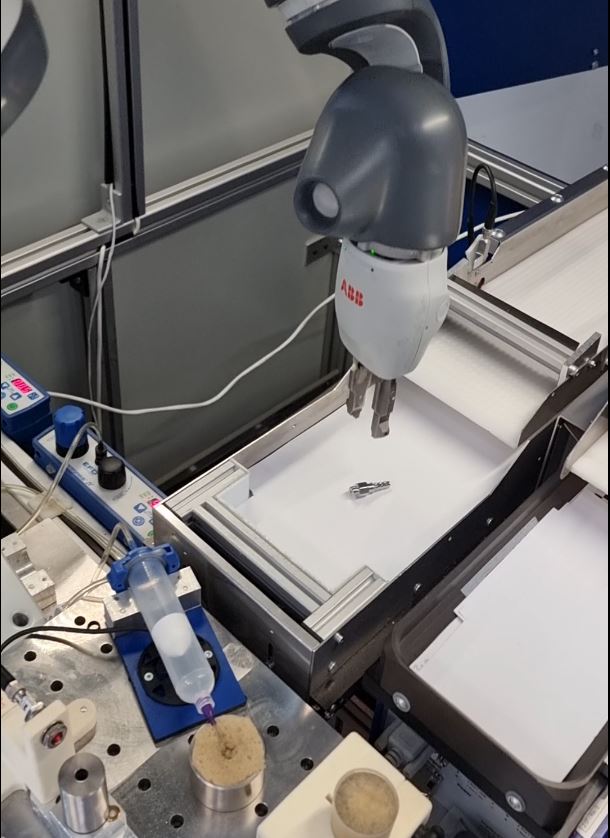}
        \label{fig:yumitask3}
    }
    \subfigure[YM1-4]{
        \includegraphics[width=0.23\linewidth]{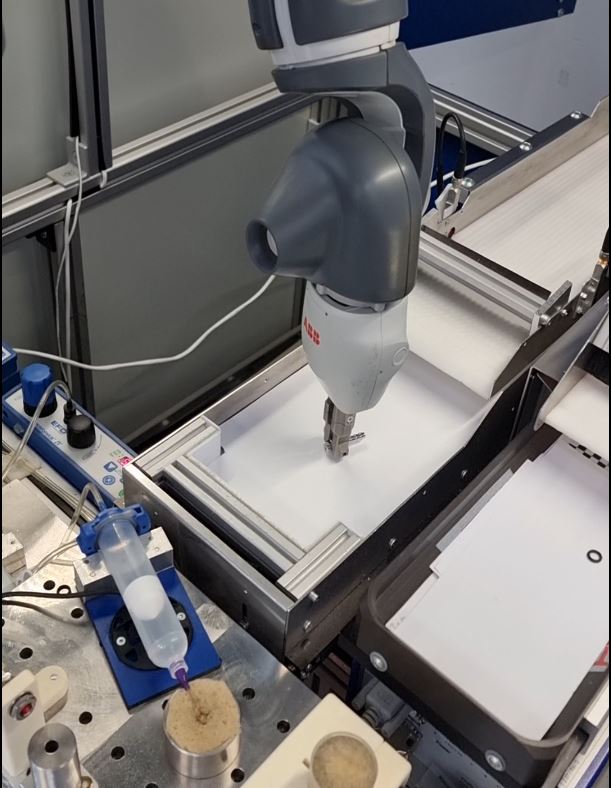}
        \label{fig:yumitask4}
    }
    \subfigure[YM2-1]{
        \includegraphics[width=0.23\linewidth]{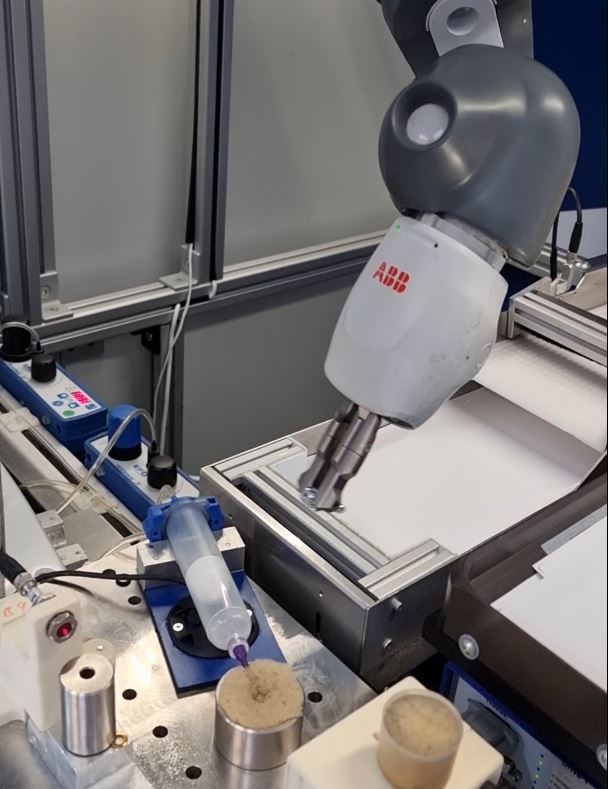}
        \label{fig:yumitask5}
    }
    \subfigure[YM2-2]{
        \includegraphics[width=0.23\linewidth]{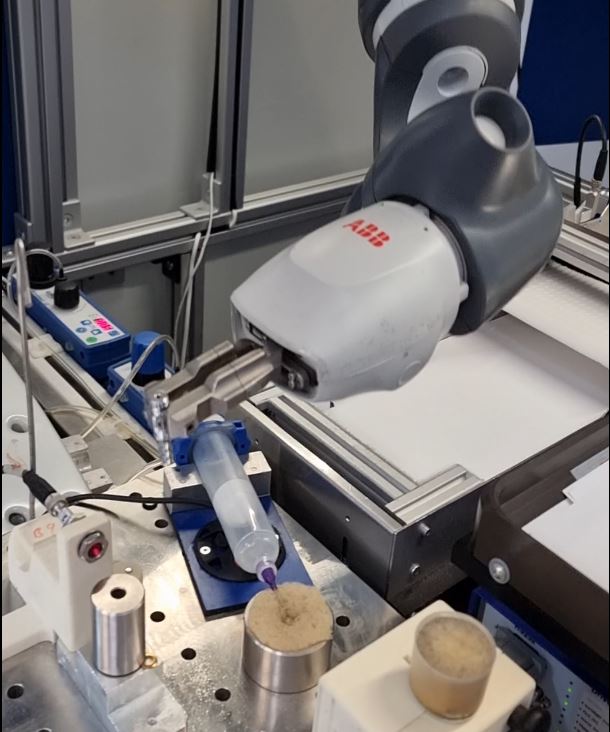}
        \label{fig:yumitask6}
    }
    \subfigure[YM2-3]{
        \includegraphics[width=0.23\linewidth]{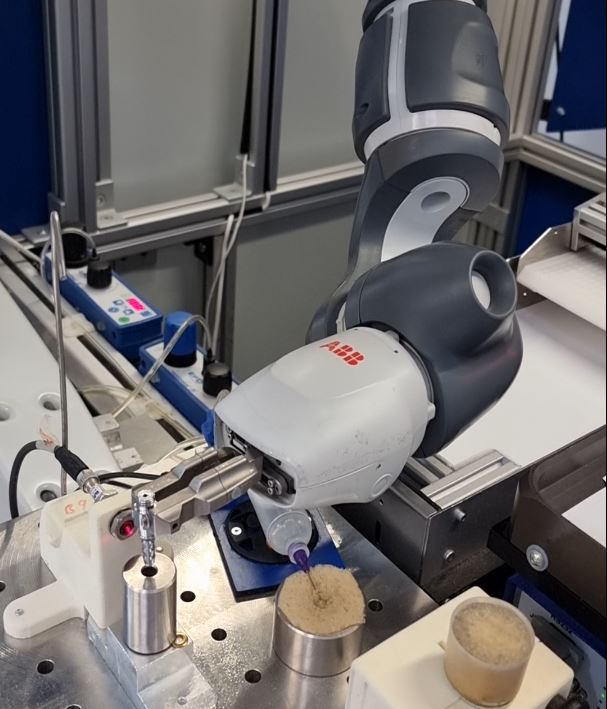}
        \label{fig:yumitask7}
    }
    \subfigure[YM2-4]{
        \includegraphics[width=0.23\linewidth]{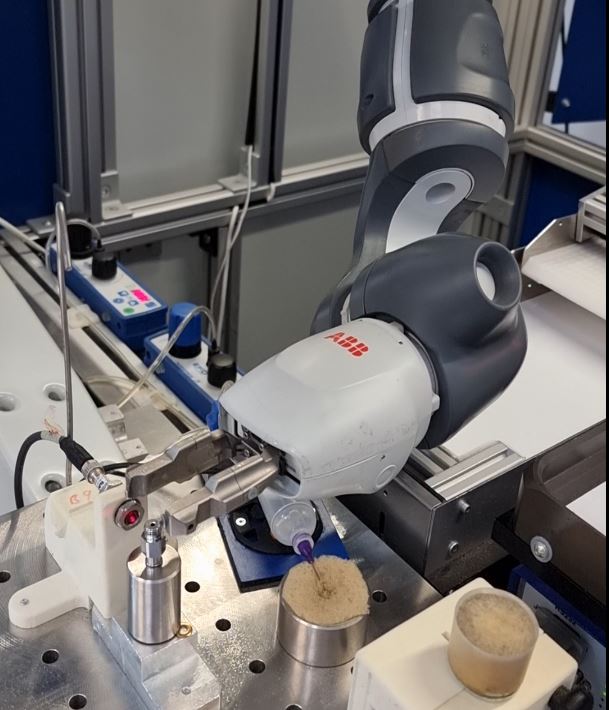}
        \label{fig:yumitask8}
    }
    \subfigure[YM3-1]{
        \includegraphics[width=0.23\linewidth]{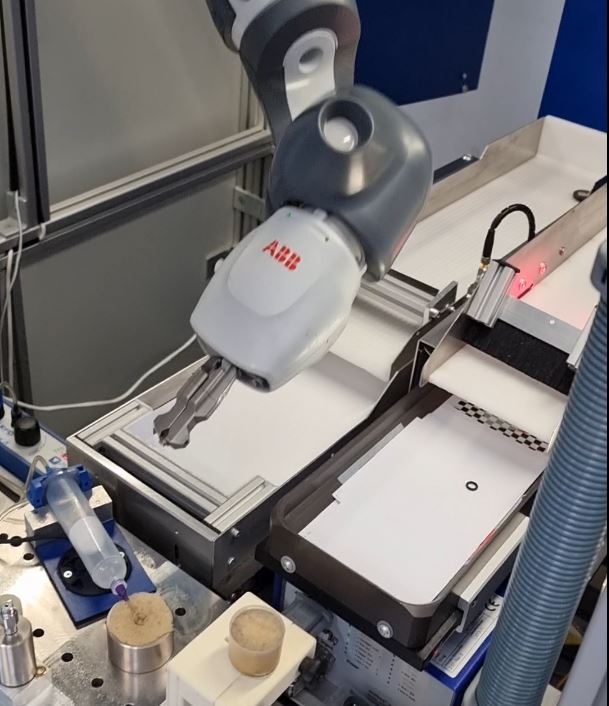}
        \label{fig:yumitask9}
    }
    \subfigure[YM3-2]{
        \includegraphics[width=0.23\linewidth]{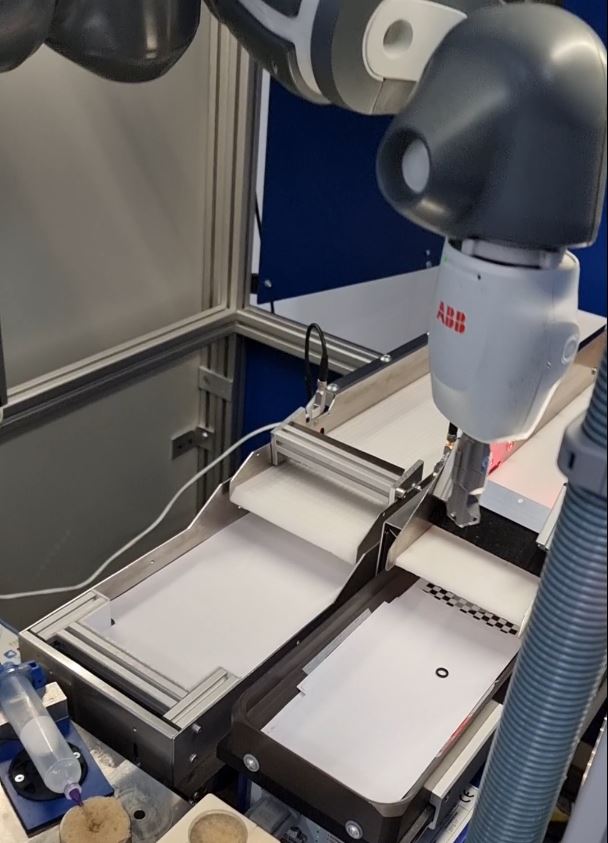}
        \label{fig:yumitask10}
    }
    \subfigure[YM3-3]{
        \includegraphics[width=0.23\linewidth]{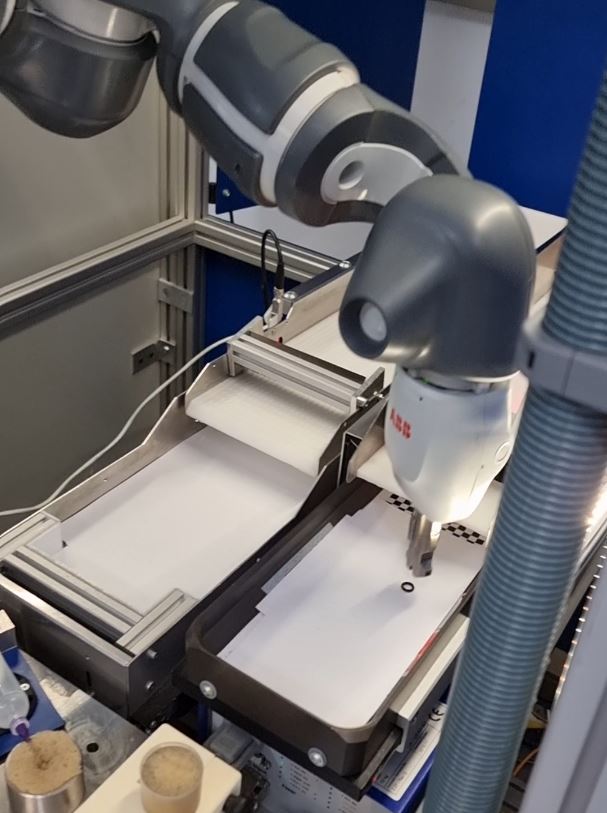}
        \label{fig:yumitask11}
    }
    \subfigure[YM3-4]{
        \includegraphics[width=0.23\linewidth]{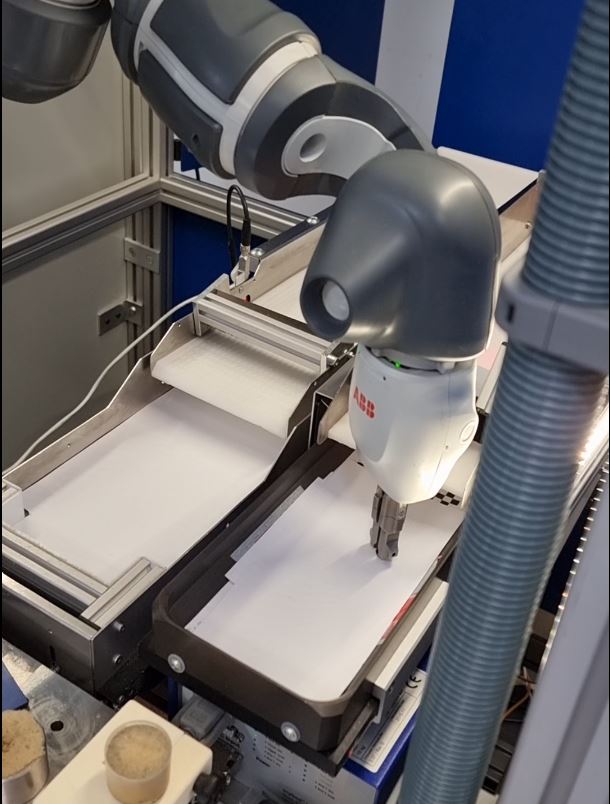}
        \label{fig:yumitask12}
    }
    \subfigure[YM4-1]{
        \includegraphics[width=0.23\linewidth]{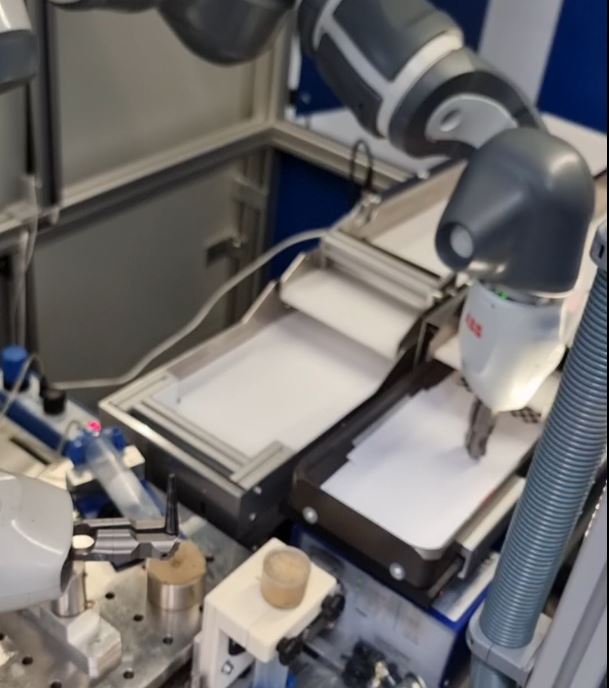}
        \label{fig:yumitask13}
    }
    \subfigure[YM4-2]{
        \includegraphics[width=0.23\linewidth]{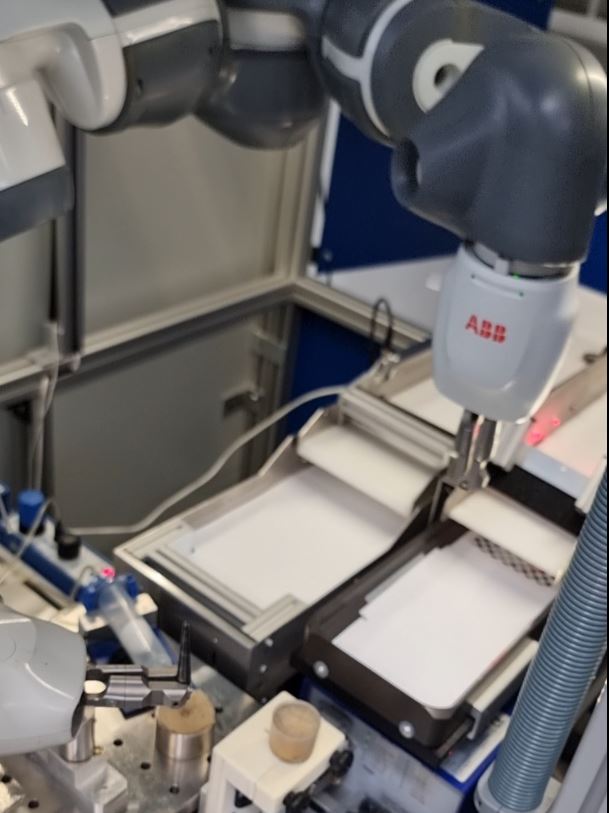}
        \label{fig:yumitask14}
    }
    \subfigure[YM4-3]{
        \includegraphics[width=0.23\linewidth]{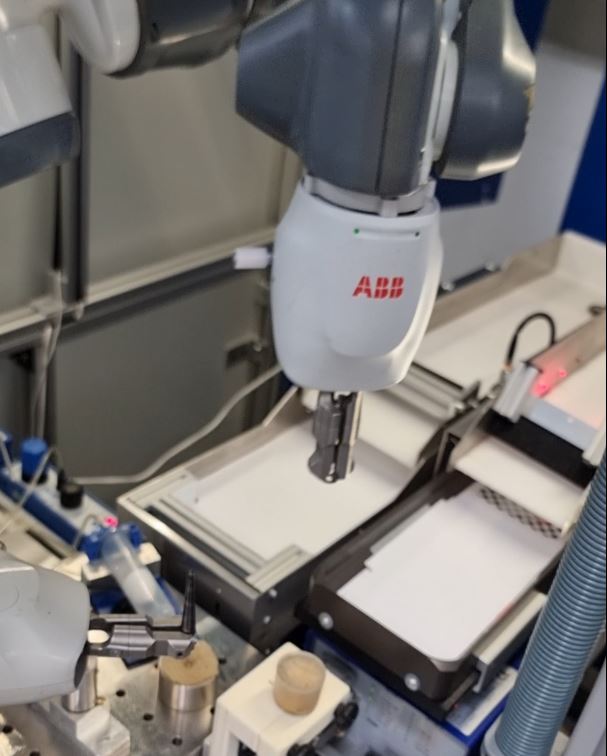}
        \label{fig:yumitask15}
    }
    \subfigure[YM4-4]{
        \includegraphics[width=0.23\linewidth]{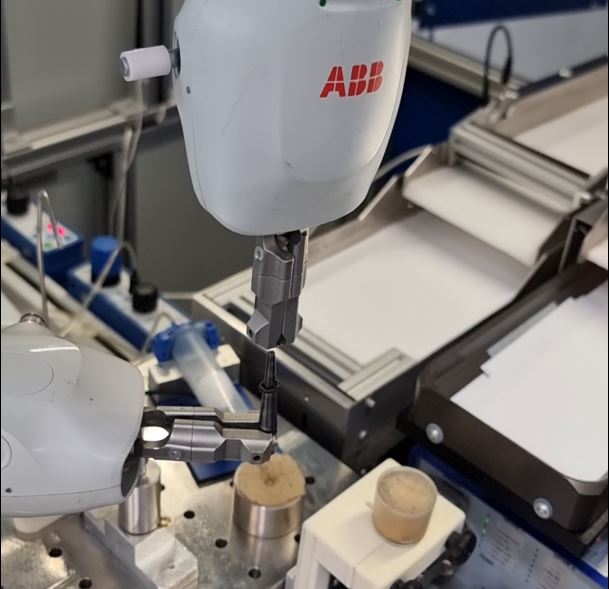}
        \label{fig:yumitask16}
    }
    \caption{Visualization of YuMi tasks YM1 to YM4 in the production site.}
    \label{fig:yumitask}
    \end{minipage}
\end{figure*}

The experiments with the ABB YuMi include automating a portion of a real manufacturing task conducted at a production site. This task involved using both arms of the ABB YuMi to assemble a component of a larger product. In this study, we partially automated this task using DFL-TORO and DMPs. The traditional programming method for the motions of the left YuMi arm was replaced by an LfD approach, optimized by DFL-TORO and generalized through DMPs. Figure \ref{fig:yumitask} illustrates the steps of the subtasks automated via LfD, designated as YM1-YM4. The details of these tasks are described as follows:


\begin{enumerate}
    \item \textbf{YM1}: The left arm reaches for the metallic object to grasp it from the tray (Fig. \ref{fig:yumitask1} to Fig. \ref{fig:yumitask4}). The object's location is detected by an overhead camera and transformed into the robot's coordinate system. These coordinates are then converted into the desired joint configuration, which serves as the goal configuration for the DMP. Using this information, YM1 generalizes the trajectory to grasp the object.
    \item \textbf{YM2}: The left arm moves the object to its desired location, releasing it into its mounting station (Fig. \ref{fig:yumitask5} to Fig. \ref{fig:yumitask8}). 
    \item \textbf{YM3}: The left arm reaches for the plastic ring on the other tray, grasping it from the inside  (Fig. \ref{fig:yumitask9} to Fig. \ref{fig:yumitask12}). Similarly, the location of the ring is detected via the overhead camera. 
    \item \textbf{YM4}: The left arm moves the ring on top of an auxiliary cone, held by the right arm, releasing it to fall on the cone (Fig. \ref{fig:yumitask13} to Fig. \ref{fig:yumitask16}). From here, the right arm mounts the cone on top of the metallic object and inserts the ring onto it. The right arm is programmed in a traditional way.
\end{enumerate}

The task specifications of YM1-YM4 present a complex and precise manufacturing task in a small-part assembly process. Fig. \ref{fig:visym1} to \ref{fig:visym4} show the recorded demonstration trajectories. It is important to note that due to safety measures, the lead-through mode of industrial robots such as ABB YuMi inherently produces more noise in recorded demonstrations compared to the gravity compensation mode used in the FR3. This is because, in Lead-through mode, the robot's movements are guided by joint torque sensors that detect the forces applied by the operator. Unlike the gravity compensation mode, which can maintain smooth and steady movements by counteracting gravitational forces, the Lead-through mode can have sudden movements or stops based on human input. This can lead to inconsistencies and fluctuations in the trajectory, resulting in noisier data. Therefore, in these experiments, the role of noise removal becomes even more crucial.

\subsubsection{DMP Case Study}
\begin{figure*}[htbp]
    \centering
    \subfigure[YM1]{
        \includegraphics[width=0.22\linewidth]{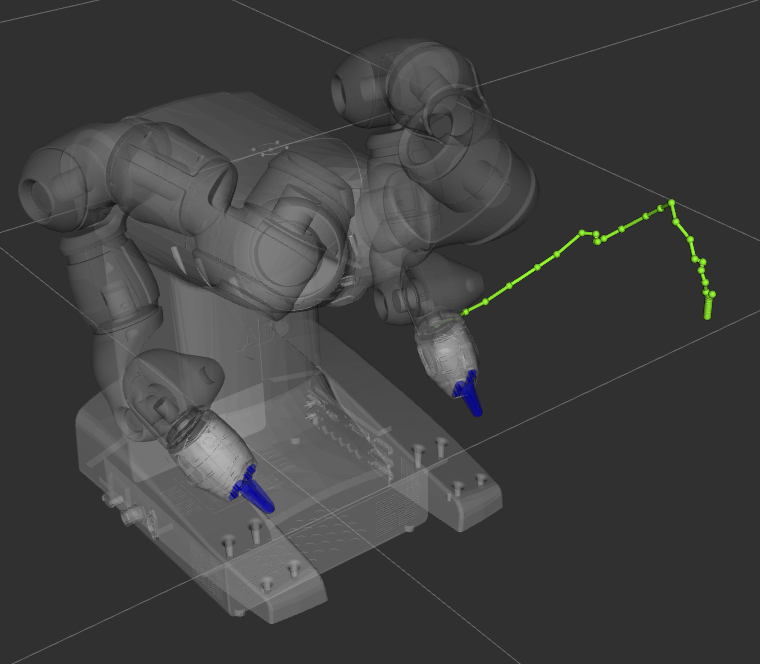}
        \label{fig:visym1}
    }
    \subfigure[YM2]{
        \includegraphics[width=0.22\linewidth]{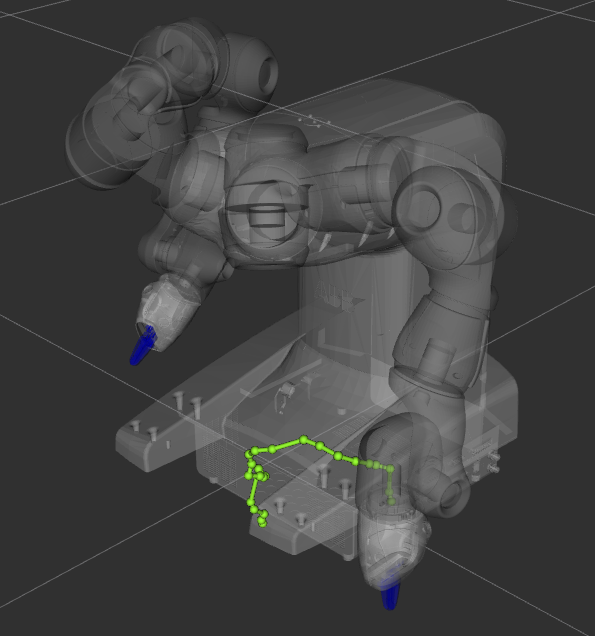}
        \label{fig:visym2}
    }
    \subfigure[YM3]{
        \includegraphics[width=0.22\linewidth]{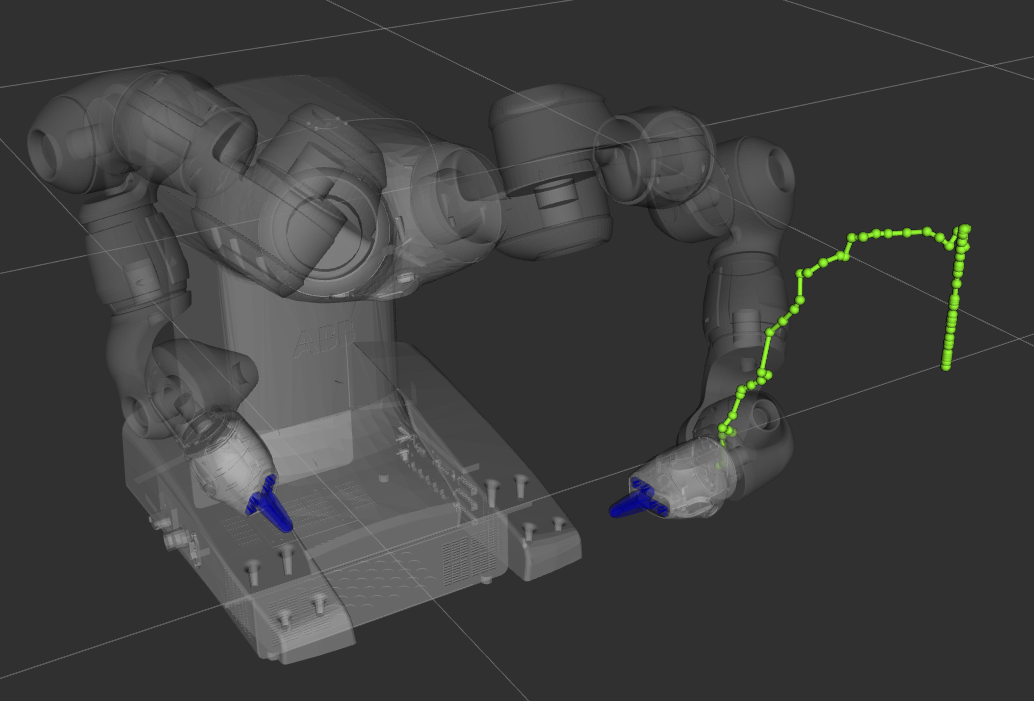}
        \label{fig:visym3}    
    }
    \subfigure[YM4]{
        \includegraphics[width=0.22\linewidth]{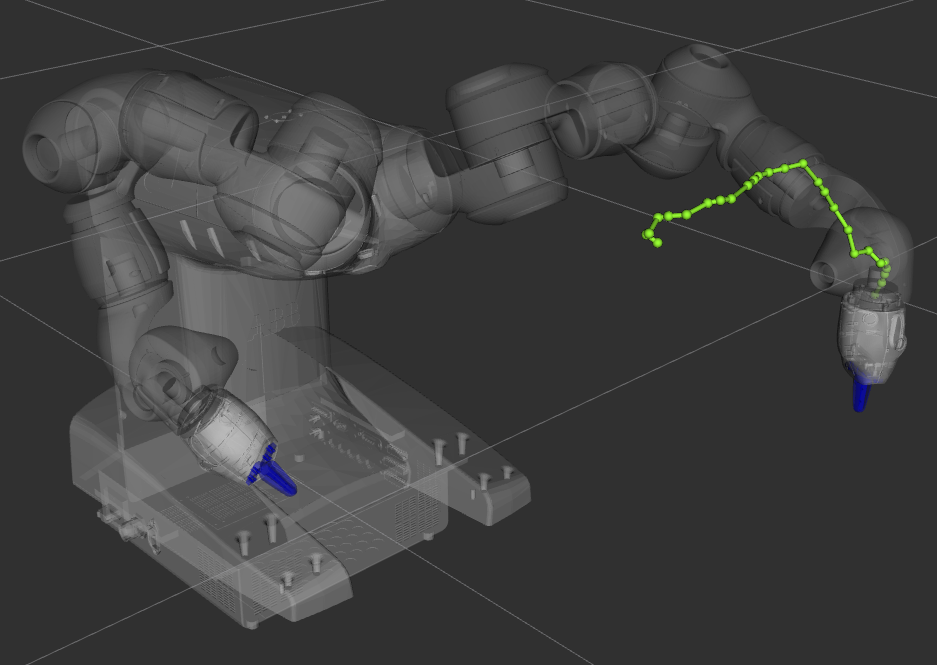}
        \label{fig:visym4}
    }
    \subfigure[3D path of YM1]{
        \includegraphics[width=0.22\linewidth]{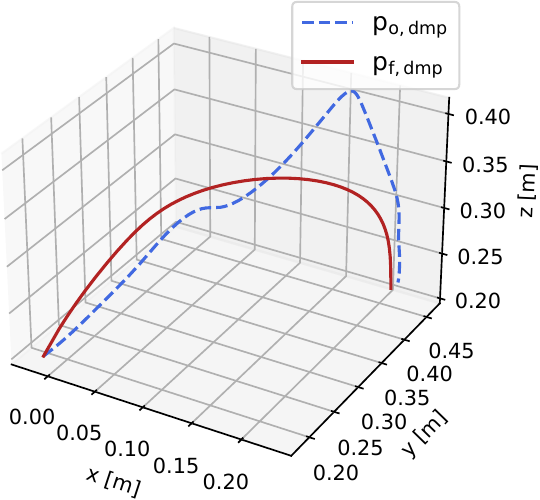}
        \label{fig:pathym1}
    }
    \subfigure[3D path of YM2]{
        \includegraphics[width=0.22\linewidth]{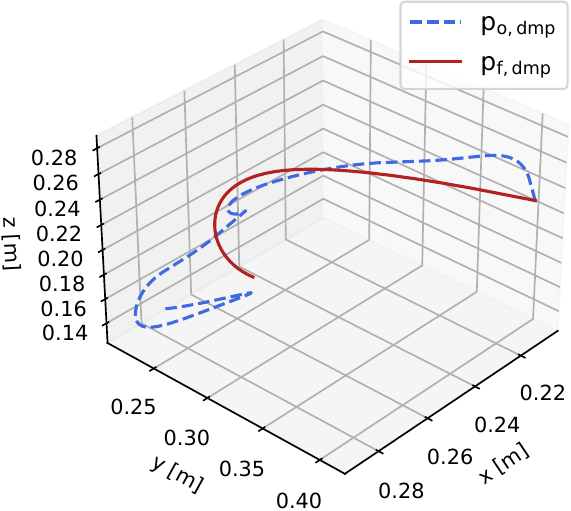}
        \label{fig:pathym2}
    }
    \subfigure[3D path of YM3]{
        \includegraphics[width=0.22\linewidth]{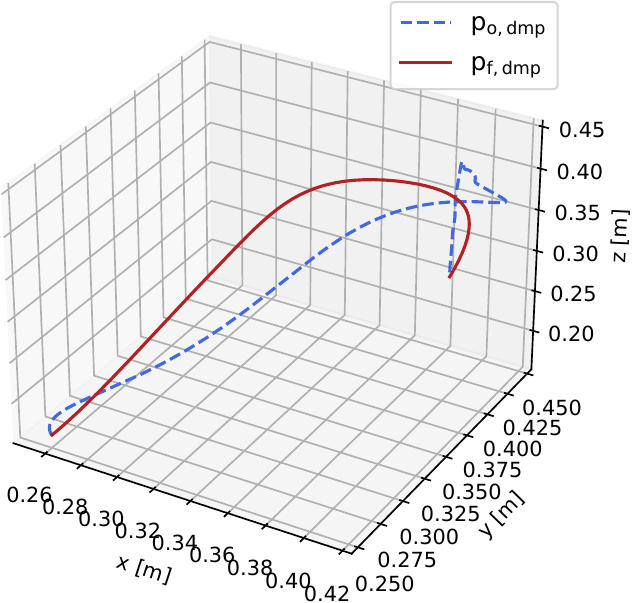}
        \label{fig:pathym3}
    }
    \subfigure[3D path of YM4]{
        \includegraphics[width=0.22\linewidth]{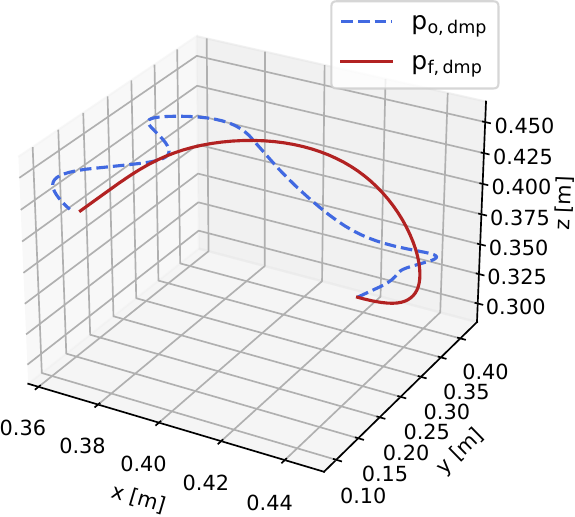}
        \label{fig:pathym4}
    }
    \caption{Visualization of demonstration trajectories recorded using ABB YuMi, alongside the 3D paths generalized by ${\mathrm{DMP}}_o$ and ${\mathrm{DMP}}_f$. The paths generated by ${\mathrm{DMP}}_o$ show noticeable effects of noise, which can cause the robot to collide with obstacles in the workspace or impact execution efficiency.}
    \label{fig:yumivisual}
\end{figure*}

Similar to Section \ref{sec:fr3:dmp} we train two sets of DMPs: ${\mathrm{DMP}}_o$ using the original trajectories $\bm{q}_o(t)$ and ${\mathrm{DMP}}_f$ using the optimized trajectories $\bm{q}_f(t)$, reproducing $\bm{q}_{o,dmp}(t)$ and $\bm{q}_{f,dmp}(t)$, respectively, for the tasks YM1-YM4. Fig. \ref{fig:pathym1} to \ref{fig:pathym4} illustrate the paths of $\bm{p}_{o,dmp}(t)$ and $\bm{p}_{f,dmp}(t)$ for these tasks.


The presence of noise in the original demonstrations led to several reliability and safety issues, as evidenced by overfitting effects in each task. In YM1, $\bm{p}_{o,dmp}(t)$ shows an unnecessary movement by rising too far above the object tray before approaching the object. In YM2, the noise significantly affects the motion towards the end, causing the robot to potentially collide with the mounting station for the metallic object. Additionally, the path of $\bm{p}_{o,dmp}(t)$ fails to reach the goal configuration, requiring additional time for the DMP to converge to the goal. In YM3, the noise introduces an unnecessary curve in the end-effector's path before it descends to approach the ring, risking a collision with the back of the ring tray. Finally, in YM4, the noise results in several unnecessary curves, which negatively impact the efficiency of the motion execution.


Following a similar approach to Section \ref{sec:fr3:dmp}, Table \ref{tab:comparison2} provides a summary of the time and jerk values for YM1-YM4. The data highlight a significant improvement in both execution time and MANJ when DFL-TORO is incorporated into the LfD process. As reflected in the underlying path of $\bm{p}_{o,dmp}(t)$, the inherent noise-removal capabilities of standard LfD algorithms alone are insufficient to achieve optimal trajectories in a manufacturing context, where execution time and jerk are critical factors. This underscores the importance of using DFL-TORO, which significantly enhances the quality and efficiency of the generated trajectories by effectively filtering out noise and optimizing the motion.


\begin{table}

\renewcommand{\arraystretch}{1.2}
\caption{Comparison of time [$s$] and Maximum Absolute Normalized Jerk (MANJ) [$rad/s^3$] for YM1-4.}
\centering
\resizebox{1\linewidth}{!}{
\begin{NiceTabular}{wc{1.3cm} wc{1.3cm} wc{1.2cm} wc{1.2cm} wc{0.1cm} wc{1.2cm} wc{1.2cm}}
\CodeBefore
  \rectanglecolor{mygray}{3-2}{3-7}
  \rectanglecolor{mygray}{5-2}{5-7}
  \rectanglecolor{mygray}{7-2}{7-7}
  \rectanglecolor{mygray}{9-2}{9-7}
  \rectanglecolor{mygray}{11-2}{11-7}
\Body 
\Xhline{3\arrayrulewidth}

\Block{2-1}{\bf Experiment} & \Block{2-1}{\bf Time/Jerk} &  \Block{1-2}{\bf Demonstration} & &  & \Block{1-2}{\bf DMP} &      \\
                                                     \cmidrule{3-4} \cmidrule{6-7} 
                                                      & & \bf $\bm{q}_o(t)$ & \bf $\bm{q}_f(t)$ & & \bf $\bm{q}_{o,dmp}(t)$ & \bf $\bm{q}_{f,dmp}(t)$  \\

\midrule
\Block{2-1}{YM1} & Time & 9.75 & \textbf{1.50} &  & 9.75 & \textbf{1.50}  \\
                    &  MANJ & 139034.03 & \textbf{492.11} &  & 2386.13 & \textbf{433.93}                        \\
\Xhline{2\arrayrulewidth}
\Block{2-1}{YM2} & Time & 13.35 & \textbf{1.10} &  & 13.35 & \textbf{1.10}  \\
                    &  MANJ & 16148.34 & \textbf{114.13} &  & 3798.41 & \textbf{168.37}                        \\
\Xhline{2\arrayrulewidth}
\Block{2-1}{YM3} & Time & 79.75 & \textbf{1.28} &  & 79.75 & \textbf{1.28}  \\
                    &  MANJ & 5152541.33 & \textbf{232.17} &  & 16996.27 & \textbf{227.79}                        \\
\Xhline{2\arrayrulewidth}
\Block{2-1}{YM4} & Time & 14.15 & \textbf{1.19} &  & 14.15 & \textbf{1.19}  \\
                    &  MANJ & 153770.69 & \textbf{190.80} &  & 1839.68 & \textbf{634.43}                        \\
\Xhline{2\arrayrulewidth}

\Xhline{3\arrayrulewidth}
\end{NiceTabular}
}\\
\label{tab:comparison2}
\end{table}

\section{Conclusion}
\label{sec:conclusion}




In this paper, we presented DFL-TORO, a novel demonstration framework within the LfD process, addressing the poor quality and efficiency of human demonstrations. The framework captures task requirements intuitively with no need for multiple demonstrations, as well as locally adjusting the velocity profile. The obtained trajectory was ensured to be time-optimal, noise-free, and jerk-regulated while satisfying the robot's kinematic constraints. The effectiveness of DFL-TORO was experimentally evaluated using a FR3 in several reaching and moving task scenarios. Moreover, a case study via DMP showed that optimizing the demonstrations via DFL-TORO before the LfD algorithm significantly enhances the quality of generalized trajectories. This aspect is further showcased in a real manufacturing task operated by ABB YuMi.

The methods and approaches proposed in this study for the DFL-TORO workflow have some limitations that could be addressed in future work. In the Optimization-based Smoothing module, the performance of Trajectory Generation depends on the default tolerance values, $\bm{\epsilon}_p^d$ and $\epsilon^d_\theta$. Currently, these values must be manually assigned, which limits the framework's autonomy. Since these default values can vary based on the required task accuracy, improperly defined values may lead to inaccuracies. Automating this process through a semantic understanding of the task could improve accuracy and consistency.

Another issue concerns the Refinement Phase, where the human teacher refines the trajectory replayed at a slower speed $\mathrm{v}_{min}^r$. While refining in the actual speed of $\bm{q}_f(t)$ is cognitively demanding for the human teacher, refining at a slower speed than the actual operational speed may require the human teacher to adjust to this mismatch, potentially affecting the intuitiveness of the process. Additionally, future research could delve deeper into how tolerance values encapsulate task-specific semantic nuances and explore ways to enhance the richness of information conveyed through these values.

In the overall DFL-TORO workflow, step (C) involves replaying $\bm{q}_f(t)$ and having a human supervisor assess whether refinement is needed. This step can pose risks, especially if the robot's workspace is cluttered with obstacles or if the manipulated object is delicate, potentially leading to damage or safety hazards. To mitigate these risks, this step could be enhanced with Augmented Reality (AR), allowing the human supervisor to review the motion virtually without executing it physically. Finally, to further assess the intuitiveness and practical utility of the workflow, a user study involving industry experts, such as process engineers, in real manufacturing settings is beneficial.

\newpage
\bibliographystyle{ieeetr}
\bibliography{references}

\end{document}